\documentclass[letterpaper]{article} % DO NOT CHANGE THIS
\usepackage{aaai2026}  % DO NOT CHANGE THIS
\usepackage{times}  % DO NOT CHANGE THIS
\usepackage{helvet}  % DO NOT CHANGE THIS
\usepackage{courier}  % DO NOT CHANGE THIS
\usepackage[hyphens]{url}  % DO NOT CHANGE THIS
\usepackage{graphicx} % DO NOT CHANGE THIS
\usepackage{natbib}  % DO NOT CHANGE THIS AND DO NOT ADD ANY OPTIONS TO IT
\usepackage{caption} % DO NOT CHANGE THIS AND DO NOT ADD ANY OPTIONS TO IT
\usepackage{algorithm}
\usepackage{algorithmic}
\usepackage{subfigure}
\usepackage{amsmath}
\usepackage{amssymb}
\usepackage{xcolor}
\usepackage{boldline}
\usepackage{multicol}
\usepackage{multirow}
\usepackage{makecell}
\usepackage{colortbl}

\usepackage{newfloat}
\usepackage{listings}
\DeclareCaptionStyle{ruled}{labelfont=normalfont,labelsep=colon,strut=off} % DO NOT CHANGE THIS
\floatstyle{ruled}
\newfloat{listing}{tb}{lst}{}
\floatname{listing}{Listing}
\nocopyright

\title{Generalising Traffic Forecasting to Regions without Traffic Observations}
\author{
    Xinyu Su\textsuperscript{\rm 1,\rm2}\footnote{Work partially done after Xinyu Su joined The Hong Kong University of Science and Technology (Guangzhou).},
    Majid Sarvi\textsuperscript{\rm 1},
    Feng Liu\textsuperscript{\rm 1},
    Egemen Tanin\textsuperscript{\rm 1},
    Jianzhong Qi\textsuperscript{\rm 1}\footnote{Corresponding author.}
}
\affiliations{
    \textsuperscript{\rm 1}School of Computing and Information Systems, The University of Melbourne\\
    \textsuperscript{\rm 2}Artificial Intelligence Thrust, The Hong Kong University of Science and Technology (Guangzhou)\\
    \{suxs3@student., majid.sarvi@, feng.liu1@, etanin@, jianzhong.qi@\}unimelb.edu.au
}

\usepackage{bibentry}
\newcommand{\model}{{GenCast}}

\begin{document}

\maketitle

\begin{abstract}
Traffic forecasting is essential for intelligent transportation systems. Accurate forecasting relies on continuous observations collected by traffic sensors. However, due to high deployment and maintenance costs, not all regions are equipped with such sensors. This paper aims to  
forecast for regions without traffic sensors, where the lack of historical traffic observations challenges the generalisability of existing models. 
We propose a model named \textbf{\model}, the core idea of which is to exploit external knowledge to compensate for the missing observations and to enhance generalisation. We integrate physics-informed neural networks into \model, enabling physical principles to regularise the learning process. We introduce an external signal learning module to explore correlations between traffic states and external signals such as weather conditions, further improving model generalisability. Additionally, we design a spatial grouping module to filter localised features that hinder model generalisability. Extensive experiments show that \model\ consistently reduces forecasting errors on multiple real-world datasets.

\end{abstract}

\begin{links}
    \link{Code}{https://github.com/suzy0223/GenCast}
\end{links}

\section{Introduction}
Traffic forecasting is essential for intelligent transportation systems, enabling optimisations such as real-time route planning and transportation scheduling. Accurate traffic forecasting yields substantial social and economic benefits by improving travel efficiency, reducing congestion-related losses, and supporting sustainable urban development~\cite{gman,gamcn}. However, high deployment costs of traffic sensors often result in their sparse and limited spatial coverage~\cite{ignnk}, creating a gap between limited traffic observations and the need for fine-grained, wide-coverage forecasting. 

To bridge this gap, recent works study  {\textit{kriging and extrapolation}. Kriging models estimate current traffic conditions at locations of interest without sensors, i.e., \emph{unobserved locations}~\cite{increase,kits}, while extrapolation models take a step further and forecast for such locations~\cite{STGNP,stgp}. Although these models have produced promising results for scattered unobserved locations~(Fig.~\ref{fig:introduction}a), they struggle when applied to large continuous regions without traffic sensors~(Fig.~\ref{fig:introduction}b), i.e., traffic forecasting for \emph{unobserved regions}~\cite{stsm}. 

\begin{figure}[t]
     \centering
     \begin{subfigure}
         \centering
         \includegraphics[width=\columnwidth]{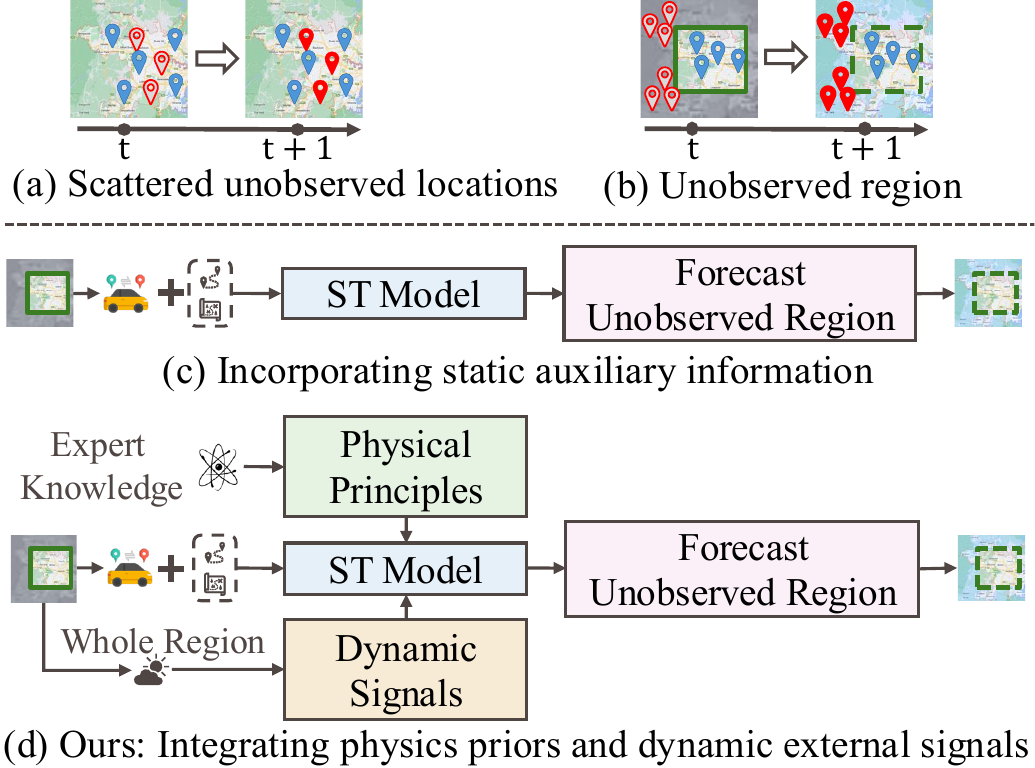}
     \end{subfigure}
     \caption{
      Illustration of the problem setting and modelling strategies.
\textbf{Upper}:
(a) Scattered unobserved locations vs. (b) Unobserved region (\textbf{our focus}). Blue bubbles represent observed locations; red hollow bubbles denote unobserved ones; and red solid bubbles denote the target forecasts. 
\textbf{Lower}: Comparison of different modelling strategies:
(c) Incorporating static auxiliary features (e.g., learning from locations with similar POIs or geo-coordinates);
(d) Integrates physics priors and dynamic external signals (e.g., weather) to improve generalisation~(\textbf{ours}).}
     \label{fig:introduction}
\end{figure}

We consider traffic forecasting for a region without traffic observations that is adjacent to (or enclosed by) observed regions. As Fig.~\ref{fig:introduction}b
illustrates, the region with red hollow bubbles does not have traffic observations at current time $t$, which is adjacent to the region with observations as denoted by the blue bubbles. We aim to forecast for the unobserved region for a future time (the red bubbles at time $t+1$). 
This setting is practical due to the staged deployment of sensors or unbalanced regional development~\cite{stsm}. 

The state-of-the-art (SOTA) model for this setting, STSM~\cite{stsm}, masks locations in observed regions that are similar to the unobserved region and is trained to forecast for such locations, with the aim to generalise to unobserved regions. The similarity is defined based on static auxiliary features, i.e., POI categories and geo-coordinates (Fig.~\ref{fig:introduction}c), which, however, do \emph{not} capture the dynamic nature of traffic patterns, thus limiting generalisation capacity.

Another work, KITS~\cite{kits}, creates virtual nodes at random locations and trains a model to forecast for such nodes, instead of masking the already-observed locations. It implicitly assumes scattered unobserved locations of a known density (the virtual nodes are created to match this density), thereby introducing a structural prior that limits generalisability to large continuous unobserved regions.

To address these limitations, we propose to exploit more versatile forms of guidance to enhance the generalisation of traffic forecasting models to regions without traffic observations. We explore two guidance signals: (1)~\emph{physical principles} that encode inherent traffic dynamics generalisable across regions; and (2)~\emph{dynamic external signals} available across regions that closely correlate with traffic patterns. These guidance signals offer a principled way to bridge the gap between observed and unobserved regions. Accordingly, we propose a traffic fore\emph{\underline{cast}}ing model \emph{\underline{gen}}eralisable to unobserved regions named \textbf{\model}.

To incorporate physical principles, we exploit 
the Lighthill--Whitham--Richards (LWR)  equation~\cite{lwr1,lwr2} as a soft learning constraint (i.e., a loss term). LWR governs the relationships between traffic density and flow in a road network.  
Using it poses two technical challenges: (1)~The LWR equation considers both traffic density and flow, while density data are typically unavailable. (2)~Applying LWR in model learning requires partial derivative over space, while most traffic forecasting models consider a graph over locations of interest which are  discrete and not directly differentiable.  

To address these challenges, we reformulate the LWR constraint based on traffic speed, yielding a physical constraint without requiring explicit density/flow data. We introduce two continuous spatial embeddings that enable automatic differentiation:
(i)~SE-L, a large language model (LLM)-based embedding that encodes semantic attributes (e.g., POIs and road structure); and
(ii)~SE-H, a GeoHash-based embedding that preserves spatial locality. 
These embeddings serve as differentiable proxies for locations on a graph, enabling residual-based physical loss computation.

To further utilise dynamic external signals, we exploit global weather observations from ECMWF~\cite{ecmwf_era5}, for their universal availability and strong correlation with dynamic traffic patterns~\cite{nigam2023hybrid}. We introduce an attention-based fusion module to learn the correlations between weather and traffic patterns. 

Beyond introducing external guidance signals, it is also crucial to filter patterns local to individual locations (e.g., induced by a traffic accident) that are non-generalisable to unobserved regions.
We propose a spatial grouping module that dynamically learns to group locations based on their intrinsic spatial-temporal patterns. The resulting groups enable \model\ to learn shared patterns of a group and suppress disruptive signals local to individual locations~\cite{lin2023diversifying}.

Overall, our contributions are summarised as follows:

(1)~We propose a traffic forecasting model, \model, that aims to generalise to regions without observations.

(2)~We design (i)~a physical principle-guided loss together with continuously differentiable spatial embeddings and (ii)~a cross-domain fusion module to fuse dynamic weather signals with traffic observations. These modules guide \model\ to generalise to regions without traffic observations through external knowledge and signals.  

(3)~We propose a spatial grouping module that filters out noisy localised signals while preserving region-invariant patterns to further strengthen model generalisability.%, thereby improving model robustness and generalisation.

(4)~We conduct extensive experiments on real-world datasets. The results show that \model\ consistently outperforms SOTA  baselines, reducing forecasting errors by up to 3.1\% and improving R\textsuperscript{2} scores by up to 125.6\%. %Ablation studies further validate the effectiveness of each component in our framework.

\section{Related Work}
Kriging and extrapolation in traffic forecasting aim to infer current or future observations at unobserved locations~\cite{STGNP,dualSTN,ustd,stgp}. However, as discussed in the introduction, existing methods often struggle to generalise, particularly when forecasting for regions without traffic observations. Next, we briefly review representative strategies to improve model generalisation.

Physics-guided approaches have been proposed to enhance generalisability of spatial-temporal models~\cite{hwang2021climate,hettigeairphynet,ClimODE}. In the context of traffic forecasting, these approaches fall into two categories: (1) simulating latent dynamics via neural ODEs or energy-based models~\cite{STDEN,st-pef}, which typically require fully observed traffic data to initialise hidden model states; and (2) imposing traffic flow constraints (e.g., LWR) via physics-informed neural networks (PINNs)~\cite{raissi2019physics,shi2021physics,zhang2024physics}, which often assume \emph{continuous} spatial locations. 
Our model introduces differentiable spatial embeddings to enable PINNs on traffic graphs which are discrete.

Other studies use external signals to enhance forecasting performance by capturing invariant spatial-temporal patterns from external-domain datasets~\cite{li2024opencity} or environmental features (e.g., weather~\cite{mystakidis2024traffic} or events~\cite{zhou2024sdformer,dualcast,ruan2025retrieval}). These studies use  such signals to help detect irregular events or handle short-term missing data. The use of such signals to guide model generalisation to unobserved regions, combined with advanced spatial-temporal graph networks, remains underexplored.
A full discussion of these works is included in Appendix~A.

\section{Preliminaries}
\paragraph{Region and Region Graph.} Following~\citet{stsm}, we represent a region as a graph $G=(V, E)$, where $V$ denotes $N$ locations of interest and $E$ represents their connections based on spatial proximity. Each location is associated with a feature vector, forming a matrix $\mathbf{L} \in \mathbb{R}^{N \times F}$ that encodes static attributes such as geo-coordinates, road network information, or regional descriptors. The specific features, which can be raw attributes or embeddings, may vary across different methods. For each $v_i \in V$, we use $\mathbf{x}_i$ to denote the series of traffic observations at location $v_i$ and $\mathbf{x}_i^t \in \mathbb{R}^C$ to denote the $C$ different types of observations (e.g., speed and volume) at time $t$, if there are such observations collected. 

\paragraph{Observed and Unobserved Regions.} We consider two disjoint but adjacent regions (i.e., graphs): an \emph{observed region} $G_o = (V_o, E_o)$ and an \emph{unobserved region} $G_u = (V_u, E_u)$, where $V_o \cap V_u = \emptyset$. 
We denote $N_o = |V_o|$ and $N_u = |V_u|$. Graphs $G_o$ and $G_u$ together form the input region graph $G = (V, E)$, where $V = V_o \cup V_u$, $N = N_o + N_u$ and $E_o + E_u \subseteq E$.

\paragraph{Weather Data.} We collect weather data from the ECMWF ERA5 dataset~\cite{ecmwf_era5}, provided in a gridded format with $9\,\text{km} \times 9\,\text{km}$ resolution. For each grid cell $i$ at time $t$, the weather observation is denoted as $\mathbf{x}_{w,i}^{t} \in \mathbb{R}^{C_w}$, where $C_w=4$ is the number of weather attributes (2-meter temperature, surface net solar radiation, surface runoff, and total precipitation). We denote the full weather observations at time $t$ as $\mathbf{X}_{w}^{t} \in \mathbb{R}^{N_w \times C_w}$, where $N_w$ is the number of grid cells overlapping the input region $G$.

\paragraph{Problem Statement.}
Given region $G = (V, E) = G_o \cup G_u$, location features $\mathbf{L}$ corresponding to $V$, traffic observations $\mathbf{X}_{G_o}^{t-T+1:t}$ for $G_o$ over the past $T$ time steps, and weather observations $\mathbf{X}_{w}^{t-T_{w}+1:t}$ for $G$ over the past $T_w$ time steps ($T$ and $T_w$ may vary and hence are denoted differently), we aim to learn a function $f$ to forecast the traffic conditions $\hat{\mathbf{X}}_{G_u}^{t+1:t+T'}$ for $G_u$ over the next $T'$ time steps:
\begin{equation}\label{eq:task_definition}
\hat{\mathbf{X}}_{G_u}^{t+1:t+T'}
= f(\mathbf{X}_{G_o}^{t-T+1:t}; \mathbf{X}^{t-T_{w}+1:t}_{w}; G; \mathbf{L}).
\end{equation}
We note that observed (or unobserved) regions (locations) refer to regions (locations) with (or without) \emph{traffic} observations. Both types of regions (locations) have weather observations from the ECMWF ERA5 dataset.

\begin{figure*}[!ht]
     \centering
     \begin{subfigure}
         \centering
         \includegraphics[width=0.88\textwidth]{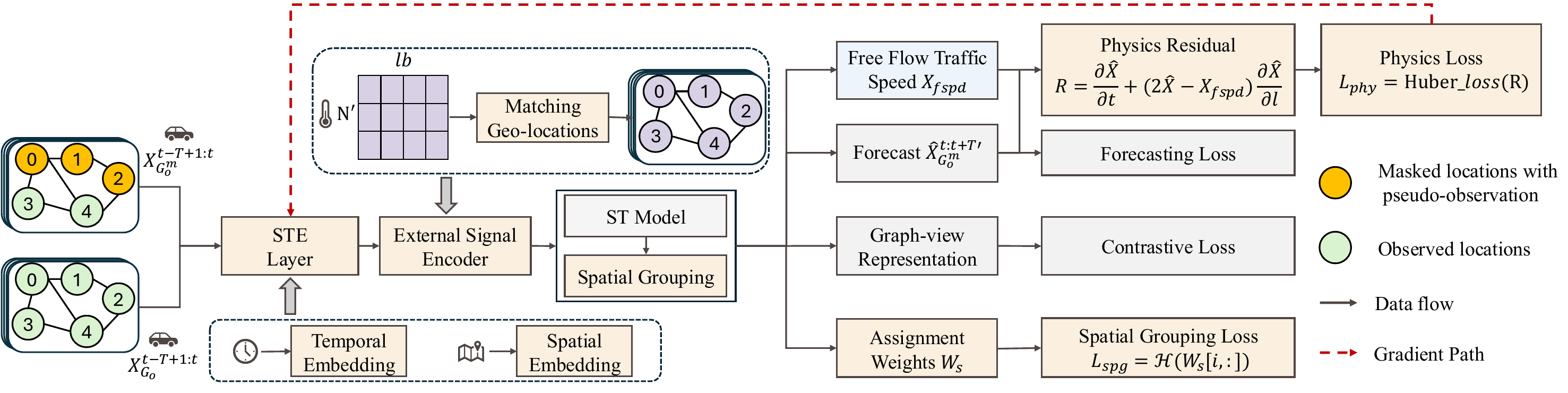}
     \end{subfigure}
     \caption{Overview of \model. Given observed traffic data $G_o$, a masked view $G_o^m$ is generated by randomly masking a subgraph. Temporal and spatial embeddings $\mathbf{TE}_{enc}$ and $\mathbf{L}_{enc}$ are fused with inputs $\mathbf{X}_{G_o^m}^{t-T+1:t}$  and $\mathbf{X}_{G_o}^{t-T+1:t}$ via a spatial-temporal embedding (STE) layer to produce initial features $\mathbf{H}^0_m$ and $\mathbf{H}^0$, respectively. External signals (weather) are matched to graph nodes via geo-coordinates and are integrated with the node features using an external signal encoder, yielding $\mathbf{H}^0_{m, fuse}$ and $\mathbf{H}^0_{fuse}$.These feature matrices are passed to a spatial-temporal (ST) model with a spatial grouping module, producing forecasts, graph representations, and learnable soft grouping scores for each node. Three loss terms are used: forecast loss $L_{pred}$ measures forecast errors;  contrastive loss $L_{cl}$ measures node representation consistency across $G^m_o$ and $G_o$; and 
     group-aware loss $L_{spg}$ measures group assignment confidence. The physics module computes a residual $R$ based on LWR to form a fourth (i.e., physics) loss $L_{phy}$, to regularise \model\ by physical principles of traffic dynamics.}
     \label{fig:framework}
\end{figure*}

\section{Methodology}
\subsection{Model Overview} 

Fig.~\ref{fig:framework} presents an overview of \model. The backbone (in \emph{gray}) adopts a contrastive architecture with a pseudo-observation generator, following~\cite{stsm}. Each epoch constructs a masked graph $G_o^m$ by randomly masking a subgraph of $G_o$, producing two input views: original $\mathbf{X}{G_o}^{t-T+1:t}$ and masked $\mathbf{X}{G_o^m}^{t-T+1:t}$.
Pseudo-observations are generated for masked locations to compute spatial (geo-coordinate) and temporal (traffic-series) similarity matrices, which feed GCNs and TCNs to capture respective dependencies.

The two views go through the GCNs and TCNs (i.e., a spatial-temporal (ST) model) separately to produce forecasts and graph representations $\mathbf{Z}^{t+T'}_{G_o}$ and $\mathbf{Z}^{t+T'}_{G_o^m}$, where the representation is taken from the final time step $t+T'$ following~\citet{whdocl}.
A contrastive loss, $L_{cl}$, is applied over $\mathbf{Z}^{t+T'}_{G_o}$ and $\mathbf{Z}^{t+T'}_{G_o^m}$ to promote consistent forecasts with and without masked (i.e., unobserved) locations, thereby achieving generalisation to the unobserved region. 
Since both views go through identical pipelines, we omit subscripts `$G_o$' and `$G^m_o$' in the subsequent discussion as long as the context is clear. More details about the backbone contrastive learning process are in Appendix~{B}.

\paragraph{Our Proposed Modules.}
We power \model\ with four modules to achieve high generalisability to unobserved regions: \textbf{spatial and temporal encoder}, \textbf{external signal encoder}, \textbf{spatial grouping module}, and \textbf{physics-informed module} ({coloured in \emph{almond}}). 

We design a \emph{spatial-temporal encoder} to embed node geo-locations and time indices into differentiable representations ($\mathbf{L}_{enc}$ and $\mathbf{TE}_{enc}$), enabling back-propagation for model optimisation guided by a physical principle-based loss. These embeddings are fused with the input $\mathbf{X}^{t-T+1:t}$ through an STE layer to produce the initial representation $\mathbf{H}^0 \in \mathbb{R}^{T\times N \times D}$, where $D$ is the hidden dimensionality.

To further utilise dynamic external signals, we match the nodes with weather data by their geo-locations. We denote the matched weather data by $\mathbf{X}^{t-T_w+1:t}_{wx}$. An \emph{external signal encoder} (i.e., a cross-attention module) fuses these signals with $\mathbf{H}^0$, producing an enriched representation $\mathbf{H}^0_{fuse}$.

Then, $\mathbf{H}^0_{fuse}$ is fed into the ST model as part of the contrastive learning process. 
To improve model generalisability, we encourage the learning of essential patterns by filtering out localised signals.  We introduce a spatial grouping module that softly assigns each node to a small number of spatial groups via learnable weights. To avoid group representations being impacted by ad hoc local features, we apply an entropy loss $L_{spg}$ to promote confident, near one-hot assignments. This allows each node to primarily contribute to one representative group, supporting clearer group-level representations and better generalisation to unobserved regions.

Besides outputting learned graph representations as mentioned earlier, the ST model also produces forecasts $\hat{\mathbf{X}}^{t+1:t+T'}$, which are fed into the \emph{physics-informed module} with an automatic differentiation step to compute spatial and temporal derivatives. A residual $R$ is computed based on LWR. The physics loss $L_{phy}$  minimises this residual to encourage confinement to physical laws of traffic dynamics.

\paragraph{Model Training.}
\model\ is optimised with a loss function of four terms, a forecast loss $L_{pred}$ that encourages accurate forecasts at the \emph{masked} locations, plus the contrastive loss $L_{cl}$, spatial grouping loss $L_{spg}$, and physics loss $L_{phy}$ as mentioned above. 
Note that, except for $L_{cl}$, all other losses are computed based on the masked graph $G_o^m$ to simulate unobserved locations and enhance generalisation.
{\small
\begin{equation}
    L = L_{pred} + \lambda L_{cl} + \mu L_{spg} + \theta L_{phy} 
\end{equation}}
Here, $\theta$ and $\mu$ are  hyper-parameters. The contrastive learning loss weighting follows~\citet{stsm}.

\paragraph{Model Inference.} For inference, we first compute pseudo-observations for the unobserved locations, and let graph $G = G_o + G_u$ with pseudo-observations (for $G_u$) be $G_m$. Then, the model forward process described above is run to generate forecasts for $G_m$, from which forecasts for the unobserved locations can be extracted.

Next, we detail the four proposed modules of \model.

\subsection{Spatial and Temporal Encoder}
We encode temporal and spatial features into differentiable embeddings to enable physics residual computation.

\paragraph{Temporal Embedding.}
We construct a time embedding $\mathbf{TE}$ that captures daily traffic cycles. Each observed time step is assigned a position index within the day ($\mathbf{TE}[i] = i \mod T_d$), where $T_d$ is the number of time intervals per day.
The time embedding $\mathbf{TE}$ is encoded using sinusoidal functions to produce a smooth and continuous representation: $\mathbf{TE}_{enc} = \left[ \sin\left(2\pi \cdot \frac{\mathbf{TE}}{T_d} \right), \cos\left(2\pi \cdot \frac{\mathbf{TE}}{T_d} \right) \right]$. 

\paragraph{Spatial Embedding.} 
We introduce two spatial feature encoding strategies:~(1)~\emph{LLM-based spatial embedding} ({SE-L}) and~(2)~\emph{GeoHash-based spatial embedding} ({SE-H}). They differ in their input features (and hence processing mechanisms) to suit different location feature availability settings. 

\textit{LLM-based spatial embedding.} SE-L is computed with two steps: (i)~location description generation and (ii)~embedding generation following~\cite{sun2025flexireg}. 
Location description generation constructs textual descriptions for each location by prompting an LLM with geo-coordinates, geometric properties of the surrounding area, POI information and attributes of the nearest road segments (see Appendix~{B}). 
The location description of $S$ tokens, $\mathbf{L}_t \in \mathbb{R}^{N_o \times S}$, is fed into a frozen LLaMA3 8B Instruct~\cite{llama2024}. The final token embedding from the last hidden layer of the model is SE-L,  $\mathbf{L}_{{llm}} \in \mathbb{R}^{N_o \times d_{{llm}}}$ ($d_{llm}$ is the embedding dimensionality).

\textit{GeoHash-based spatial embeddings.} When location features are unavailable beyond the geo-coordinates, we use GeoHash~\cite{geohash} to compute spatial embeddings SE-H. %This approach is versatile and does not rely on additional regional information as in SE-L. 
There are two main steps: 
(i)~GeoHash string generation and (ii)~embedding generation. 
 GeoHash string generation applies GeoHash coding on 
the geo-coordinates of a location to convert them into a fixed-length alphanumeric string, the length of which decides spatial precision. 
The embedding generation step then embeds the GeoHash string using a pre-trained character-BERT~\cite{lhy_charbert_2023}. The last hidden layer output, discarding the special tokens [CLS] and [SEP], %, gives contextualised character-level embedding, 
produces 
$\mathbf{L}'_{hash} \in \mathbb{R}^{N_o \times S \times d_{bert}}$. Here, $S$ is reused to denote the GeoHash string length, and $d_{{bert}}$ is the embedding dimensionality. 
To capture semantic dependencies within $\mathbf{L}'_{{hash}}$, we feed it into a multi-layer Transformer encoder, and we apply mean pooling along the character dimension of the output to obtain the SE-H, $\mathbf{L}_{{hash}}\in \mathbb{R}^{N_o \times d_{{hash}}}$.

We use $\mathbf{L}_{{enc}}$ to denote spatial embeddings ($\mathbf{L}_{{llm}}$ or $\mathbf{L}_{{hash}}$) when the context is clear.
We add $\mathbf{TE}_{enc}$ and $\mathbf{L}_{enc}$ to obtain $\mathbf{STE}$, and concatenate it with $\mathbf{X}^{t-T+1:t}$ as $\mathbf{H}^0$.

\subsection{External Signal Encoder}
For external weather signals, each traffic location $v_i$ is matched to its nearest weather station $s_j$ based on proximity:
{\small
\begin{equation}
\label{eq:match_weather}
\mathbf{X}_{{wx}, i} = \mathbf{X}_{w, j^*(i)}, 
\text{where} \ j^*(i) = \arg\min_{s_j \in \mathcal{S}_w}\text{dist}(v_i, s_j),
\end{equation}
}
where $\mathcal{S}_w$ denotes the set of all weather stations, and $\mathbf{X}_{{wx}, i}$ is the external signal assigned to location $v_i$.
After matching, we obtain an external signal tensor $\mathbf{X}_{{wx}} \in \mathbb{R}^{T_w \times N_o \times C_w}$, where $T_w$ is the weather window length  and $C_w$ is the number of weather features. 
Empirically, we associate each traffic observation with a 12-hour weather context in the past, i.e., $T_w{=}12$, to account for the lasting impact of weather. 

To capture traffic–weather interdependencies, we apply cross-attention by projecting $\mathbf{H}^0$ into queries $\mathbf{Q}$, and $\mathbf{X}_{{wx}}$ into keys $\mathbf{K}$ and values $\mathbf{V}$. Temporal attention scores $\alpha_{t,t'}$ are used to aggregate weather signals: $\mathbf{H}^{t}_{{wx}} = \sum_{t'=1}^{T_w} \alpha_{t,t'} \mathbf{V}^{t'}$, yielding $\mathbf{H}_{{wx}} \in \mathbb{R}^{T\times N_o \times D}$. Recall that $D$ is the hidden dimensionality.

We use gated fusion to fuse weather and traffic signals:
{\small
\begin{equation}
\begin{aligned}
\mathbf{H}^0_{{fuse}} &= \text{ReLU} \left( \text{FC}_h \left( z \odot \mathbf{H}^0 + (1 - z) \odot \mathbf{H}_{wx} \right) \right),\\
z &= \sigma\left(\text{FC}_s(\mathbf{H}^0) + \text{FC}_t(\mathbf{H}_{{wx}}) \right),
\end{aligned}
\end{equation}}
where $\text{FC}$ denotes linear layers, $\sigma(\cdot)$ is the sigmoid function, and $\odot$ denotes element-wise multiplication.
The fused output $\mathbf{H}^0_{{fuse}}$ is then passed through the ST model to generate node-level forecasts $\hat{\mathbf{X}}_{G_o^m}^{t+1:t+T'}$ and graph representations $\mathbf{Z}^{t+T'}_{G_o}$ and $\mathbf{Z}^{t+T'}_{G_o^m}$ for loss computation.

\subsection{Spatial Grouping Module}
We adopt spatial grouping to softly cluster locations into latent groups (Fig.~\ref{fig:spatial_grouping}), enabling \model\ to capture shared group-level patterns and filter out ad hoc patterns at individual locations, with the help of an entropy regularisation term.

We add a spatial grouping module to each layer of the ST model. For the output feature map $\mathbf{H}^{l} \in \mathbb{R}^{N_o\times T\times D}$ from the $l$-th layer, we first perform temporal average pooling to obtain a static spatial representation $\mathbf{H'}^{l} \in \mathbb{R}^{N_o \times D}$. 
We then divide the $D$ channels into $cg$ (a hyperparameter) channel groups and reshape the representation into $\mathbf{Z}^{l} \in \mathbb{R}^{(N_o \cdot cg) \times d'}$, where $d' = D/cg$, producing $N_o \times cg$ samples. This transformation enables the model to capture fine-grained features across channel partitions. 

We also project $\mathbf{H'}^{l} \in \mathbb{R}^{N_o \times D}$ to obtain a learnable $\mathbf{W}_c \in \mathbb{R}^{N_o \times (sg\cdot cg \cdot cg)}$, where $sg$ is the number of spatial groups (a hyperparameter). Then, we reshap it to obtain $\mathbf{W}_c \in \mathbb{R}^{(sg\cdot cg) \times (N_o \cdot cg)}$, and we map $\mathbf{W}_c$ to obtain the representations of group centres as: $\mathbf{C} =  \mathrm{Softmax}(\mathbf{W}_c)\mathbf{Z}^{l} \in \mathbb{R}^{(sg \cdot cg) \times d'}$, i.e., the representation of each group centre is determined by all input samples. 

The group assignment score is computed via a distance-based softmax, where $\text{cdist}(\cdot, \cdot)$ computes the pairwise Euclidean distances between all location samples and centres:
{\small
\begin{equation}
\mathbf{W}_s = \text{Softmax}\left(-\text{cdist}(\mathbf{Z}^l, \mathbf{C})\right) \in \mathbb{R}^{(N_o\cdot cg)\times (sg\cdot cg)}.
\end{equation}
}

\model\ encourages each sample to be confidently assigned to a representative group, suppressing the influence of ad hoc features at individual locations that introduce inconsistent signals and obscure generalisable group-level patterns.
To this end, we apply an entropy minimisation loss on the soft assignment weights:
{\small
\begin{equation}
\mathcal{L}_{{spg}} = \frac{1}{N \cdot c_g} \sum_{i=1}^{N \cdot c_g} \mathcal{H}(\mathbf{W}_s[i, :]), 
\end{equation}
}
where $\mathbf{W}_s[i, :]$ denotes the soft assignment weights of the $i$-th sample to all $sg \cdot cg$ latent groups, forming a probability distribution over group assignments. $\mathcal{H}(\mathbf{p}) = - \sum_j p_j \log (p_j + \epsilon)$, where $p_j$ denotes the soft assignment probability of a sample to the $j$-th latent group. Lower entropy will encourage sharper group membership. 

\subsection{Physics-informed Module}
\label{subsec:Physics-informed-Constraint}
The physics-informed module introduces a constraint with the LWR  model~\cite{lwr1,lwr2}. LWR describes the evolution of traffic density over space (location $l$) and time ($t$) using a conservation law formulated as Eq.~\ref{eq:lwr_basic}, where \(\rho = \rho(l, t)\) denotes traffic density, and\(x = x(l, t)\) represents traffic velocity (i.e., speed). 
{\small
\begin{equation}
\label{eq:lwr_basic}
    \frac{\partial \rho}{\partial t} + \frac{\partial (\rho x)}{\partial l} = 0.
\end{equation}
}

As traffic density observations are \emph{not} commonly available,  we rewrite the equation using velocity, assuming a closed system with a functional density-velocity relationship~\cite{greenshields,fspd}:
$x = x_{fspd} \left(1 - \frac{\rho}{\rho_{{max}}} \right)$, where $x_{fpsd}$ denotes the free flow speed -- the speed at which vehicles travel under low traffic density, and $\rho_{max}$ denotes the maximum traffic density, where vehicles are fully packed. This can also be written as: 
%We follow previous work~\cite{fspd} to compute it. 
{\small
\begin{equation}
\rho = \rho_{{max}} \left(1 - \frac{x}{x_{fspd}} \right)
\end{equation}
}
Putting it into Eq.~\ref{eq:lwr_basic} yields:
{\small
\begin{align}
\label{eq:phsics_speed_only}
-\frac{\rho_{{max}}}{x_{fspd}} \frac{\partial x}{\partial t}
+ \rho_{{max}} \left(1 - \frac{2x}{x_{fspd}} \right) \frac{\partial x}{\partial l} = 0
\end{align}}
Multiplying both sides by $-\frac{x_{fspd}}{\rho_{{max}}}$ gives: $\frac{\partial x}{\partial t} + (2x - x_{fspd}) \frac{\partial x}{\partial l} = 0$. The left-hand side is the physics residual~$R$: 
{\small
\begin{equation}
\label{eq:phy_residual}
R=\frac{\partial x}{\partial t} + (2x - x_{fspd}) \frac{\partial x}{\partial l}
\end{equation}
}

Further details on the derivation of the physics residual are in Appendix~{B}. 

As the system is not assumed to be closed, the model enforces local flow conservation across connected segments consistent with LWR principles. The Huber loss~\cite{huber1992robust} penalises violations of flow conservation to enforce these physical constraints
{\small
\begin{equation}
\label{eq:phy_loss}
\mathcal{L}_{{phy}} =\mathrm{Huber}(R,\delta).
\end{equation}
}
Here, $\delta$ denotes the Huber threshold. To account for dataset-specific error scales, we adopt an adaptive strategy. After a warm-up run (an epoch using the RMSE loss), we compute the $\tau$-quantile of the physical residuals to set $\delta$, where $\tau$ is a tunable hyperparameter. Notably, when $\tau = 100\%$, the Huber loss reduces to RMSE. The model is then re-initialised and trained using the Huber loss with this fixed $\delta$.

During model training, Eq.~\ref{eq:phy_residual} becomes:
{\small
\begin{equation}
\label{eq:phy_loss_emb}
R =  \frac{\partial \mathbf{\hat{X}}}{\partial \mathbf{TE}_{{enc}}} + \left(2\mathbf{\hat{X}} - \mathbf{X}_{{fspd}} \right) \cdot \frac{\partial \mathbf{\hat{X}}}{\partial \mathbf{L}_{{enc}}},
\end{equation}}
where $\hat{\mathbf{X}} = \hat{\mathbf{X}}_{G_o^m}^{t+1:t+T'}$ denotes model forecasts based on the masked graph, and $\mathbf{X}_{{fspd}} \in \mathbb{R}^{N \times 1}$ represents estimated free-flow speed at each location~\cite{fspd}.
The partial derivatives with respect to $\mathbf{TE}_{{enc}}$ and $\mathbf{L}_{{enc}}$ are computed via automatic differentiation. 

\section{Experiments}
\subsection{Experimental Setup}
\paragraph{Datasets.} 
We evaluate \model\ on four highway datasets (PEMS-Bay, PEMS07, PEMS08, and METR-LA) and one urban dataset (Melbourne), with 5-min or 15-min intervals. Dataset details, including statistics, processing, and visualisations, are in Appendix~C.1. We use \textbf{ERA5-Land weather data}~\cite{ecmwf_era5} (hourly, 9km$\times$9km) with four traffic-related variables: temperature, solar radiation, precipitation, and runoff. \textbf{Region and road network data} for LLM-based embeddings are from~\citet{osm}.

Following prior work~\cite{increase,stsm}, we use traffic records from the past two hours to forecast for the next two hours, i.e., $T = T' = 2$ hours. Each dataset is split into training, validation, and test sets in a 4:1:5 spatial ratio, ensuring spatial adjacency within each split. Locations in the training and validation sets are treated as observed, while test locations are unobserved. The space-based split is performed horizontally or vertically based on geo-coordinates. We generate four spatial splits per dataset and report results on average. Temporally, the first 70\% of data is used for training, and the remaining 30\% for testing.

\paragraph{Competitors.} We compare with a transductive Kriging model \textbf{GE-GAN}~\cite{ge-gan}, inductive Kriging models \textbf{IGNNK}~\cite{ignnk},  \textbf{INCREASE}~\cite{increase} and \textbf{KITS}~\cite{kits}, and the SOTA model for unobserved region forecasting \textbf{STSM}~\cite{stsm}.

\paragraph{Implementation Details.} We use the default settings of all baselines. For imputation-based models, we adapt their objective to forecast future values. 
Our model is trained with Adam (initial learning rate: 0.01), batch size 32, and masking ratio $\sigma_m=0.5$, with hyperparameters selected on the validation set. All experiments are run on an NVIDIA A100 (80GB) GPU. We report RMSE, MAE, MAPE, and R\textsuperscript{2} -- the first three measure errors and R\textsuperscript{2} reflects improvement over historical averages. More details are in Appendix~C.1.

\begin{table*}[t]
\centering
{\small
\begin{tabular}{c|l|ccccc|cc|>{\columncolor{gray!15}}c}
\hlineB{3}
\multicolumn{1}{l|}{\textbf{Dataset}} & Metric & GE-GAN      & IGNNK  & INCREASE & STSM        & KITS        & \model-H & \model-L & Improve  \\ \hline \hline
\multirow{4}{*}{\textbf{PEMS07}}    & RMSE$\downarrow$   & 20.772      & 11.398 & 8.399    & \underline{8.390} & 9.574       & 8.285       & \textbf{8.253}   & 1.64\%        \\
                            & MAE$\downarrow$    & 15.436      & 9.016  & 5.396    & \underline{5.111} & 5.150       & 5.116       & \textbf{5.073}   & 0.74\%        \\
                            & MAPE$\downarrow$   & 0.270       & 0.179  & 0.124    & \underline{0.123} & 0.135       & 0.122       & \textbf{0.121}   & 1.63\%        \\
                            & R\textsuperscript{2}$\uparrow$     & -4.174      & -0.618 & 0.168    & \underline{0.169} & 0.094       & 0.193       & \textbf{0.197}   & 16.57\%       \\ \hline
\multirow{4}{*}{\textbf{PEMS08}}    & RMSE$\downarrow$   & 23.405      & 10.646 & 8.375    & \underline{7.925} & 8.182       & 7.880       & \textbf{7.863}   & 0.79\%        \\
                            & MAE$\downarrow$    & 17.613      & 8.138  & 5.097    & 4.899 & \underline{4.863}       & 4.776       & \textbf{4.728}   & 2.78\%        \\
                            & MAPE$\downarrow$   & 0.298       & 0.160  & 0.118    & \underline{0.114} & 0.115       & 0.113       & \textbf{0.112}   & 1.75\%        \\
                            & R\textsuperscript{2}$\uparrow$     & -6.531      & -0.642 & 0.031    & \underline{0.136} & 0.083       & 0.146       & \textbf{0.150}   & 10.51\%       \\ \hline
\multirow{4}{*}{\textbf{PEMS-Bay}}   & RMSE$\downarrow$   & 25.801      & 10.051 & 8.860    & \underline{8.773} & 9.435       & \textbf{8.683}       & 8.692   & 1.03\%        \\
                            & MAE$\downarrow$    & 24.822      & 6.596  & 5.339    & 5.390       & \underline{5.270} & 5.192       & \textbf{5.139}   & 2.49\%        \\
                            & MAPE$\downarrow$   & 0.407       & 0.160  & 0.134    & \underline{0.134} & 0.138       & \textbf{0.131}       & \textbf{0.131}   & 2.10\%        \\
                            & R\textsuperscript{2}$\uparrow$     & -5.856      & 0.042  & 0.196    & \underline{0.210} & 0.094       & \textbf{0.228}       & 0.225   & 8.43\%\\    \hline
\multirow{4}{*}{\textbf{METR-LA}}    & RMSE$\downarrow$   & 32.303      & 14.825 & {13.151}   & \underline{12.952}      & 13.916     & \textbf{12.720}      &    12.886     &    1.79\%     \\
                            & MAE$\downarrow$    & 26.371      & 12.119 & {9.062}    & 9.010       &    \underline{8.910}    & \textbf{8.792} &    8.799     &    1.33\%    \\
                            & MAPE$\downarrow$   & 0.507       & 0.311  & 0.272    & \underline{0.270}       & 0.293       & \textbf{0.265}       &     0.267    & 1.85\%         \\
                            & R\textsuperscript{2}$\uparrow$     & -4.901      & -0.258 & 0.025    & \underline{0.048}       & -0.086       &  \textbf{0.086}    &     0.063    &    79.58\%     \\ \hline
\multirow{4}{*}{\textbf{Melbourne}}  & RMSE$\downarrow$   & 10.233      & 14.262 & 9.579    & \underline{9.175} & 10.026      & \textbf{9.009}       & 9.258   & 1.81\%        \\
                            & MAE$\downarrow$    & 7.891       & 12.296 & 7.627    & \underline{7.308} & 7.971       & \textbf{7.083}       & 7.253   & 3.08\%        \\
                            & MAPE$\downarrow$   & \underline{0.374} & 0.939  & 0.408    & 0.388       & 0.415       & \textbf{0.366}       & 0.370   & 2.27\%        \\
                            & R\textsuperscript{2}$\uparrow$     & -0.213      & -1.810 & -0.042   & \underline{0.027} & -0.165      & \textbf{0.061}       & 0.012   & 125.56\%      \\
  
\hlineB{3}
\end{tabular}
}

\caption{Overall model performance.  ``$\downarrow$''/``$\uparrow$'' indicates lower/larger is better. The best baseline results are \underline{underlined}, and the best results are in \textbf{bold}. ``Improve'' means the errors reduced by \model-L compared with the best baseline model.}
\label{tab:overall_result}
\end{table*}

\subsection{Results}
\paragraph{Overall Results.} 
Table~\ref{tab:overall_result} reports forecast errors averaged over two hours (24 time steps for highway datasets, and 8 time steps for urban traffic datasets). Our model, with either variant \model-H (using GeoHash embeddings) or \model-L (using LLM embeddings), consistently outperforms all competitors. Our model reduces forecast errors by up to 3.1\% and increases $\mathrm{R}^2$ by up to 125.6\% on the Melbourne dataset. We further conducted paired t-tests and Wilcoxon signed-rank tests between our model and the best baseline across all datasets. The results show that our model consistently outperforms the baselines with statistically significant improvements $p\ll 10^{-8}$.

STSM, the SOTA model, is the best baseline in most cases (16 out of 20). GE-GAN and IGNNK underperform due to limited information flow. GE-GAN relies on static similarity, while IGNNK struggles with message propagation path construction. INCREASE iteratively masks and reconstructs nodes using partial observations and static similarities, overlooking dynamic dependencies. KITS creates virtual nodes inside the observed region. This setup limits generalisability to outside, unobserved regions, which is our setting.

\paragraph{Comparison between Spatial Embedding Strategies.} As shown in Table~\ref{tab:overall_result}, \model-L achieves better performance on PEMS07 and PEMS08, while \model-H performs better on PEMS-Bay, METR-LA, and Melbourne.
We attribute this discrepancy to the quality of SE-L. The data for PEMS-Bay and METR-LA were collected a long time ago (see Table~\ref{tab:dataset_info} in Appendix~C.1), whereas the regional information used for SE-L is  retrieved from the latest OpenStreetMap data. Changes in the physical environment over time may result in a mismatch between the generated embeddings and the conditions at the data collection time. In addition, the Melbourne dataset covers a small, homogeneous area concentrated in Melbourne CBD, making it difficult for SE-L to capture distinctive spatial features or meaningful propagation patterns. In contrast, SE-H is generated from geo-coordinates and is updated during training, making it more adaptable 
 and robust to different environments.

\paragraph{Ablation Study.}
We compare \model\ with five variants: \textbf{w/o-phy}, \textbf{w/o-spg}, and \textbf{w/o-wx}  remove the physics constraint, spatial grouping loss, and cross-domain encoder (i.e., weather), respectively; \textbf{w/o-SE} and \textbf{w/o-TE} remove spatial and temporal embeddings, respectively, together with the physics constraint.
As Fig.~\ref{fig:ablation} shows, all variants lead to higher errors, confirming the effectiveness of the modules. 

For \model-L, the physics constraint is particularly important, as the frozen spatial embeddings are less effective without additional guidance. This is evidenced by the high errors of {w/o-TE}, where only spatial embeddings are used. In contrast, \model-H uses simpler, trainable spatial embeddings, which function without physical constraints. 

In contrast, {w/o-spg} has similar impact across datasets, showing its generalise applicability.
We further compare the entropy of spatial grouping weights $\mathbf{W}_s$ between \model-L and its w/o-spg variant across all splits on PEMS07 (Fig.~\ref{fig:sgp_entropy}). \model-L has sharp peaks near zero entropy, suggesting confident and sparse (i.e., close to one-hot) assignments. The variant {w/o-spg} has more flat distributions and higher entropy, reflecting more diffuse and ambiguous groupings. These results confirm that our spatial grouping module effectively encourages confident group selection, filtering out localised features that could otherwise harm generalisation. 

\begin{figure}[!t]
    \centering
    \subfigure[PEMS07]{\includegraphics[width=0.48\columnwidth]{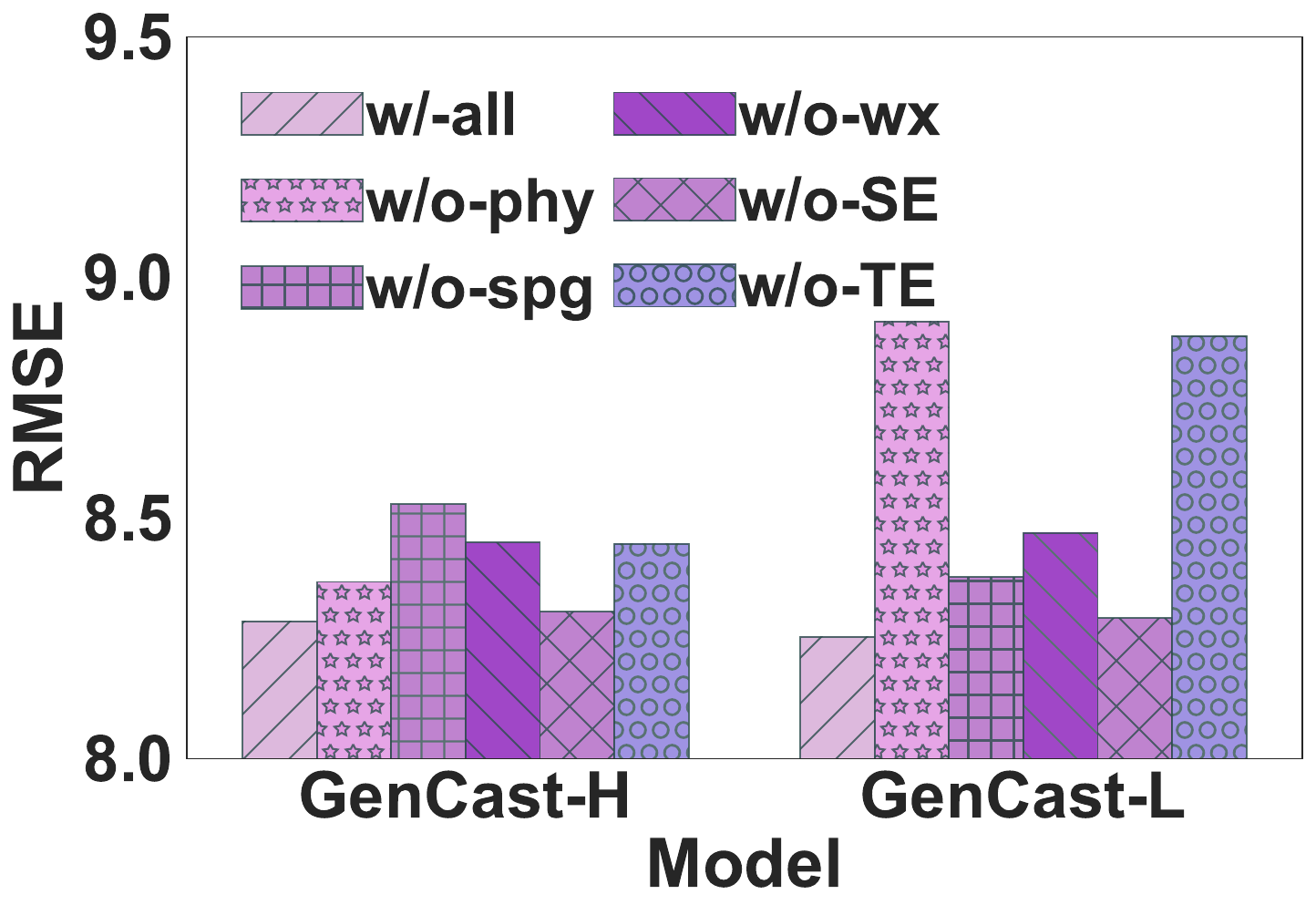}} 
    \hspace{-2mm}
    \subfigure[PEMS-Bay]{\includegraphics[width=0.48\columnwidth]{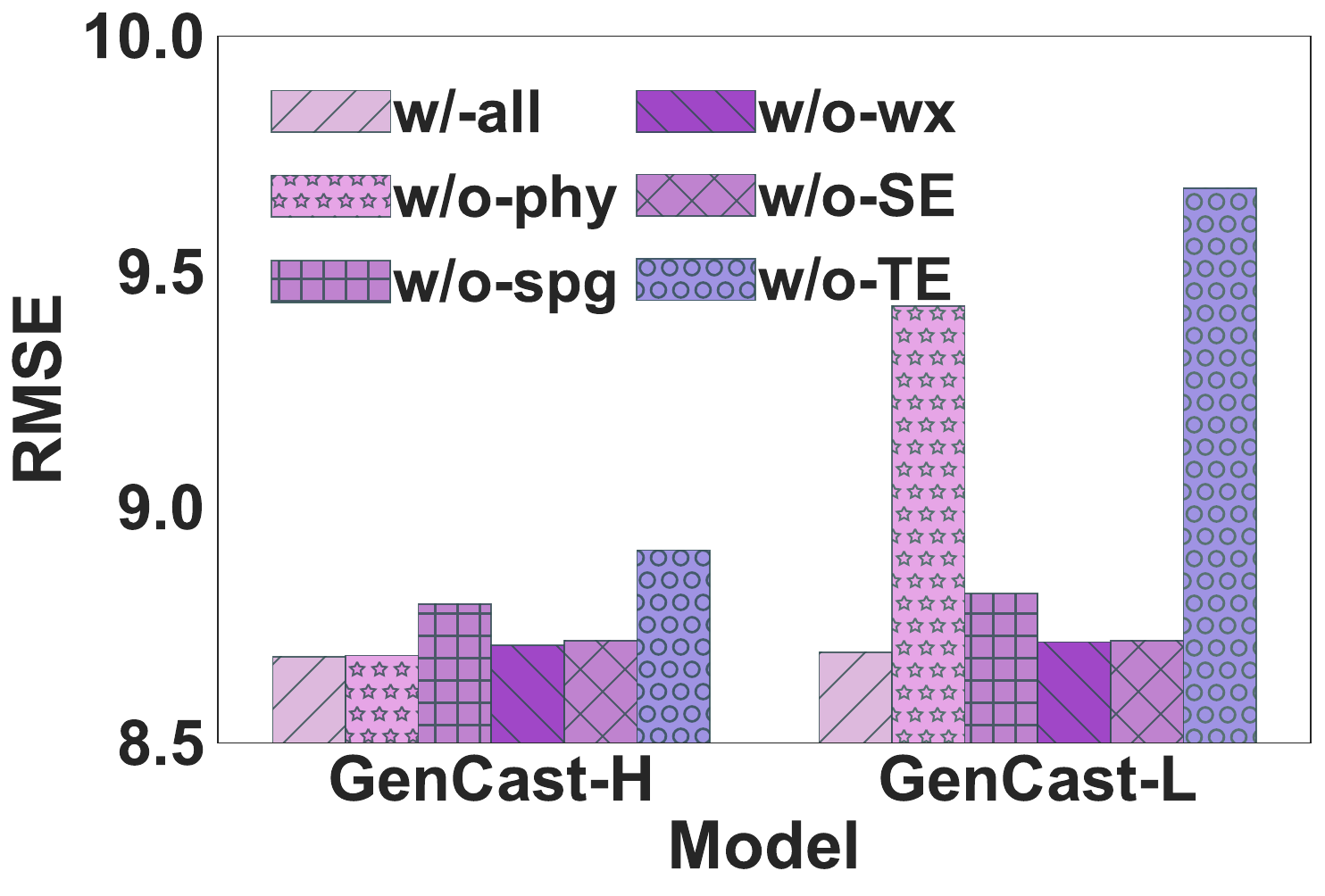}} 
    \caption{Ablation study results. We include results on the other datasets in the appendix. Same below.}
    \label{fig:ablation}
\end{figure}

\begin{figure}[!t]
    \centering
    \subfigure[$\mathbf{W}_s$ of Layer 1]{\includegraphics[width=0.47\columnwidth]{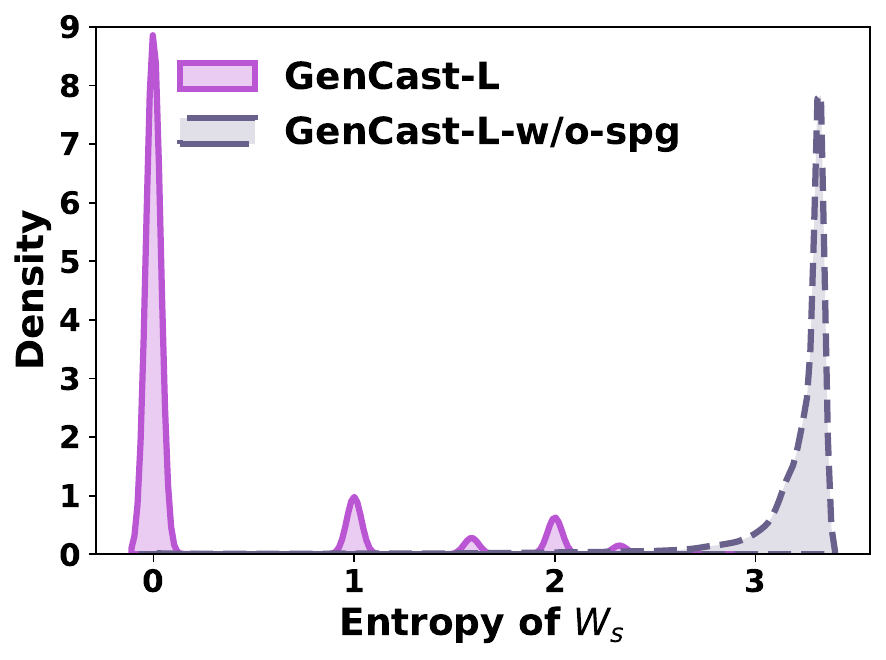}} 
    \hspace{-2mm}
    \subfigure[$\mathbf{W}_s$ of Layer 2]{\includegraphics[width=0.47\columnwidth]{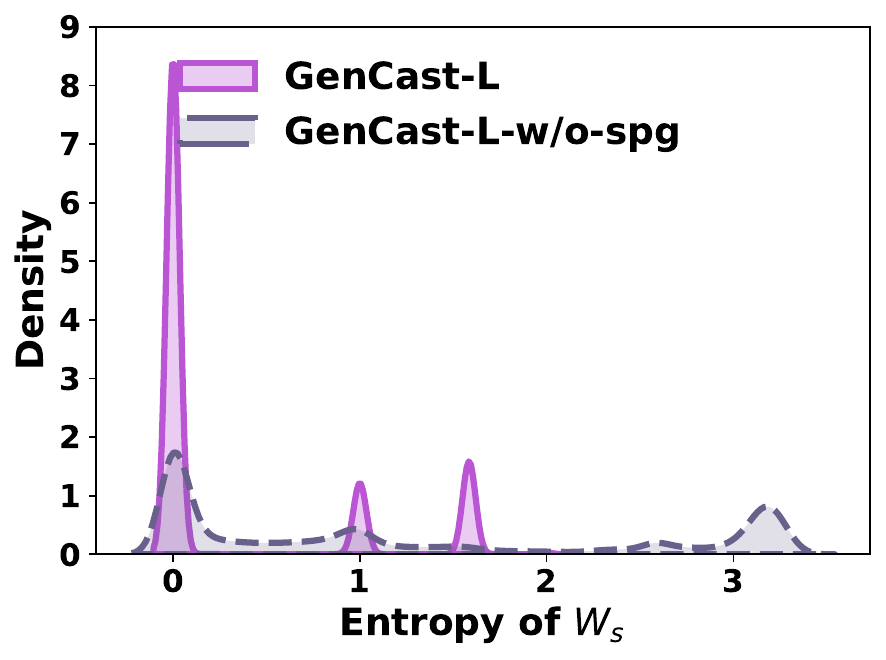}}
    \caption{SPG entropy: \model-L vs. {w/o-spg} (PEMS07).}
    \label{fig:sgp_entropy}
\end{figure}

\begin{figure}[!t]
    \centering
    \subfigure[PEMS07]{\includegraphics[width=0.48\columnwidth]{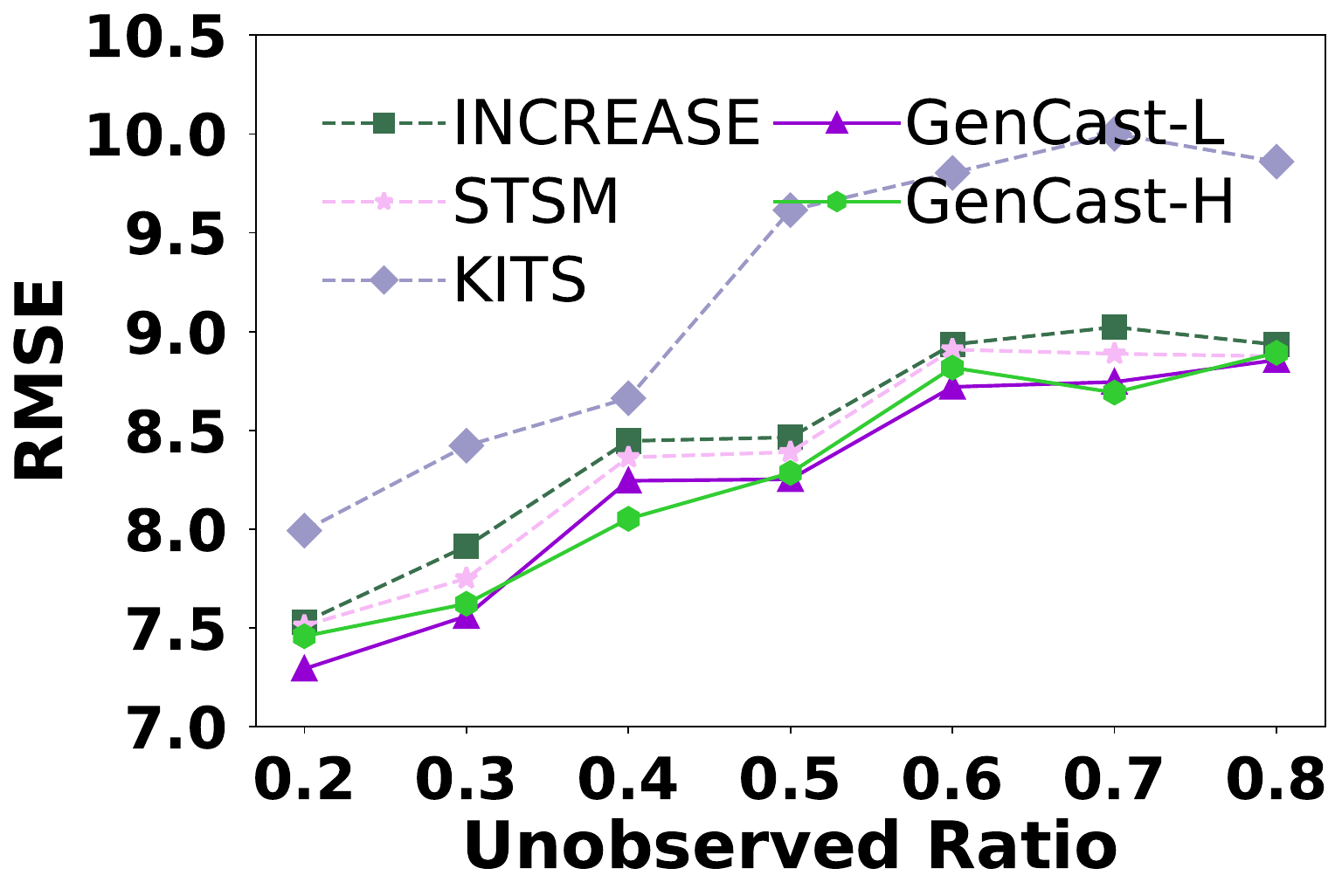}} 
    \hspace{-2mm}
    \subfigure[PEMS-Bay]{\includegraphics[width=0.48\columnwidth]{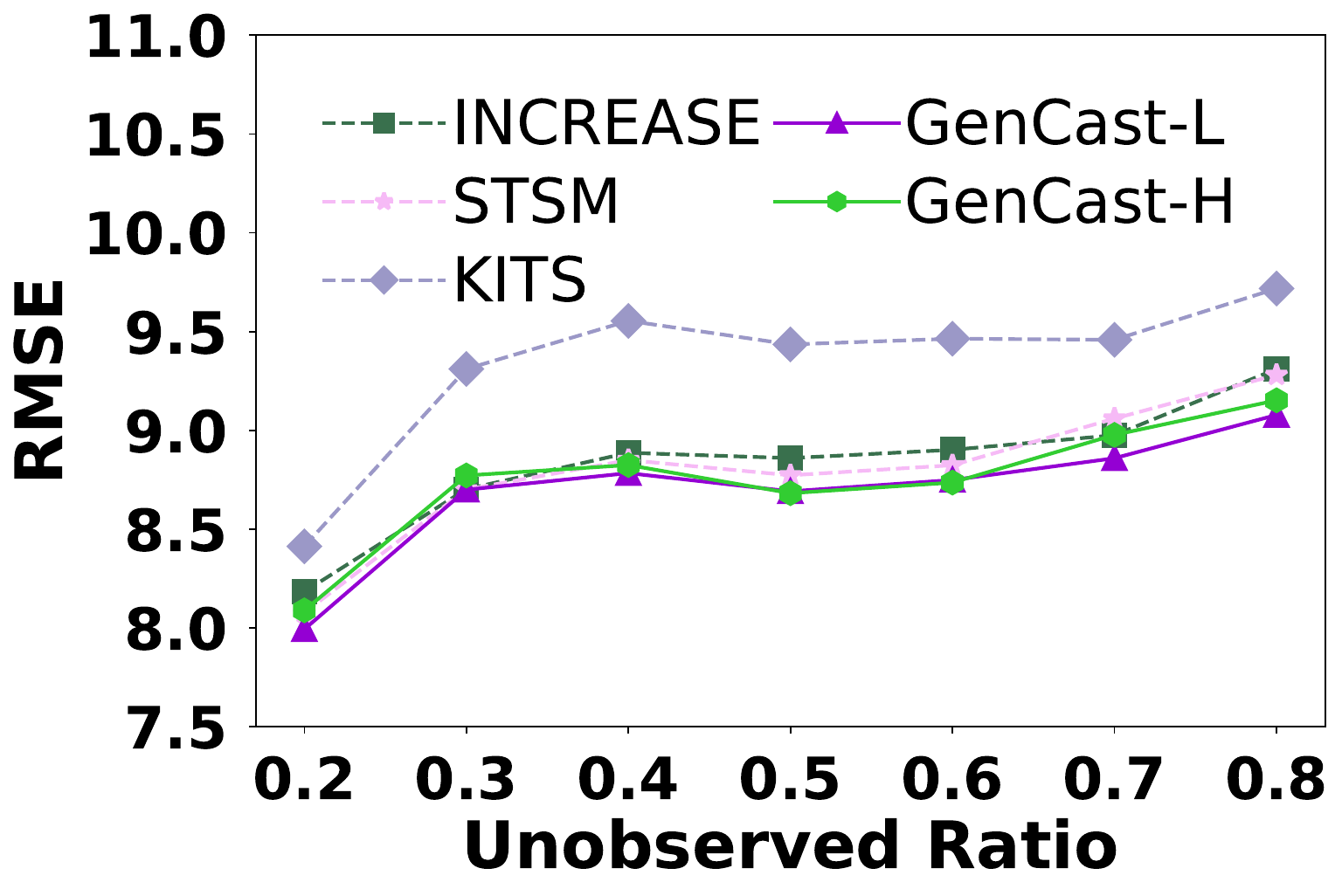}} 
    \caption{Model performance vs. unobserved ratio.}
    \label{fig:unknow_ratio}
\end{figure}

\begin{table}[t]
\centering

{\small
\begin{tabular}{l|cccc}
\hlineB{3}
 \multirow{2}{*}{\textbf{Model}}&\multicolumn{4}{c}{{PEMS-Bay (Ring Split)}} \\ 
 \cline{2-5}
 & \textbf{RMSE}$\downarrow$ & \textbf{MAE}$\downarrow$ & \textbf{MAPE}$\downarrow$ & \textbf{R$^2$}$\uparrow$ \\
\hline
GE-GAN     & 26.073 & 25.147 & 0.411 & -6.395 \\
IGNNK      & 12.881 & 10.056 & 0.198 & -0.808 \\
INCREASE   & 8.662  & 5.126  & \underline{0.126} & 0.178 \\
STSM       & \underline{8.599}  & 5.052  & 0.129 & \underline{0.189} \\
KITS       & 9.087  & \underline{4.942}  & 0.134 & 0.098 \\ \hline
\model-H    & \textbf{8.323}  & 4.929  & \textbf{0.123} & \textbf{0.241} \\
\model-L    & 8.583  & \textbf{4.734}  & 0.125 & 0.191 \\
\hline
\rowcolor{gray!15}
{Improve} & 3.21\% & 4.21\% & 2.38\% & 27.51\% \\
\hline \hline
\multirow{2}{*}{\textbf{Model}}&\multicolumn{4}{c}{{NREL}} \\ 
 \cline{2-5}
 & \textbf{RMSE}$\downarrow$ & \textbf{MAE}$\downarrow$ & \textbf{MAPE}$\downarrow$ & \textbf{R$^2$}$\uparrow$ \\
\hline 
GE-GAN        & 12.142 & 9.169  & 7.444 & -0.358 \\
IGNNK         & 13.732 & 10.444 & 2.409 & -0.697 \\
INCREASE      & 8.534  & 6.045  & 2.730 & 0.177 \\
STSM          & \underline{7.733}  & \underline{5.050}  & 1.789 & \underline{0.326} \\
KITS          & 8.704  & 5.513  & \underline{1.776} & 0.314 \\ \hline
\model-w/wx & \textbf{7.620}  & \textbf{4.770}  & \textbf{1.750} & \textbf{0.345} \\ \hline
\rowcolor{gray!15}
{Improve} & 1.46\% & 5.54\% & 1.46\% & 5.83\% \\
\hlineB{3}
\end{tabular}
}
\caption{Results on PEMS-Bay (Ring Split) and NREL.}
\label{tab:case_study}
\end{table}

\paragraph{Parameter Study.} 
We study the impact of key hyperparameters, including the number of spatial/channel groups in the grouping module, loss weights $\theta$, $\mu$, $\tau$, and weather window length $T_w$. The results (see Appendix~C) show that \model\ performs well under consistent settings across datasets, i.e., \model\ does \emph{not} need heavy tuning.

\paragraph{Impact of Unobserved Ratio.} We vary the ratio of unobserved nodes in $G$ from 0.2 to 0.8 on all datasets. Each dataset is split horizontally or vertically, and results are averaged over four setups. Fig.~\ref{fig:unknow_ratio} compares the top three baselines with our model. \model\ consistently performs the best, confirming its robustness against the unobserved ratio.    

\paragraph{Impact of Space Splits.} The relative position of the observed and unobserved regions may also impact model performance. We test model robustness with a ring split on PEMS-Bay, reflecting city layouts. Experiments show that \model\ again consistently outperform all baseline models, achieving up to a 27.5\% improvement in R$^2$ (Table~\ref{tab:case_study}).

\paragraph{Domain Generalisability of \model.} 
We further adapt \model\ to the solar power NREL dataset~\cite{nrel}, using \model-w/wx (i.e., without spatial embeddings or physics constraints) due to domain complexity. As shown in Table~\ref{tab:case_study}, it still outperforms all baselines, demonstrating strong generalisability.

\paragraph{Additional Results.} Appendix~C presents further results on the other datasets, model running time, and generalisability of the external signal encoder.

\section{Conclusion}
We proposed a model named \model\ to address the challenges in traffic forecasting for unobserved regions. 
Unlike purely data-driven approaches, \model\ uses physics knowledge and external spatial-temporal data (e.g., weather) to enhance generalisation to unobserved regions. It employs two continuously differentiable spatial embeddings to support physics-informed learning.
A spatial grouping module filters localised, non-transferable features.
We evaluated \model\ on real-world traffic datasets. The results show that it consistently outperforms SOTA models across different settings, achieving up to 3.1\% reduction in forecast error and up to 125.6\% improvement in $\mathrm{R}^2$.
Although \model\ achieves strong generalisation across regions, the relatively low $\mathrm{R}^2$ indicates room for improvement. Future work will focus on balancing cross-region generalisability and local fidelity through adaptive or hybrid modelling strategies.

\section{Ethical Statement}
All datasets used in this study are publicly available and do not contain any personally identifiable information. There are no ethical concerns associated with this work.

\section{Acknowledgments}
This work is in part supported by the Australian Research Council (ARC) via Discovery Projects DP230101534 and DP240101006. Jianzhong Qi is supported by ARC Future Fellowship FT240100170. Feng Liu is supported by the ARC with grant numbers DE240101089, LP240100101, DP230101540 and the NSF\&CSIRO Responsible AI program with grant number 2303037.

\bibliography{aaai2026}

@inproceedings{ignnk,
  title={{Inductive Graph Neural Networks for Spatiotemporal Kriging}},
  author={Wu, Yuankai and Zhuang, Dingyi and Labbe, Aurelie and Sun, Lijun},
  booktitle={AAAI},
  pages={4478--4485},
  year={2021}
}

@inproceedings{increase,
  title={{INCREASE: Inductive Graph Representation Learning for Spatio-Temporal Kriging}},
  author={Zheng, Chuanpan and Fan, Xiaoliang and Wang, Cheng and Qi, Jianzhong and Chen, Chaochao and Chen, Longbiao},
  booktitle={WWW},
  pages={673--683},
  year={2023}
}

@inproceedings{dcrnn,
  title={{Diffusion Convolutional Recurrent Neural Network: Data-Driven Traffic Forecasting}},
  author={Li, Yaguang and Yu, Rose and Shahabi, Cyrus and Liu, Yan},
  booktitle={ICLR},
  year={2018}
}

@inproceedings{graphwavenet,
  title={{Graph WaveNet for Deep Spatial-Temporal Graph Modeling}},
  author={Zonghan Wu and Shirui Pan and Guodong Long and Jing Jiang and Chengqi Zhang},
  year={2019},
  booktitle={IJCAI},
  pages = {1907--1913}
}

@inproceedings{gman,
  title={{GMAN: A Graph Multi-Attention Network for Traffic Prediction}},
  author={Zheng, Chuanpan and Fan, Xiaoliang and Wang, Cheng and Qi, Jianzhong},
  booktitle={AAAI},
  pages={1234--1241},
  year={2020}
}

@inproceedings{whdocl,
  title={{When Do Contrastive Learning Signals Help Spatio-Temporal Graph Forecasting?}},
  author={Liu, Xu and Liang, Yuxuan and Huang, Chao and Zheng, Yu and Hooi, Bryan and Zimmermann, Roger},
  booktitle={SIGSPATIAL},
  pages={5:1--5:12},
  year={2022}
}

@article{gamcn,
  title={{A Graph and Attentive Multi-Path Convolutional Network for Traffic Prediction}},
  author={Qi, Jianzhong and Zhao, Zhuowei and Tanin, Egemen and Cui, Tingru and Nassir, Neema and Sarvi, Majid},
  journal={IEEE Transactions on Knowledge and Data Engineering},
  pages={6548--6560},
  year={2022}
}

@article{ge-gan,
  title={{GE-GAN: A Novel Deep Learning Framework for Road Traffic State Estimation}},
  author={Xu, Dongwei and Wei, Chenchen and Peng, Peng and Xuan, Qi and Guo, Haifeng},
  journal={Transportation Research Part C: Emerging Technologies},
  pages={102635},
  year={2020}
}

@misc{osm,
  author = {{OpenStreetMap}},
  title = {{OpenStreetMap US Northeast Data Dump}},
  howpublished = {\url{https://download.geofabrik.de/}},
  year = {2018}
}

@misc{ecmwf_era5,
  author       = {Muñoz Sabater, Joaquín},
  title        = {{ERA5-Land Hourly Data from 1950 to Present}},
  year         = {2019},
  howpublished = {Copernicus Climate Change Service (C3S) Climate Data Store (CDS)},
  url          = {https://cds.climate.copernicus.eu/cdsapp#!/dataset/reanalysis-era5-land}
}

@misc{pems,
  author = {{CalTrans}},
  title = {{California Department of Transportation  Performance Measurement System (PeMS)}},
  howpublished = {\url{https://pems.dot.ca.gov}},
  year={2001},
}

@inproceedings{dtw,
  title={{Using Dynamic Time Warping to Find Patterns in Time Series}},
  author={Berndt, Donald J and Clifford, James},
  booktitle={KDD workshop},
  pages={359--370},
  year={1994}
}

@article{cast,
  title={{Deciphering Spatio-Temporal Graph Forecasting: A Causal Lens and Treatment}},
  author={Xia, Yutong and Liang, Yuxuan and Wen, Haomin and Liu, Xu and Wang, Kun and Zhou, Zhengyang and Zimmermann, Roger},
  journal={NeurIPS},
  pages = {37068--37088},
  year={2023}
}

@inproceedings{stgp,
  title={Prompt-Based Spatio-Temporal Graph Transfer Learning},
  author={Hu, Junfeng and Liu, Xu and Fan, Zhencheng and Yin, Yifang and Xiang, Shili and Ramasamy, Savitha and Zimmermann, Roger},
  booktitle={CIKM},
  pages={890--899},
  year={2024}
}

@inproceedings{hafusion,
  author={Sun, Fengze and Qi, Jianzhong and Chang, Yanchuan and Fan, Xiaoliang and Karunasekera, Shanika and Tanin, Egemen},
  booktitle={IEEE International Conference on Data Engineering}, 
  title={{Urban Region Representation Learning with Attentive Fusion}}, 
  year={2024},
  pages={4409-4421}
}

@article{raissi2019physics,
  title={{Physics-informed Neural Networks: A Deep Learning Framework for Solving Forward and Inverse Problems Involving Nonlinear Partial Differential Equations}},
  author={Raissi, Maziar and Perdikaris, Paris and Karniadakis, George E},
  journal={Journal of Computational physics},
  pages={686--707},
  year={2019},
  publisher={Elsevier}
}

@inproceedings{STGNP,
  title={{Graph Neural Processes for Spatio-temporal Extrapolation}},
  author={Hu, Junfeng and Liang, Yuxuan and Fan, Zhencheng and Chen, Hongyang and Zheng, Yu and Zimmermann, Roger},
  booktitle={KDD},
  pages={752--763},
  year={2023}
}

@article{dualcast,
  title={{DualCast: Disentangling Aperiodic Events from Traffic Series with a Dual-Branch Model}},
  author={Su, Xinyu and Liu, Feng and Chang, Yanchuan and Tanin, Egemen and Sarvi, Majid and Qi, Jianzhong},
  journal={arXiv preprint arXiv:2411.18286},
  year={2024}
}

@inproceedings{ClimODE,
  title={{ClimODE: Climate and Weather Forecasting with Physics-informed Neural ODEs}},
  author={Verma, Yogesh and Heinonen, Markus and Garg, Vikas},
  booktitle={ICLR},
  year={2024}
}

@inproceedings{hettigeairphynet,
  title={{AirPhyNet: Harnessing Physics-Guided Neural Networks for Air Quality Prediction}},
  author={Hettige, Kethmi Hirushini and Ji, Jiahao and Xiang, Shili and Long, Cheng and Cong, Gao and Wang, Jingyuan},
  booktitle={ICLR},
  year={2024}
}

@inproceedings{STDEN,
  title={STDEN: Towards Physics-guided Neural Networks for Traffic Flow Prediction},
  author={Ji, Jiahao and Wang, Jingyuan and Jiang, Zhe and Jiang, Jiawei and Zhang, Hu},
  booktitle={AAAI},
  pages={4048--4056},
  year={2022}
}

@inproceedings{hwang2021climate,
  title={{Climate Modeling with Neural Diffusion Equations}},
  author={Hwang, Jeehyun and Choi, Jeongwhan and Choi, Hwangyong and Lee, Kookjin and Lee, Dongeun and Park, Noseong},
  booktitle={ICDM},
  pages={230--239},
  year={2021}
}

@article{zhang2024physics,
  title={{Physics-informed Deep Learning for Traffic State Estimation based on the Traffic Flow Model and Computational Graph Method}},
  author={Zhang, Jinlei and Mao, Shuai and Yang, Lixing and Ma, Wei and Li, Shukai and Gao, Ziyou},
  journal={Information Fusion},
  pages={101971},
  year={2024}
}

@inproceedings{shi2021physics,
  title={{Physics-informed Deep Learning for Traffic State Estimation: A Hybrid Paradigm Informed by Second-order Traffic Models}},
  author={Shi, Rongye and Mo, Zhaobin and Di, Xuan},
  booktitle={AAAI},
  pages={540--547},
  year={2021}
}

@article{li2024opencity,
  title={{OpenCity: Open Spatio-temporal Foundation Models for Traffic Prediction}},
  author={Li, Zhonghang and Xia, Long and Shi, Lei and Xu, Yong and Yin, Dawei and Huang, Chao},
  journal={arXiv preprint arXiv:2408.10269},
  year={2024}
}

@inproceedings{ustd,
  title={{Towards Unifying Diffusion Models for Probabilistic Spatio-temporal Graph Learning}},
  author={Hu, Junfeng and Liu, Xu and Fan, Zhencheng and Liang, Yuxuan and Zimmermann, Roger},
  booktitle={SIGSPATIAL},
  pages={135--146},
  year={2024}
}

@article{dualSTN,
  title={{Decoupling Long-and Short-term Patterns in Spatiotemporal Inference}},
  author={Hu, Junfeng and Liang, Yuxuan and Fan, Zhencheng and Liu, Li and Yin, Yifang and Zimmermann, Roger},
  journal={IEEE Transactions on Neural Networks and Learning Systems},
  pages={16328-16340},
  year = {2024}
}

@inproceedings{kits,
  title={{KITS: Inductive Spatio-temporal Kriging with Increment Training Strategy}},
  author={Xu, Qianxiong and Long, Cheng and Li, Ziyue and Ruan, Sijie and Zhao, Rui and Li, Zhishuai},
  booktitle={AAAI},
  pages={12945--12953},
  year={2025}
}

@article{lwr1,
  title={{On Kinematic Waves. II. A Theory of Traffic Flow on Long Crowded Roads}},
  author={Lighthill, M. J. and Whitham, G. B.},
  journal={Proceedings of the Royal Society of London. Series A. Mathematical and Physical Sciences},
  pages={317--345},
  year={1955},
  publisher={The Royal Society}
}

@article{lwr2,
  title={{Shock Waves on the Highway}},
  author={Richards, P. I.},
  journal={Operations Research},
  pages={42--51},
  year={1956}
}

@article{greenshields,
  title={{A Study of Traffic Capacity}},
  author={Greenshields, B. D.},
  journal={Proceedings of the Highway Research Board},
  pages={448--477},
  year={1935}
}

@article{st-pef,
  title={{Traffic Flow Prediction based on Spatiotemporal Potential Energy Fields}},
  author={Wang, Jingyuan and Ji, Jiahao and Jiang, Zhe and Sun, Leilei},
  journal={IEEE Transactions on Knowledge and Data Engineering},
  pages={9073--9087},
  year={2022},
  publisher={IEEE}
}

@article{fspd,
  title={{Detecting Spatiotemporal Propagation Patterns of Traffic Congestion from Fine-grained Vehicle Trajectory Data}},
  author={Xiong, Haoyi and Zhou, Xun and Bennett, David A},
  journal={International Journal of Geographical Information Science},
  pages={1157--1179},
  year={2023}
}

@misc{geohash,
  author= {Gustavo Niemeyer},
  title= {{GeoHash}},
  year= {2008},
  howpublished={\url{https://github.com/davetroy/geohash}}
}

@misc{llama2024,
  author       = {{Meta}},
  title        = {{LLaMA Models}},
  year         = {2024},
  howpublished = {\url{https://www.llama.com/}}
}

@misc{lhy_charbert_2023,
  author       = {Li, Hongyi},
  title        = {{Char-BERT: Character-level BERT Model}},
  year         = {2023},
  howpublished = {\url{https://huggingface.co/lhy/char-bert-base-uncased}}
}

@inproceedings{stsm,
  title={{Spatial-temporal Forecasting for Regions without Observations}},
  author={Su, Xinyu and Qi, Jianzhong and Tanin, Egemen and Chang, Yanchuan and Sarvi, Majid},
  year={2024},
  booktitle={EDBT},
  pages = {488--500}
}

@article{nigam2023hybrid,
  title={{Hybrid Deep Learning Models for Traffic Stream Variables Prediction during Rainfall}},
  author={Nigam, Archana and Srivastava, Sanjay},
  journal={Multimodal Transportation},
  year={2023},
pages = {100052}
}

@article{sun2025flexireg,
  title={{FlexiReg: Flexible Urban Region Representation Learning}},
  author={Sun, Fengze and Chang, Yanchuan and Tanin, Egemen and Karunasekera, Shanika and Qi, Jianzhong},
  journal={arXiv preprint arXiv:2503.09128},
  year={2025}
}

@misc{ruan2025retrieval,
      title={{A Retrieval Augmented Spatio-Temporal Framework for Traffic Prediction}},
      author={Ruan, Weilin and Dang, Xilin and Zhou, Ziyu and Lyu, Sisuo and Liang, Yuxuan},
      year={2025},
      eprint={2508.16623},
      archivePrefix={arXiv},
      primaryClass={cs.CL}
}

@misc{nrel,
  author = {{NREL}},
  title = {{Solar Power Data for Integration Studies}},
  howpublished = {\url{https://www.nrel.gov/grid/solar-power-data.html}},
  year = {2018}
}

@incollection{huber1992robust,
  title={{Robust Estimation of a Location Parameter}},
  author={Huber, Peter J},
  booktitle={Breakthroughs in Statistics: Methodology and distribution},
  pages={492--518},
  year={1992}
}

@article{mystakidis2024traffic,
  title={{Traffic Congestion Prediction and Missing Data: A Classification Approach Using Weather Information}},
  author={Mystakidis, Aristeidis and Tjortjis, Christos},
  journal={International Journal of Data Science and Analytics},
  pages={1--20},
  year={2024}
}

@article{lin2023diversifying,
  title={Diversifying Spatial-temporal Perception for Video Domain Generalization},
  author={Lin, Kun-Yu and Du, Jia-Run and Gao, Yipeng and Zhou, Jiaming and Zheng, Wei-Shi},
  journal={NeurIPS},
  pages={56012--56026},
  year={2023}
}

@inproceedings{zhou2024sdformer,
  title={{SDformer: Transformer with Spectral Filter and Dynamic Attention for Multivariate Time Series Long-term Forecasting}},
  author={Zhou, Ziyu and Lyu, Gengyu and Huang, Yiming and Wang, Zihao and Jia, Ziyu and Yang, Zhen},
  booktitle={IJCAI},
  year={2024}
}
\clearpage
\newpage
\appendix
\section{A~Related Work}
Extrapolation, kriging, and forecasting are core spatial-temporal tasks that often share backbone spatial-temporal learning models~\cite{ustd,stgp}, with recent approaches combining graph and sequence models to capture spatial-temporal dependencies~\cite{graphwavenet,gman,STGNP,dualSTN}.
Kriging and extrapolation are particularly challenging due to the lack of ground truth at unobserved target locations. To simulate such settings during training, existing models typically mask observed locations~\cite{ignnk,increase} or interpolate between them~\cite{kits}, and are trained to reconstruct the masked values. While being effective for scattered missing observations, these approaches often fail in more realistic scenarios involving large, continuous unobserved regions.
To address this issue, STSM~\cite{stsm} introduces a selective masking strategy based on location similarity, encouraging the model to learn from locations that resemble those in the unobserved region. However, because the similarity is derived from static auxiliary features (e.g., POI categories and geographic coordinates), it fails to capture dynamic traffic patterns, thereby limiting the STSM’s generalisability in complex real-world environments.

To improve generalisability, recent efforts have explored (1)~\emph{contrastive learning}, (2)\emph{physics-guided models}, and (3)~\emph{external data signals}.

\paragraph{Contrastive Learning (CL)-Based Models.} 
CL-based models learn transferable representations by aligning similar inputs while distinguishing dissimilar ones through positive and negative sample pairs. For  spatial-temporal forecasting, models often construct different graph views by perturbing nodes or edges, helping to uncover invariant patterns across space and time~\cite{whdocl}. To enable generalisation to unobserved regions, STSM~\cite{stsm} contrasts the forecasts learned for the full input graph with those learned for selectively masked graphs, encouraging consistency under simulated unobserved scenarios. We also follow a contrastive learning backbone. Instead of selective masking, we adopt a more flexible random subgraph masking strategy, which does not require knowledge about the similarity between the locations in the observed and the unobserved regions as mentioned above. 

\paragraph{Physics-Guided Models.} Physics-guided models have shown strong potential for enhancing generalisation in scientific domains such as fluid dynamics and climate modelling, by embedding known physical laws into deep learning. Recent traffic forecasting studies have explored physics guidance to introduce inductive biases and improve model accuracy given sparse or irregular observations.

Physics-guided forecasting models generally follow two directions: (1) using neural networks to solve or approximate Partial Differential Equations (PDEs) or (2) incorporating physical constraints into the training loss.

Models in the first category leverage neural Ordinary Differential Equations (ODEs) or diffusion-inspired dynamics to simulate latent physical processes~\cite{hwang2021climate,STDEN,st-pef,ClimODE}. Such models embed physics priors implicitly through model design, capturing continuous-time evolution aligned with domain dynamics. They require fully observed spatial domains to initialise physical states, limiting their applicability to settings with  unobserved regions.

Models in the second category explicitly incorporate physical constraints into the training objective. Physics-Informed Neural Networks (PINNs)~\cite{raissi2019physics} extend this idea by introducing residuals of governing equations directly into the loss function, enabling models to learn solutions consistent with physical laws. PINNs have been successfully applied to domains such as air quality estimation~\cite{hettigeairphynet} and climate dynamics~\cite{hwang2021climate}, where robustness and generalisation with sparse or noisy observations are critical. For traffic forecasting, classical physical models such as the Lighthill-Whitham-Richards~\cite{lwr1,lwr2} and Greenshields~\cite{greenshields} equations define macroscopic relationships among speed, density, and flow, and have been adopted as soft constraints~\cite{shi2021physics,zhang2024physics}.

Despite these advancements, existing physics-guided models are often grid-based or assume continuous spatial coordinates, posing challenges for graph-based traffic modelling. Adapting PINN-style techniques to discrete graphs remains underexplored for improving spatial-temporal generalisation to unobserved regions.

\paragraph{Exploiting External Data Signals.}  
Beyond traffic observations, a variety of external (auxiliary) signals have been explored to enhance robustness and generalisation in spatial-temporal forecasting tasks. These signals can be broadly grouped into: (1)~multi-sourced spatial-temporal data e.g., traffic flow data, taxi demand data, and traffic index statics~\cite{li2024opencity};  (2)~external contextual features, e.g., weather conditions, major events, and holidays~\cite{dualcast}.
Most existing work focuses on using these signals to enhance sequence-based models to detect irregular events or handle short-term missing data. The use of such signals for guiding model generalisation to unobserved regions, especially when combined with advanced spatial-temporal graph networks, remains underexplored.

\section{B~Additional Model Details}
\label{appsub:methology}
\subsection{B.1~Details of \model\ Backbone}
Following~\citet{stsm}, our model backbone takes a contrastive learning architecture (Fig.~\ref{app:fig:backbone}). At each training epoch, a masked graph view $G_o^m$ is constructed by randomly masking a subgraph from the observed graph $G_o$, yielding two input sequences (i.e., two \emph{views}): the original $\mathbf{X}_{G_o}^{t-T+1:t}$ and the masked $\mathbf{X}_{G_o^m}^{t-T+1:t}$. Then, we generate pseudo-observations for the masked locations, allowing the construction of both spatial (i.e., geo-coordinate) proximity-based and temporal (i.e., traffic series) similarity-based adjacency matrices.

\begin{figure}[!t]
     \centering
     \begin{subfigure}
         \centering
         \includegraphics[width=\columnwidth]{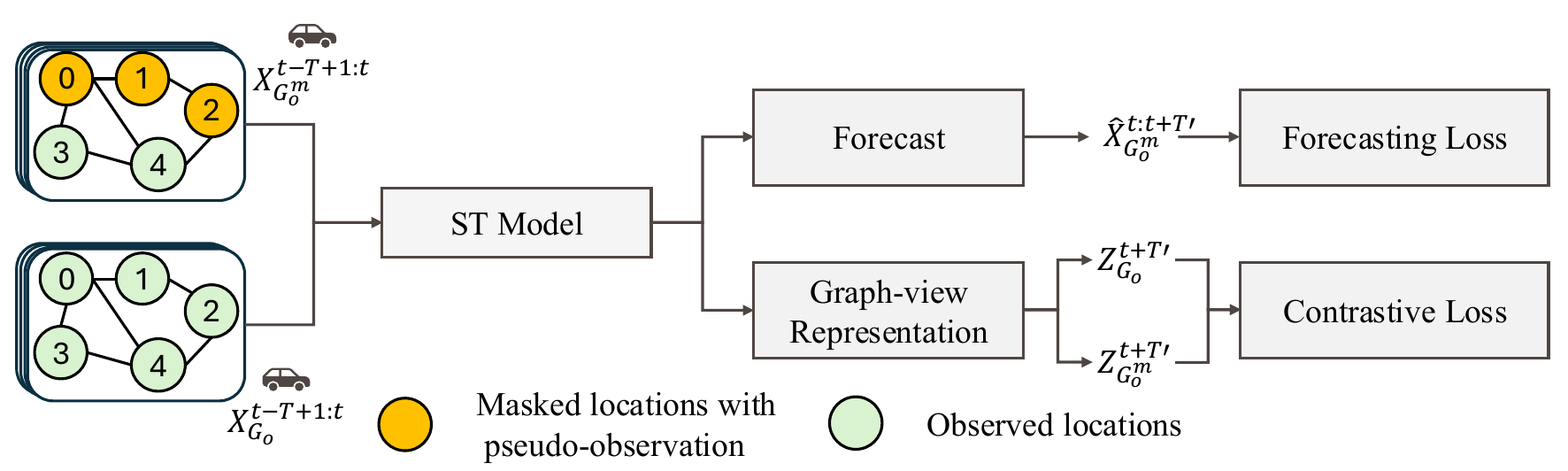}
     \end{subfigure}
     \vspace{-6mm}
     \caption{Framework of \model\ backbone.}
     \label{app:fig:backbone}
\end{figure}

The spatial adjacency matrix $\textbf{A}_{sg}$ is defined by Eq.~\ref{eq:A_sg}, where $\epsilon_{sg}$ is a hyperparameter (value set following~\citet{stsm}), and $dist(c_i,c_j)$ denotes the Euclidean distance between locations $i$ and $j$ ($c_i$ and $c_j$ are their geo-coordinates.)
\begin{equation}\label{eq:A_sg}
  A_{sg,ij}=\begin{cases}
  1 & \text{exp}(-\frac{\text{dist}(c_i,c_j)^2}{\sigma^2}) \geq \epsilon_{sg}, \\
  0 & \text{otherwise}.
\end{cases}
\end{equation}

\model\ uses Dynamic Time Warping (DTW)~\cite{dtw} to compute the temporal similarity-based (non-symmetric) adjacency matrix, which links $q_{kk}$ most similar observed–observed location pairs and $q_{ku}$ most similar observed–unobserved (masked) pairs. \model\ constructs directed edges from observed to unobserved nodes only, to prevent unobserved representations from affecting observed node embeddings (by message passing). The temporal adjacency matrix for model training is denoted as $\mathbf{A}_{{dtw}}^{{train}} \in \mathbb{R}^{N_o \times N_o}$, and that for model testing is $\mathbf{A}_{{dtw}} \in \mathbb{R}^{N \times N}$. As masking is dynamic across epochs, $\mathbf{A}_{{dtw}}^{{train}}$ is updated at each training epoch.

Then, \model\ feeds $\mathbf{X}_{G_o}^{t-T+1:t}$ and $\mathbf{X}_{G_o^m}^{t-T+1:t}$ into a spatial-temporal (ST) model composed of stacked layers, each containing a temporal block and a spatial block operating in parallel. The temporal block uses 1-D temporal convolution networks (TCNs) to extract temporal dependencies, while the spatial block applies two graph convolution layers (GCNs) based on $\mathbf{A}_{dtw}^{train}$ and $\mathbf{A}_{sg}^{train}$, respectively. Their outputs are aggregated via element-wise maximum to form the spatial representation. The outputs of the TCN and GCN blocks are summed to produce the output of the $l$-th layer.

The final output of the ST model is passed through two linear layers with activation functions to generate a forecast $\hat{\mathbf{X}}^{t+1:t+T'}_{G_o^m}$. To obtain graph representations, we extract the output at the last predicted time step and apply a linear projection over all nodes, resulting in $\mathbf{Z}_{G_o}^{t+T'}$ and $\mathbf{Z}_{G_o^m}^{t+T'}$. For a batch size of $|B|$, this yields $2|B|$ representations.

To ensure that the representations are robust and invariant to masking, we apply contrastive learning between the original graph $G_o$ and its masked variant $G_o^m$. For each time window $t$, the pair $(\mathbf{Z}_{G_o}^{t+T'}, \mathbf{Z}_{G_o^m}^{t+T'})$ is treated as a positive pair, encouraging consistency under masking. In contrast, pairs from different time windows $(t' \ne t)$ are treated as negative samples. Accordingly, the contrastive loss is defined as:
{\small
\begin{equation}
  L_{cl} = -\frac{1}{|B|}\sum_{t} \log\frac{\exp(\text{sim}(\mathbf{Z}_{G_o}^{t+T'}, \mathbf{Z}_{G_o^m}^{t+T'})/\omega)}{\sum_{t' \ne t} \exp(\text{sim}(\mathbf{Z}_{G_o}^{t+T'}, \mathbf{Z}_{G_o^m}^{t'+T'})/\omega)},
\end{equation}
}
where $\omega=0.5$ is a temperature parameter and $\text{sim}(\cdot)$ denotes a similarity function (e.g., cosine similarity).

The final loss combines forecast error (RMSE) and representation alignment: $L = L_{pred} + \lambda L_{cl}$, where $\lambda$ is a weighting factor. We follow the default hyperparameter from~\citet{stsm} without fine-tuning in \model.

\paragraph{Sub-graph and Random Sub-graph Masking.} We define the sub-graph of an observed location as its 1-hop neighbours. Given a masking ratio $\delta_m$, we aim to mask approximately $N_o \cdot \delta_m$ observed locations. Since sub-graph sizes vary, \model\ randomly and iteratively selects observed nodes and masks both the node and its 1-hop neighbours, until the desired number of masked nodes is reached.

\subsection{B.2~Spatial Embeddings of \model}

\begin{figure}[t]
     \centering
     \begin{subfigure}
         \centering
         \includegraphics[width=\columnwidth]{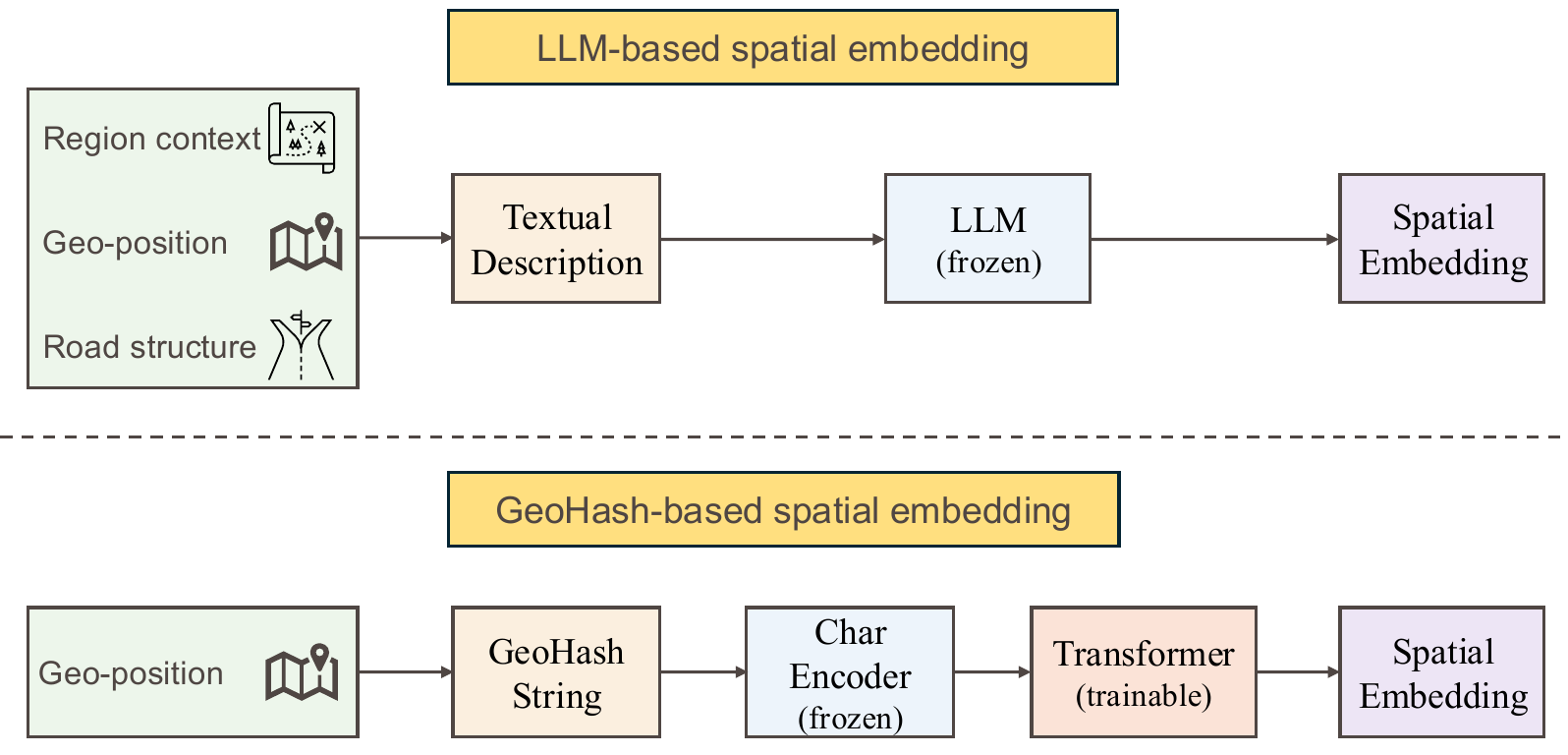}
     \end{subfigure}
     \caption{Comparison of two spatial embedding strategies. Upper: LLM-Based spatial embedding uses regional and road network information for spatial embedding. Lower: GeoHash-based spatial embedding only uses geo-location.}
     \label{app:fig:spatial_embedding}
\end{figure}

\paragraph{Generating Process.} We introduce two methods for generating continuously differentiable spatial embeddings: (1)~\emph{LLM-based spatial embedding} (SE-L), and (2)~\emph{GeoHash-based spatial embedding} (SE-H). These two methods differ in both their generation processes and update mechanisms. As illustrated in Fig.~\ref{app:fig:spatial_embedding}, SE-L is produced using a frozen large language model (LLM), ensuring that the geographic knowledge acquired during pre-training is retained. These embeddings remain fixed during training. In contrast, SE-H is generated using a pre-trained character-level encoder, and such embeddings are updated during training via a Transformer model.

\paragraph{Location Description Generation for SE-L.} Generating location descriptions is the first step in our LLM-based spatial embedding module~\cite{sun2025flexireg}. We construct a textual description for each location by incorporating four categories of information: (i)~the location’s geo-coordinates, including latitude, longitude, and address; (ii)~geometric properties describing the shape and extent of the surrounding area, particularly the spatial coverage of nearby POIs; (iii)~POI information~\cite{stsm,hafusion}, such as the categories and counts of POIs within the region; and (iv)~attributes of the closest road segments, including road type, maximum speed, number of lanes, and whether the road is one-way. Fig.~\ref{app:fig:prompt} shows an example of a location description used as a prompt for generating spatial embeddings via a large language model (LLaMA 3 8B Instruct~\cite{llama2024}). The dimension of each spatial embedding generated by the LLM is 4,096.

\begin{figure*}[!ht]
     \centering
     \begin{subfigure}
         \centering
         \includegraphics[width=0.7\linewidth]{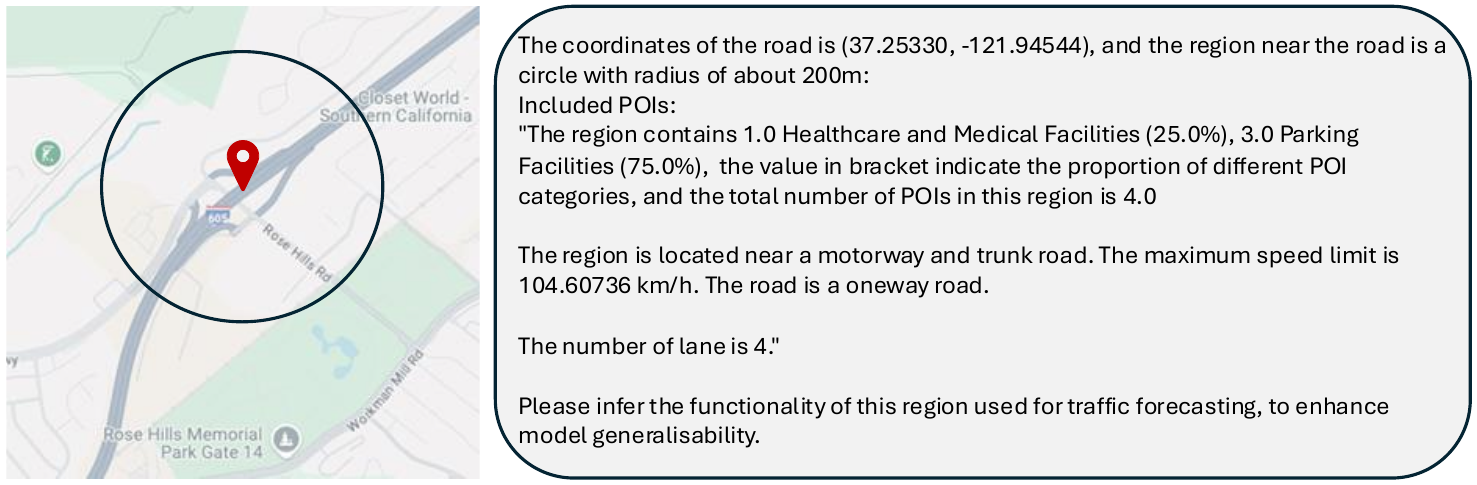}
     \end{subfigure}
     \caption{An example of location description used as a prompt for generating LLM-based spatial embeddings.}
     \label{app:fig:prompt}
\end{figure*}

\subsection{B.3~Spatial Grouping Module in \model}
Fig.~\ref{fig:spatial_grouping} illustrates the spatial grouping module, which softly clusters locations into latent groups, allowing \model\ to capture shared group-level patterns while suppressing ad hoc variations at individual locations through an entropy regularisation term.

\begin{figure}[t]
     \centering
     \begin{subfigure}
         \centering
         \includegraphics[width=\columnwidth]{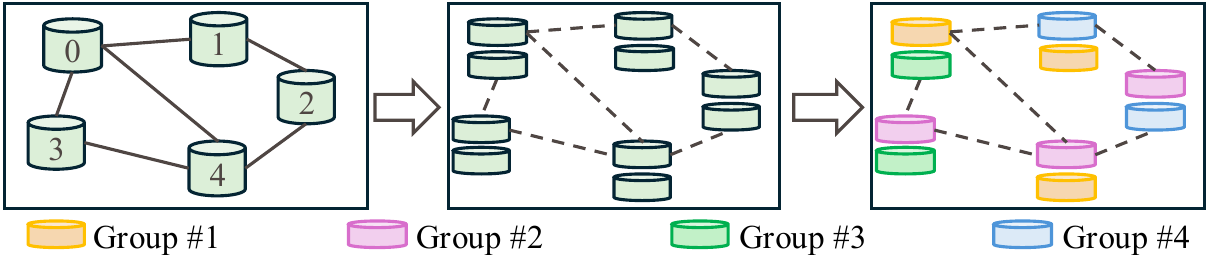}
     \end{subfigure}
     % \vspace{-5mm}
     \caption{Spatial grouping module. The feature dimensionality $D$ of each node is split into $cg$ channel groups, producing $N_o \times cg$ samples. These samples are softly assigned to $cg\times sg$ spatial groups through assignment weights, with an entropy loss encouraging confident (i.e., one-hot-like) group assignments.}
     \label{fig:spatial_grouping}
\end{figure}

\subsection{B.4~Physics-informed Module in \model}
The physics-informed module introduces a constraint based on physical principles of traffic systems formulated by the  Lighthill--Whitham--Richards (LWR) model~\cite{lwr1,lwr2}. 
LWR is a macroscopic first-order traffic flow model that describes the evolution of traffic density over space and time using a conservation law: 
{\small
\begin{equation}
\label{c5:eq:lwr_basic}
    \frac{\partial \rho}{\partial t} + \frac{\partial (\rho x)}{\partial l} = 0,
\end{equation}
}
where \(\rho = \rho(l, t)\) denotes traffic density, \(x = x(l, t)\) denotes traffic speed (i.e., traffic observation in our study), and \(t\) and \(l\) denote time and geo-location, respectively. 

As benchmark datasets often lack traffic density observations, we rewrite the formulation using velocity, assuming a closed system with a functional density–velocity relationship~\cite{greenshields}:
{\small
\begin{equation}
x = x_{fspd} \left(1 - \frac{\rho}{\rho_{{max}}} \right) \quad \Rightarrow \quad
\rho = \rho_{{max}} \left(1 - \frac{x}{x_{fspd}} \right).
\end{equation}
}
Here, $x_{fspd}$ denotes Free-Flow Traffic Speed, which refers to the speed at which vehicles travel under low traffic density (i.e., no delays caused by congestion, signals, or other disruptions). Following~\citet{fspd}, we define the 85th percentile of traffic speed during non-peak hours (i.e., 9:00am–4:00pm and 10:00pm–6:00am) as the free flow speed for each location, i.e., $
[\mathbf{X}_{i,{fspd}} = \mathrm{Percentile}_{85} \left( \mathbf{X}_i^t \mid t \in \mathcal{T}_{{non-peak}} \right), \forall i \in G_o]$. We estimate $\mathbf{X}_{i,{fspd}}$ for all observed locations from the training set. 
The spatial and temporal derivatives hence can be rewritten as:
{\small
\begin{equation}
\begin{aligned}
\frac{\partial (\rho x)}{\partial l} = \rho_{{max}} \frac{\partial}{\partial l} \left( x - \frac{x^2}{x_{fspd}} \right) = \rho_{{max}} \left( 1 - \frac{2x}{x_{{fspd}}} \right) \cdot \frac{\partial x}{\partial l}.
\end{aligned}
\end{equation}
}

Further computing the temporal derivative yields:
{\small
\begin{equation}
\frac{\partial \rho}{\partial t} 
= \frac{\partial}{\partial t} \left[ \rho_{{max}} \left(1 - \frac{x}{x_{{fspd}}} \right) \right]
= - \frac{\rho_{{max}}}{x_{{fspd}}} \cdot \frac{\partial x}{\partial t}.
\end{equation}
}

Then, this partial derivative is substituted into the conservation equation, Eq.~\ref{c5:eq:lwr_basic}:
{\small
\begin{align}
-\frac{\rho_{{max}}}{x_{fspd}} \frac{\partial x}{\partial t}
+ \rho_{{max}} \left(1 - \frac{2x}{x_{fspd}} \right) \frac{\partial x}{\partial l} = 0.
\end{align}
}
Multiplying both sides by $-\frac{x_{fspd}}{\rho_{{max}}}$ gives: $\frac{\partial x}{\partial t} + (2x - x_{fspd}) \frac{\partial x}{\partial l} = 0$. The left-hand side is the physics residual~$R$: 
{\small
\begin{equation}
\label{app:eq:phy_residual}
R=\frac{\partial x}{\partial t} + (2x - x_{fspd}) \frac{\partial x}{\partial l}
\end{equation}
}
To guide \model\ to follow the physical principles, we use the Huber loss~\cite{huber1992robust} to  minimise the physics residual (i.e., to penalise violations of the physical law):  
{\small
\begin{equation}
\label{app:eq:phy_loss}
\mathcal{L}_{{phy}} =\mathrm{Huber}(R,\delta)
\end{equation}
}
which penalises violations of the physical principles and is jointly optimised with the main forecasting objective.

\section{C Additional Experimental Details}
\subsection{C.1~Detailed Experimental Setup}
\paragraph{Datasets.} 
We conduct experiments on four highway traffic datasets~({PEMS-Bay}, {PEMS07}, {PEMS08}~\cite{pems} and {METR-LA})~and an urban traffic dataset~({Melbourne}). \textbf{PEMS-Bay}, \textbf{PEMS07}, and \textbf{PEMS08} contain traffic speed data collected by 358, 400 and 400 sensors in California, \textbf{METR-LA}~\cite{dcrnn} contains traffic speed data collected by 207 sensors in Los Angeles County. All highway traffic datasets are collected at 5-minute intervals, i.e., 288 time slots per day. The urban traffic dataset Melbourne is collected by 182 sensors in Melbourne at 15-minute intervals, i.e., 96 time slots per day. Additionally, we generalise our proposed model \model\ to a solar power dataset, named NREL~\cite{nrel}, which contains solar power data collected by 137 sensors in Alabama. Table~\ref{tab:dataset_info} lists the dataset statistics, and Fig.~\ref{app:fig:loc_vis} visualises the sensor distribution among all datasets. 

\begin{table*}[!t]
\centering
\begin{tabular}{p{1.8cm}|p{2.3cm}|p{4cm}rr}
  \hlineB{3}
  Dataset & Data Type& Time period & Interval & \#Sensors\\
  \hline \hline
  PEMS-07 & Traffic Speed & 01/09/2022 - 31/12/2022 & 5 min & 400\\
  PEMS-08 & Traffic Speed & 01/09/2022 - 31/12/2022 & 5 min & 400\\
  PEMS-Bay & Traffic Speed & 01/01/2017 - 30/06/2017 & 5 min & 325\\
  METR-LA & Traffic Speed & 01/03/2012 - 27/06/2012 & 5 min & 207\\
  Melbourne & Traffic Speed & 01/07/2022 - 30/09/2022 & 15 min &182\\
  NREL & Solar Power & 01/01/2006 - 31/12/2006 & 5 min & 137\\
  \hlineB{3}
\end{tabular}
\caption{Dataset Statistics}
  \label{tab:dataset_info}
\end{table*}

\begin{figure*}[!t]
    \centering
    \subfigure[PEM07]{\includegraphics[width=0.3\textwidth]{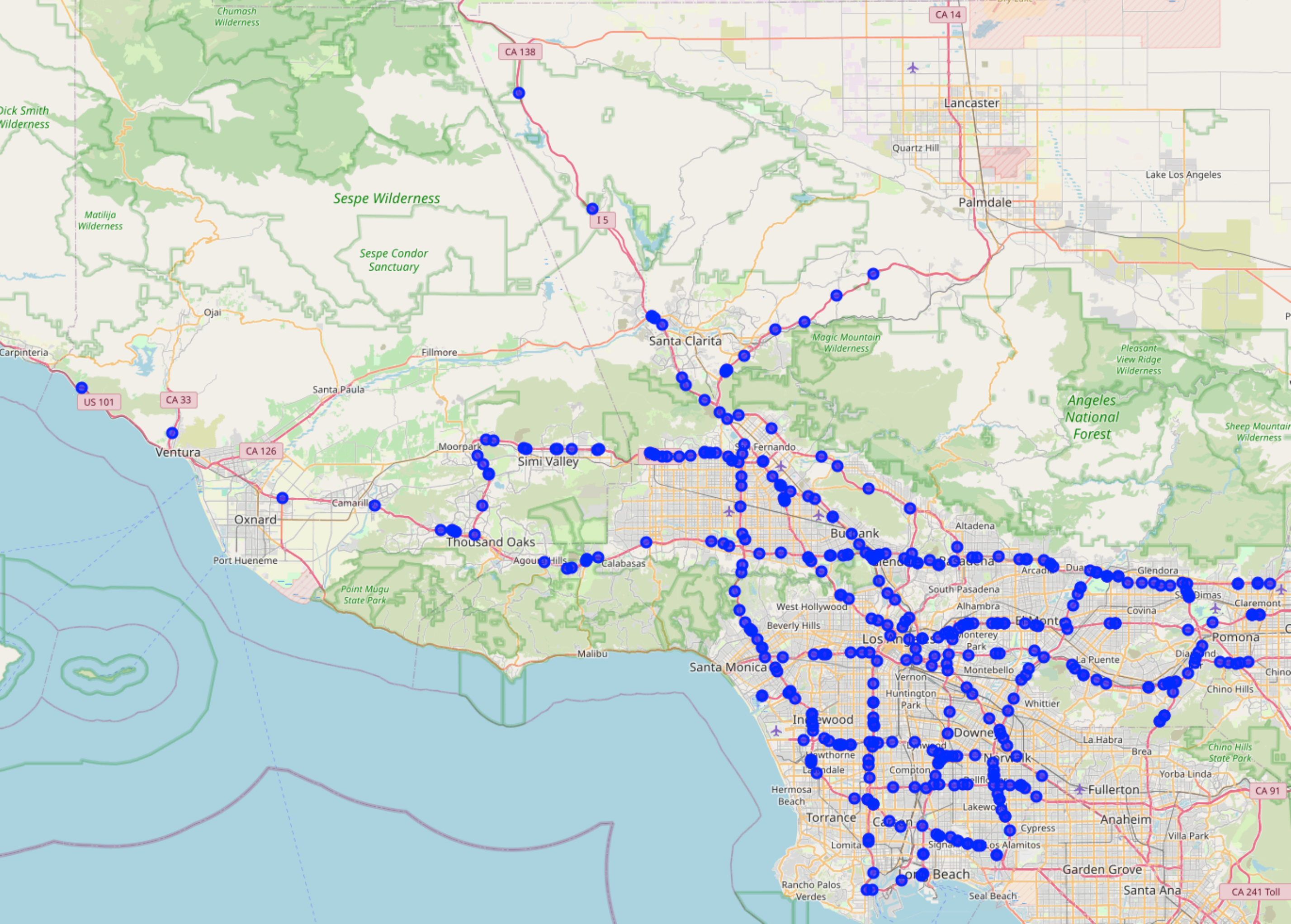}} 
    \subfigure[PEM08]{\includegraphics[width=0.3\textwidth]{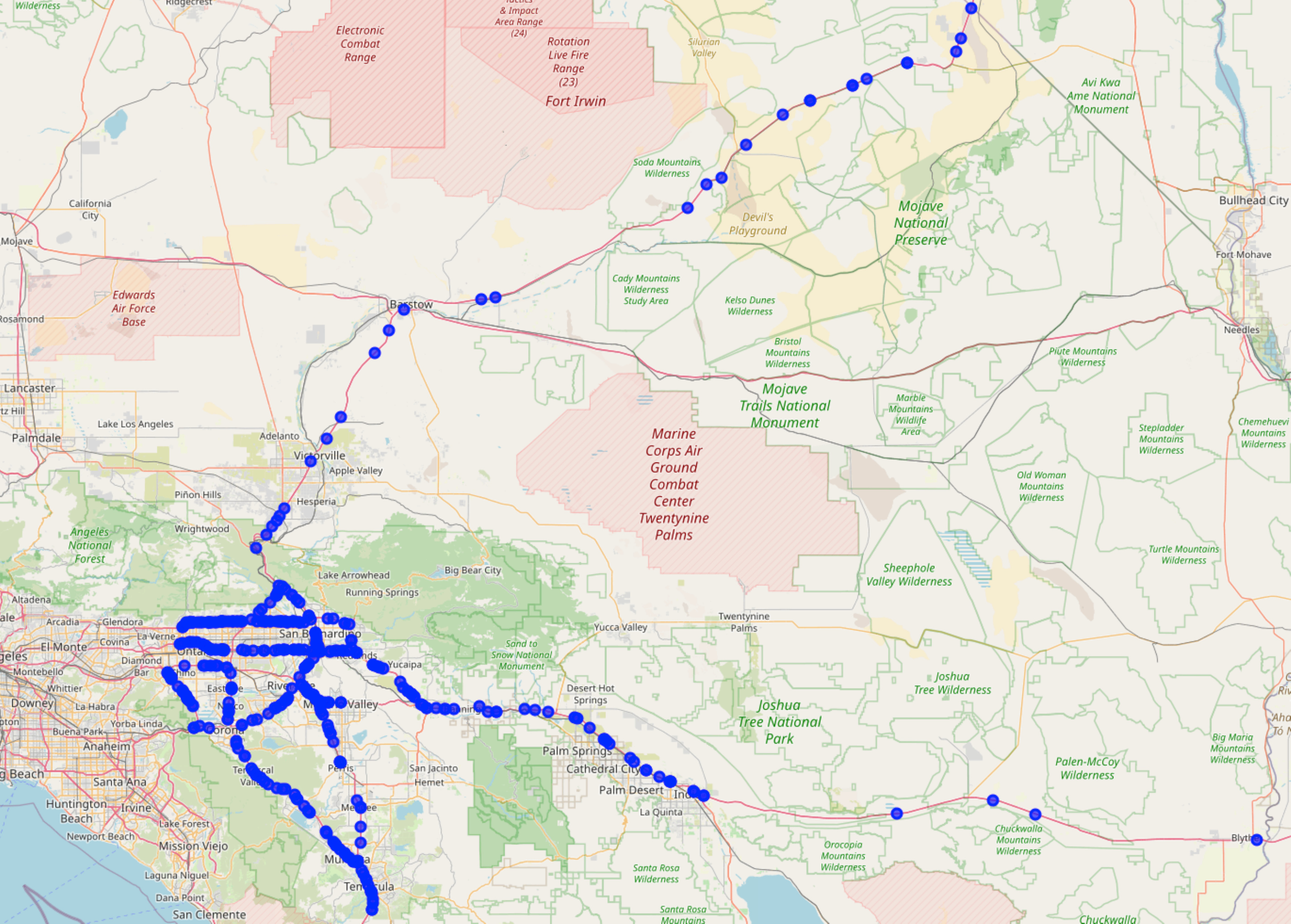}} 
    \subfigure[PEMS-Bay]{\includegraphics[width=0.3\textwidth]{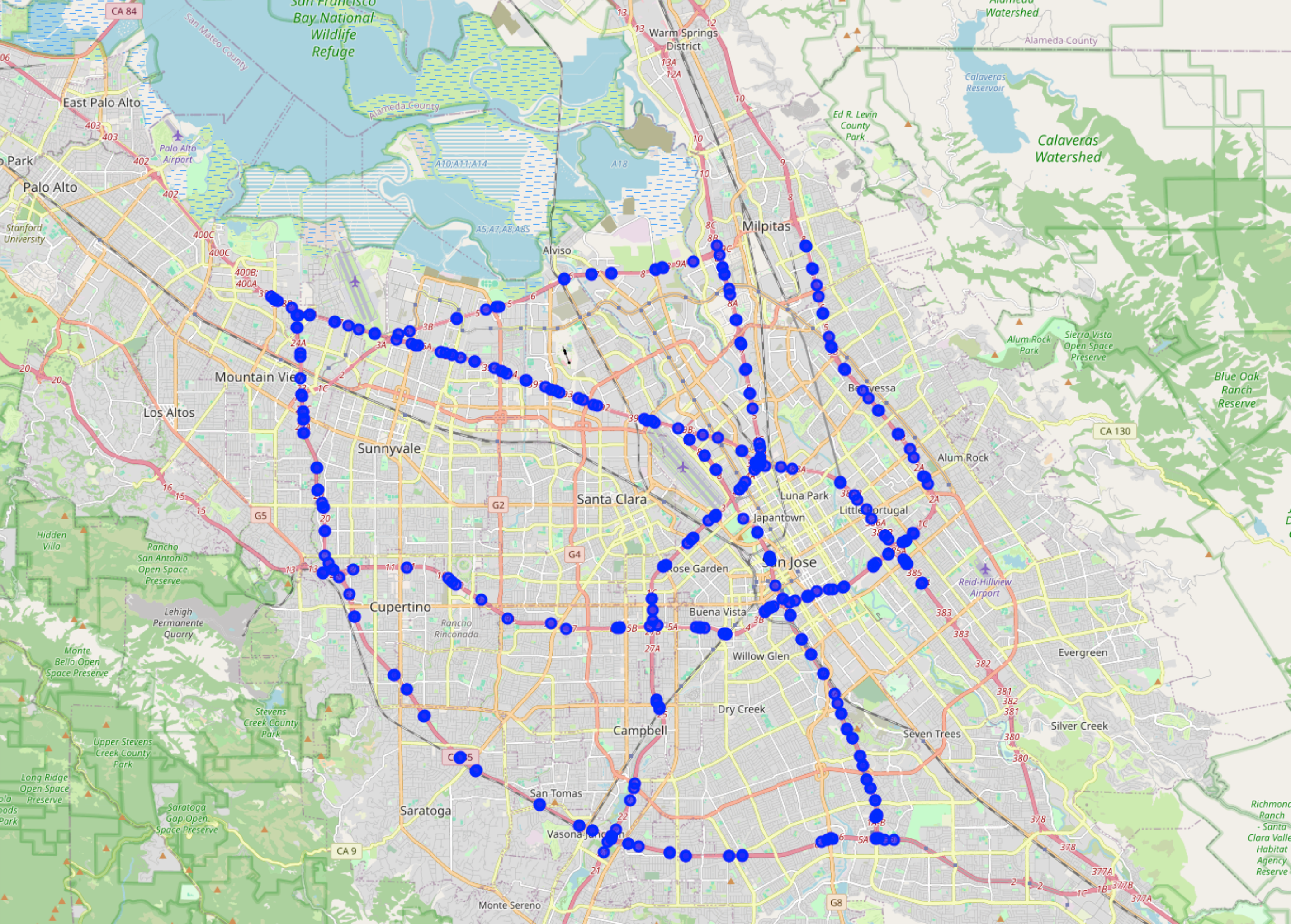}}
    \subfigure[METR-LA]{\includegraphics[width=0.3\textwidth]{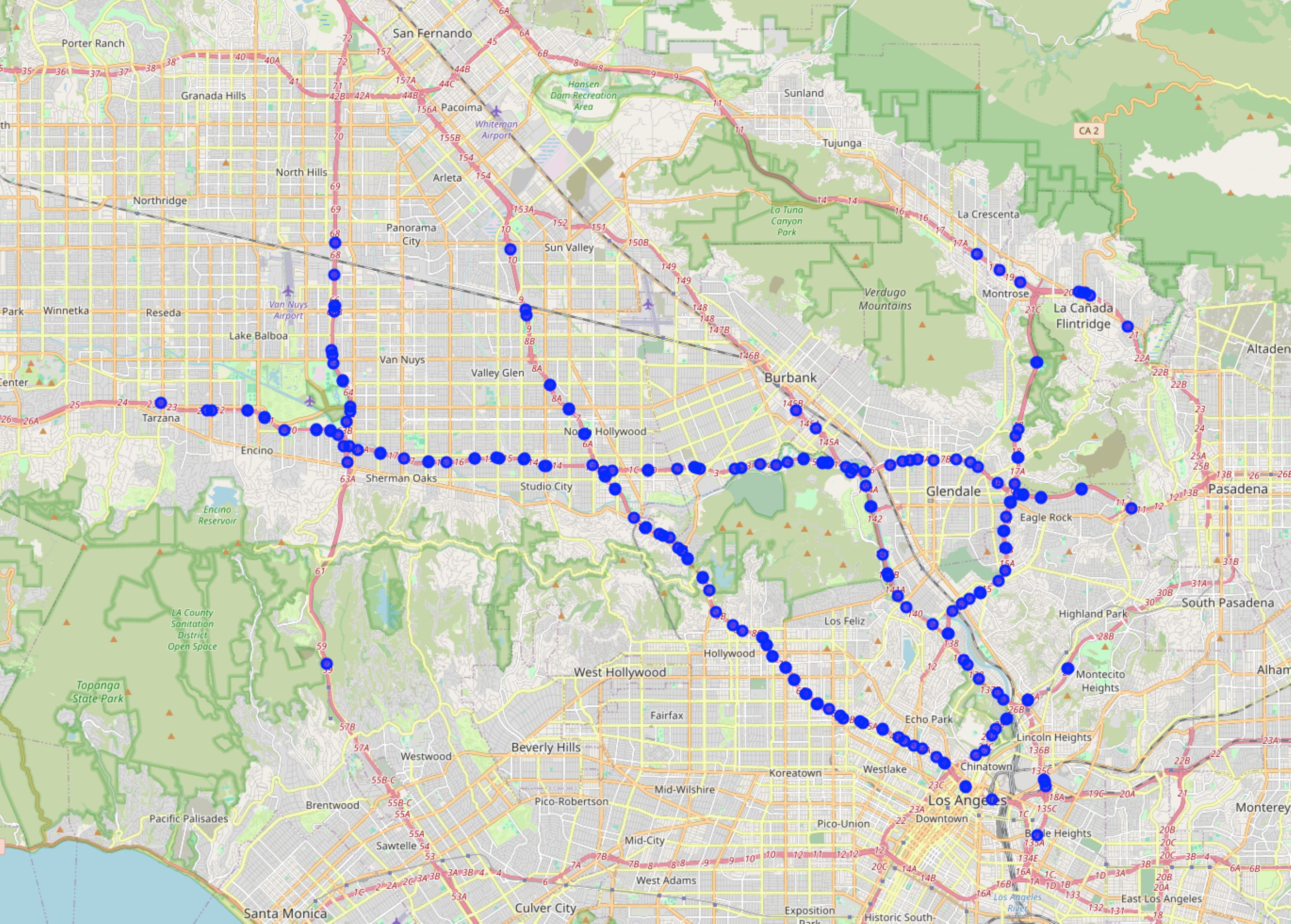}}
    \subfigure[Melbourne]{\includegraphics[width=0.3\textwidth]{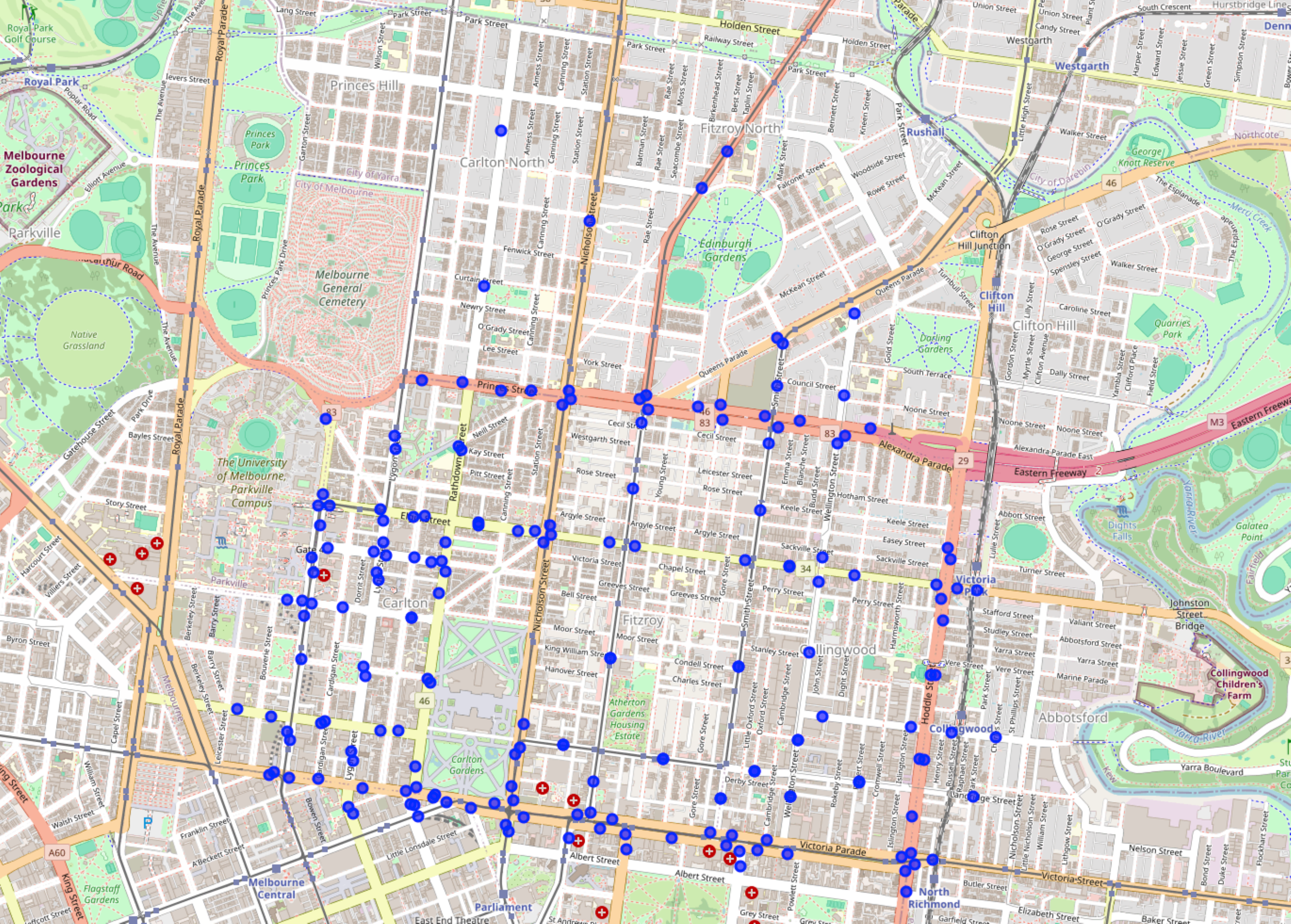}} 
    \subfigure[NREL]{\includegraphics[width=0.3\textwidth]{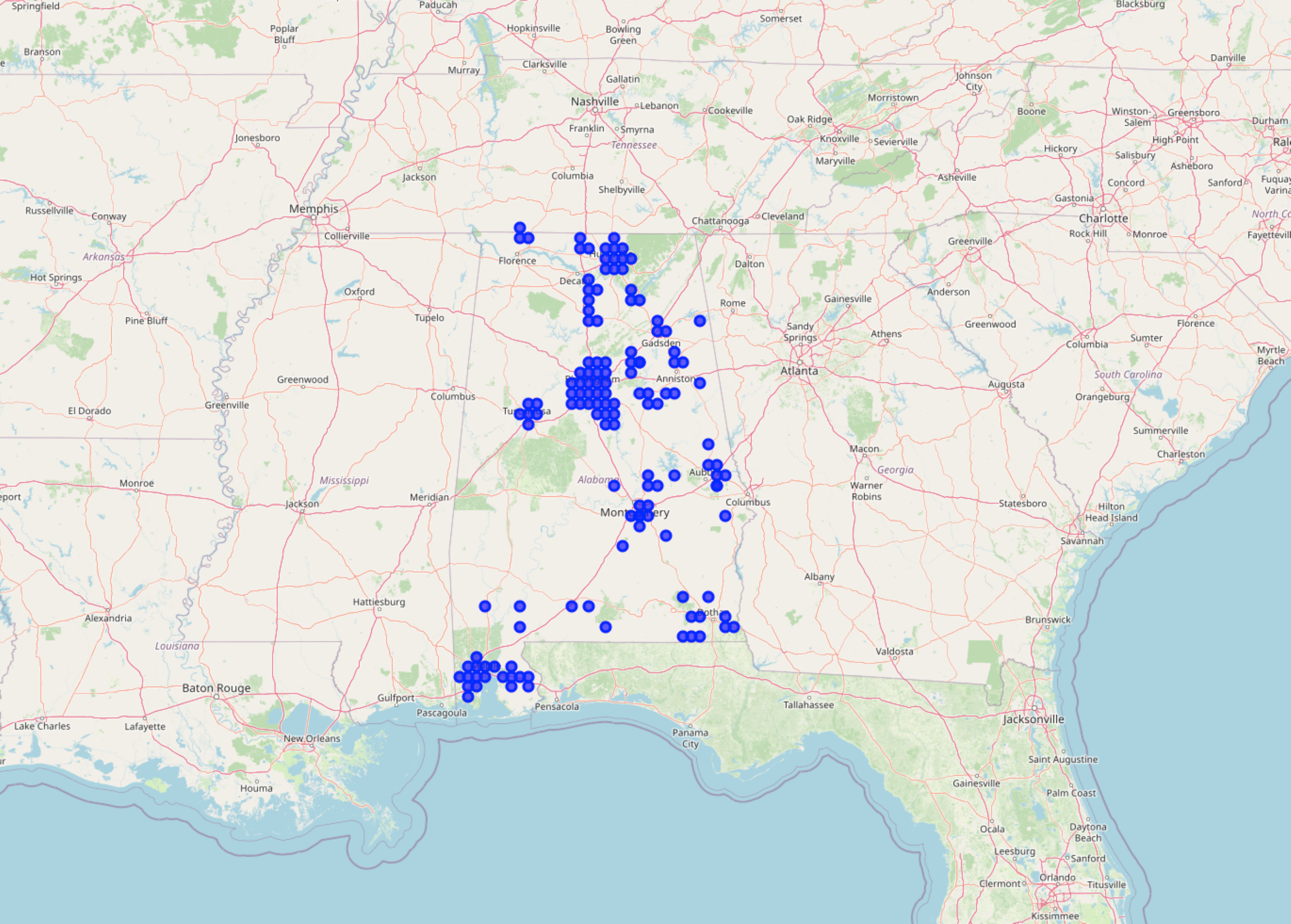}} 
    \caption{Visualisations of sensor distribution.}
    \label{app:fig:loc_vis}
\end{figure*}

%\subsection{C.2~Overall Results}

\paragraph{Implementation Details.} For datasets with temporally missing values, we first apply interpolation to obtain complete data, which is then used for training, validation, and testing. Since STSM does not report results on METR-LA and NREL, we follow the default hyperparameter settings from their paper. For NREL, due to missing regional information, we replace STSM’s selective masking with random subgraph masking. For shared model components, we adopt the same architectures and reuse STSM's reported hyperparameters to ensure fair comparison.

For our own hyperparameters, $sp=5$, $cg=2$ and $T_w=12$ are determined using the LLM-based variant on PEMS08, and \textit{the same values are applied across all datasets and model variants} without further tuning. We set $\mu = 1$ for all models on traffic datasets, and set $\mu=0.1$ on the solar power dataset.
We set $\theta = 5$ for the LLM-based model variant (\model-L) on all datasets, except for PEMS08 and Melbourne, where $\theta = 1$. For the GeoHash-based model variant (\model-H), we use $\theta = 0.01$ on PEMS07 and METR-LA, $\theta = 0.05$ on PEMS-Bay and PEMS08, and $\theta = 0.1$ on Melbourne. The observed differences in $\theta$ between \model-L and \model-H can be attributed to their distinct spatial embeddings, which yield different scales of physics residuals, e.g., on PEMS07, the 75th percentile is around 2 for \model-L and 84 for \model-H at the first training epoch.
We set $\tau = 1.0$ on PEMS08 for both model variants and for the LLM-based variant on all datasets, except for PEMS-Bay and PEMS07, where $\tau=0.85$. For the  GeoHash-based variant, we use $\tau=0.8$ on PEMS-Bay, $\tau=0.9$ on PEMS07 and Melbourne, and $\tau=0.95$ on MTRE-LA.
A GeoHash string length of 8 is used for generating SE-H, except on Melbourne (which has a smaller region and a more dense sensor distribution), where the length is set to 9.

We adopt mean square error (RMSE), mean absolute error (MAE), mean absolute percentage error (MAPE) and R-Square ($R^2$) for evaluation: 
{\small
\begin{equation}
\begin{aligned}
    \text{RMSE} &= \sqrt{ \frac{1}{n} \sum_{i=1}^{n} (\hat{x}_i - x_i)^2 }, \quad 
    \text{MAE} = \frac{1}{n} \sum_{i=1}^{n} |\hat{x}_i - x_i|\\
    \text{MAPE} &= \frac{100\%}{n} \sum_{i=1}^{n} \left| \frac{\hat{x}_i - x_i}{x_i} \right|, \quad
    R^2 = 1 - \frac{\sum_{i=1}^{n} (\hat{x}_i - x_i)^2}{\sum_{i=1}^{n} (x_i - \bar{x})^2}\\
\end{aligned}
\end{equation}
}
where, $\hat{x}$ and $x$ denote the forecasted values and ground-truth values, respectively; $\bar{x} = \frac{1}{n} \sum_{i=1}^{n} x_i$, and $n$ denotes the number of samples. 

\begin{table*}[!t]
    \centering
    {\small
    \begin{tabular}{llccccc}
        \hlineB{3}
        \textbf{Model} & \textbf{Time} & \textbf{PEMS-Bay} & \textbf{PEMS07} & \textbf{PEMS08} & \textbf{Melbourne} & \textbf{METR} \\
        \hline \hline
        \multirow{2}{*}{GE-GAN} 
            & Train (h) & 4.4 & 4.1 & 4.1 & 0.3 & 3.4 \\
            & Test (s)  & 0.9 & 0.7 & 0.8 & 0.1 & 0.3 \\
            \hline
        \multirow{2}{*}{IGNNK} 
            & Train (h) & 0.3 & 0.2 & 0.2 & 0.1 & 0.2 \\
            & Test (s)  & 8.3 & 7.8 & 8.8 & 2.4 & 2.6 \\
            \hline
        \multirow{2}{*}{INCREASE} 
            & Train (h) & 0.3 & 0.2 & 0.2 & 0.2 & 0.1 \\
            & Test (s)  & 9.5 & 7.3 & 7.5 & 4.0 & 5.2 \\
            \hline
        \multirow{2}{*}{STSM} 
            & Train (h) & 1.1 & 1.9 & 2.2 & 0.3 & 1.0 \\
            & Test (s)  & 1.6 & 1.3 & 1.2 & 0.3 & 2.0 \\
            \hline
        \multirow{2}{*}{KITS} 
            & Train (h) & 1.0 & 1.5 & 1.2 & 0.2 & 1.1 \\
            & Test (s)  & 1.4 & 1.4 & 1.4 & 0.2 & 0.6 \\
            \hline
        \multirow{2}{*}{\model-H} 
            & Train (h) & 3.0 & 2.6 & 2.2 & 2.9 & 1.4 \\
            & Test (s)  & 3.6 & 2.7 & 2.6 & 1.3 & 2.3 \\
            \hline
        \multirow{2}{*}{\model-L} 
            & Train (h) & 3.0 & 2.1 & 1.9 & 2.1 & 1.3 \\
            & Test (s)  & 3.4 & 2.4 & 2.3 & 1.2 & 2.1 \\
        \hlineB{3}
    \end{tabular}
    }
    \caption{Training and inference time comparison.}
\end{table*}

\subsection{C.2~Running Time Results}
We report training and inference times across all traffic datasets. While our model incurs slightly longer training time due to its richer architecture, all models remain within a comparable time scale. Importantly, inference speed of \model, which matters most in deployment, is on par with other methods. Given that training is conducted offline, the small increase in training time is a reasonable cost. Considering the consistent improvements in generalisation performance, this trade-off is practically justified.

\begin{figure*}[!t]
    \centering
    \subfigure[PEMS08]{\includegraphics[width=0.3\linewidth]{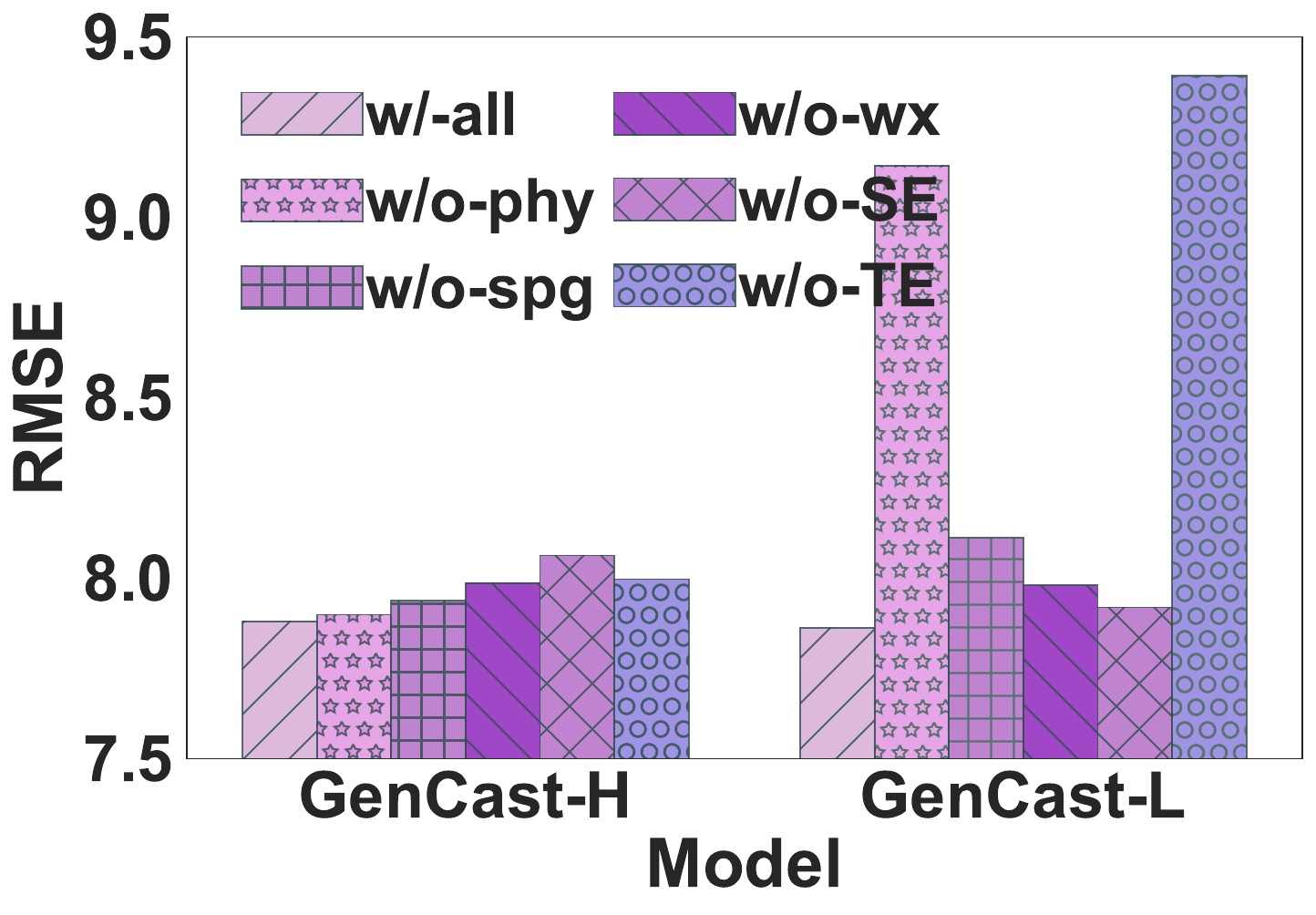}} 
    \hspace{-2mm}
    \subfigure[METR-LA]{\includegraphics[width=0.3\linewidth]{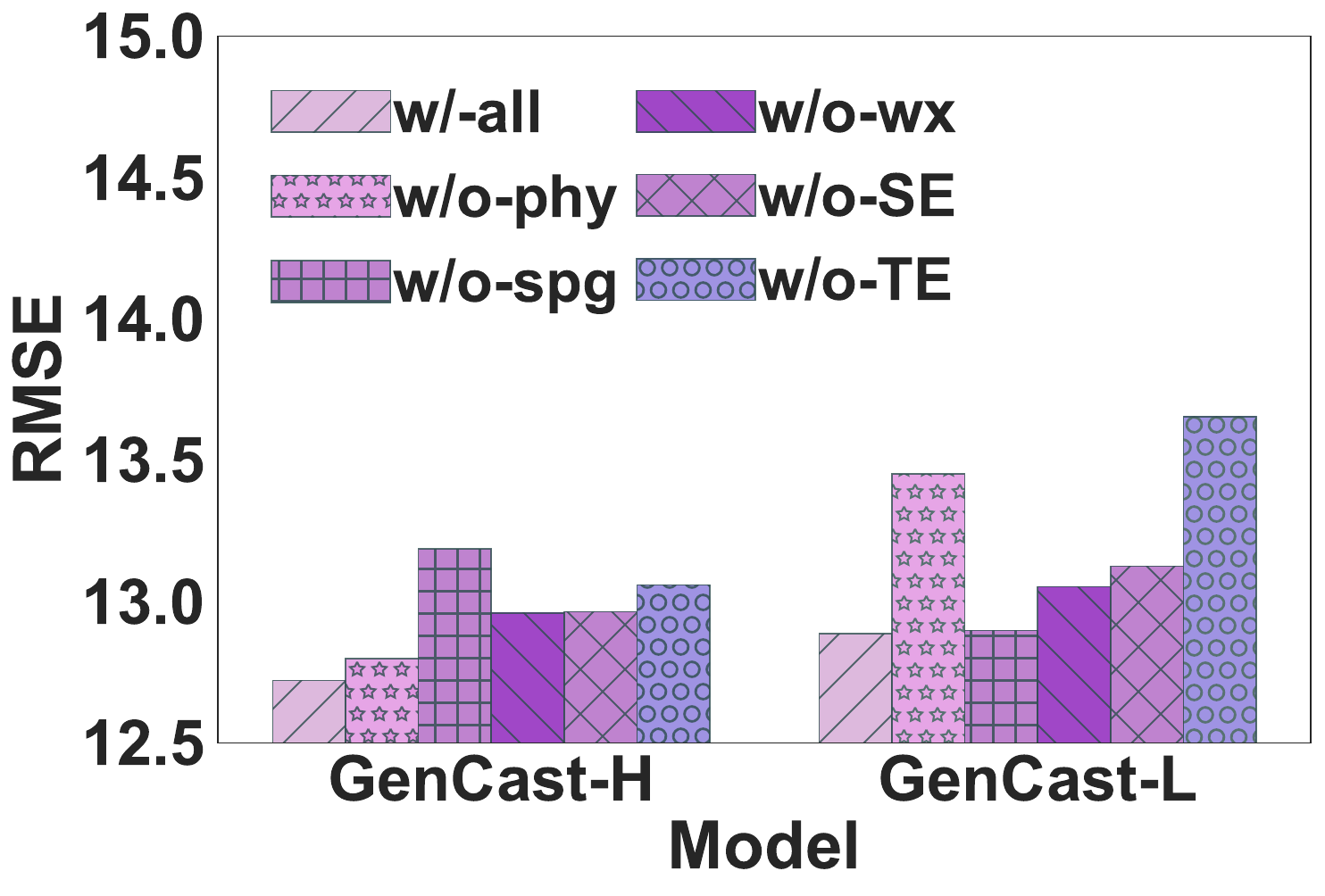}} 
    \hspace{-2mm}
    \subfigure[Melbourne]{\includegraphics[width=0.3\linewidth]{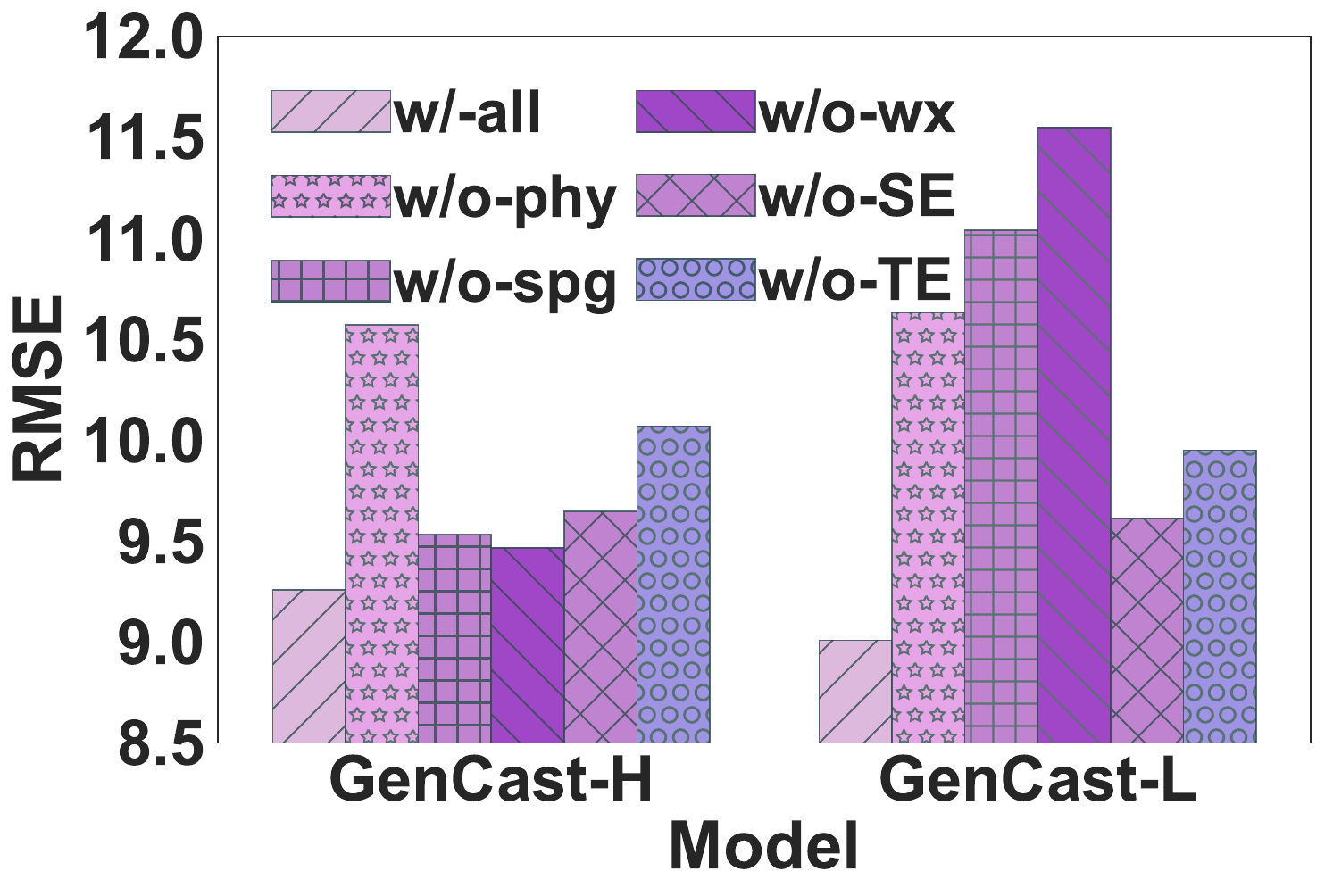}} 
    \caption{Ablation study results.}
    \label{app:fig:ablation}
\end{figure*}

\subsection{C.3 Ablation Study}
\paragraph{Effectiveness of all components.} We compare \model\ with five variants: \textbf{w/o-phy}, \textbf{w/o-spg}, and \textbf{w/o-wx}  remove the physics constraint, spatial grouping loss, and external signal (i.e., weather) encoder, respectively; \textbf{w/o-SE} and \textbf{w/o-TE} remove spatial and temporal embeddings, respectively, together with the physics constraint. Fig.~\ref{app:fig:ablation} presents ablation results on PEMS08, METR-LA, and Melbourne. All modules in \model\ contribute to its performance gains, as verified by the higher errors of all model variants. As observed on PEMS07 and PEMS-Bay (Fig.~\ref{fig:ablation} in the main paper), the physics constraint is especially important for \model-L. Notably, it is also crucial for \model-H on the Melbourne dataset. This is because Melbourne is an urban dataset with densely located sensors (see Fig.~\ref{app:fig:loc_vis}~(e)), making it more difficult to differentiate by the spatial embeddings, even with SE-H. In such settings, the physics-guided loss becomes essential for helping the model effectively utilise spatial embeddings.

Additionally, {w/o-spg} and {}w/o-wx have similar impacts across datasets, showing their general  applicability. On Melbourne, \model-L is more sensitive to the removal of the key modules, due to the high sensor density and urban spatial layout as mentioned above.

\paragraph{Effectiveness of Huber Loss.}
Table~\ref{tab:huber_comparison} compares model performance using dynamic Huber loss and fixed L2 loss (i.e., Huber with $\delta$ set to the 100\%-quantile) on the physics residual (Eq.~\ref{eq:phy_residual}). When $\delta$ is large, Huber loss behaves like the L2 loss, treating all residual values equally. We observe that the distribution of physics residuals varies across datasets -- some may contain more outliers or  heavier tails. This motivates our adaptive calibration strategy, where $\delta$ is set based on a quantile of the initial error distribution during a warm-up stage. This lightweight adjustment improves robustness to outliers while retaining sensitivity to typical errors.
Empirically, this strategy yields consistent (even though somewhat small) improvements across datasets and metrics. While simple, the method enhances training stability and generalisation with negligible overhead.

\begin{table*}[!t]
\centering
\small
\begin{tabular}{l|l|c|c|c|c}
\hlineB{3}
\textbf{Dataset} & \textbf{Metric} & \textbf{\model-H} & \textbf{\model-H (L2)} & \textbf{\model-L} & \textbf{\model-L (L2)} \\
\hline\hline
\multirow{4}{*}{PEMS-07} 
& RMSE & 8.285 & 8.425 & 8.253 & 8.311 \\
& MAE  & 5.116 & 5.286 & 5.073 & 5.024 \\
& MAPE & 0.122 & 0.125 & 0.121 & 0.122 \\
& R$^2$ & 0.193 & 0.166 & 0.197 & 0.189 \\
\hline
\multirow{4}{*}{PEMS-08} 
& RMSE & 7.880 & 7.880 & 7.863 & 7.863 \\
& MAE  & 4.776 & 4.776 & 4.728 & 4.728 \\
& MAPE & 0.113 & 0.113 & 0.112 & 0.112 \\
& R$^2$ & 0.146 & 0.146 & 0.150 & 0.150 \\
\hline
\multirow{4}{*}{METR-LA} 
& RMSE & 12.720 & 12.953 & 12.886 & 12.886 \\
& MAE  & 8.792  & 8.983  & 8.799  & 8.799  \\
& MAPE & 0.265  & 0.272  & 0.267  & 0.267  \\
& R$^2$ & 0.086  & 0.051  & 0.063  & 0.063  \\
\hline
\multirow{4}{*}{Melbourne} 
& RMSE & 9.009 & 9.530 & 9.258 & 9.258 \\
& MAE  & 7.083 & 7.496 & 7.253 & 7.253 \\
& MAPE & 0.366 & 0.387 & 0.370 & 0.370 \\
& R$^2$ & 0.061 & -0.052 & 0.012 & 0.012 \\
\hline
\multirow{4}{*}{PEMS-Bay} 
& RMSE & 8.683 & 8.799 & 8.692 & 8.752 \\
& MAE  & 5.192 & 5.193 & 5.139 & 5.143 \\
& MAPE & 0.131 & 0.132 & 0.131 & 0.132 \\
& R$^2$ & 0.228 & 0.207 & 0.225 & 0.215 \\
\hlineB{3}
\end{tabular}
\caption{Model Performance: dynamic Huber Loss vs. L2 loss on the physics residual.}
\label{tab:huber_comparison}
\end{table*}

\subsection{C.4~Parameter and Case Study}

\paragraph{Impact of Unobserved Ratio.}
Fig.~\ref{app:fig:unknow_ratio} presents model performance when we vary the unobserved ratio (i.e., percentage of unobserved nodes in $G$) from 0.2 to 0.8 on the PEMS08, METR-LA and Melbourne datasets. As before, we split each dataset either horizontally or vertically and report the performance averaged over four setups. We plot the top-3 best baselines with our models. Our \model, with either variant \model-H (using GeoHash embeddings) or \model-L (using LLM embeddings), performs the best in most cases (20 out of 21), confirming our proposed model \model's robustness against the unobserved ratio.
\begin{figure*}[!t]
    \centering
    \subfigure[PEM08]{\includegraphics[width=0.3\textwidth]{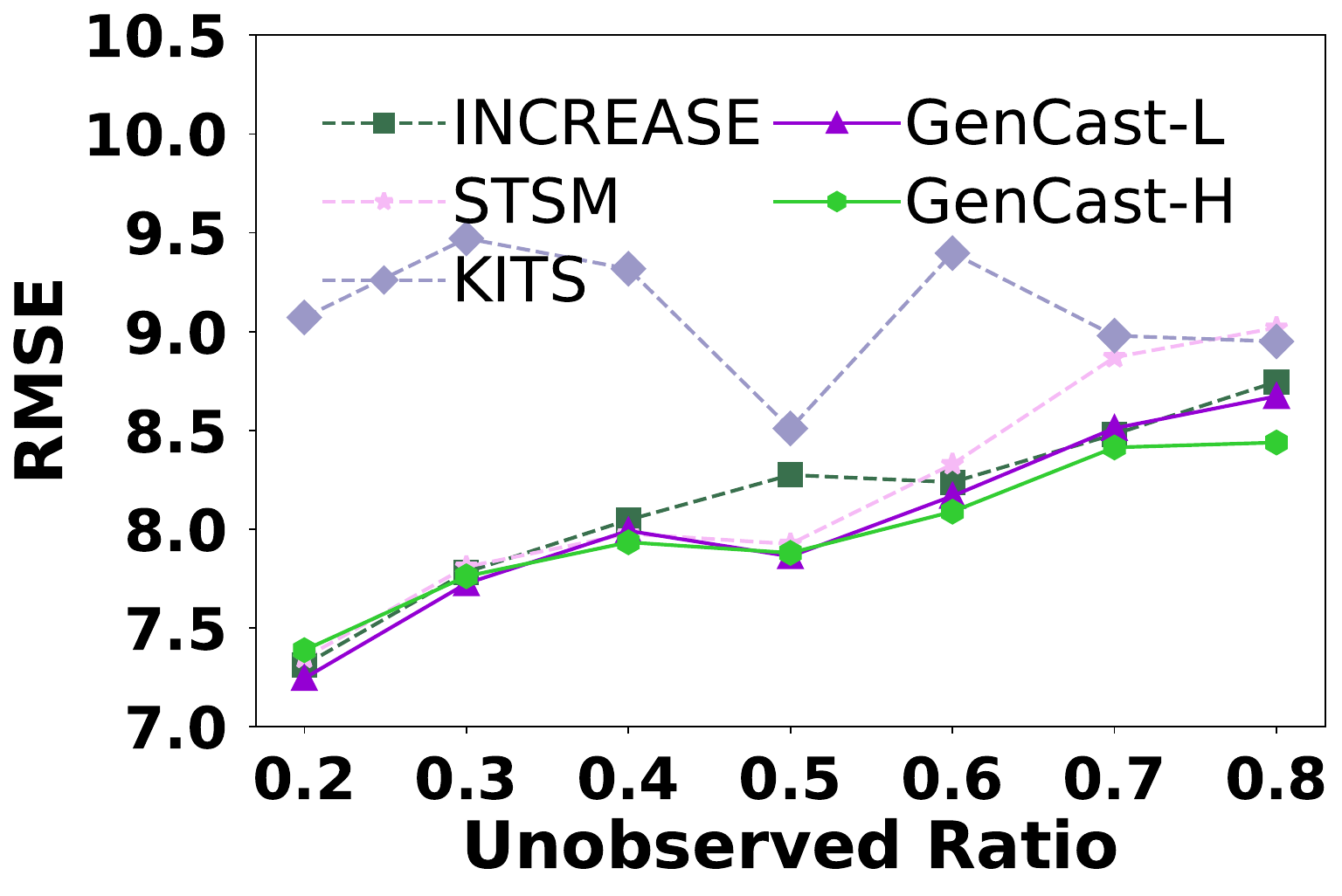}} 
    \subfigure[METR-LA]{\includegraphics[width=0.3\textwidth]{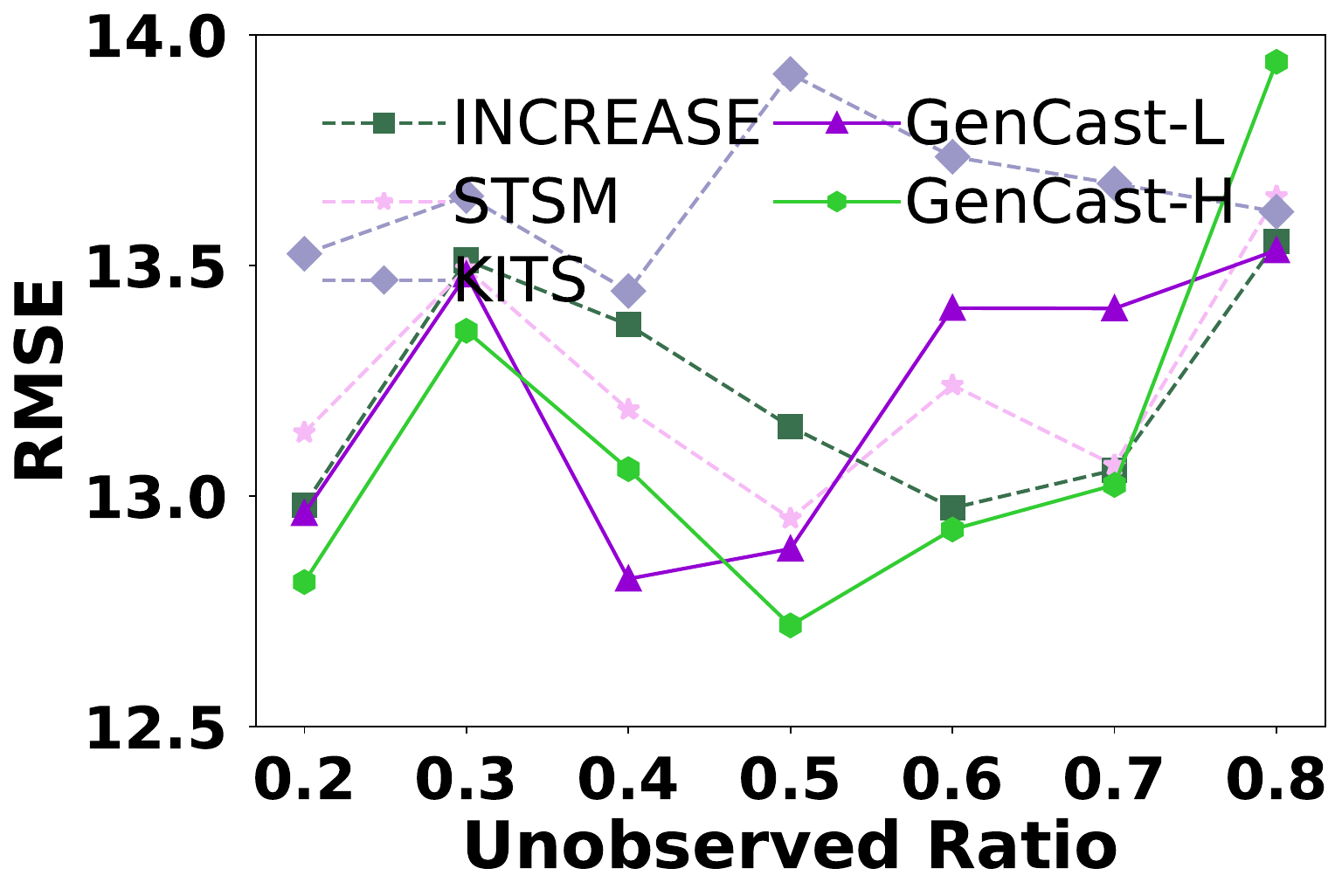}}
    \subfigure[Melbourne]{\includegraphics[width=0.3\textwidth]{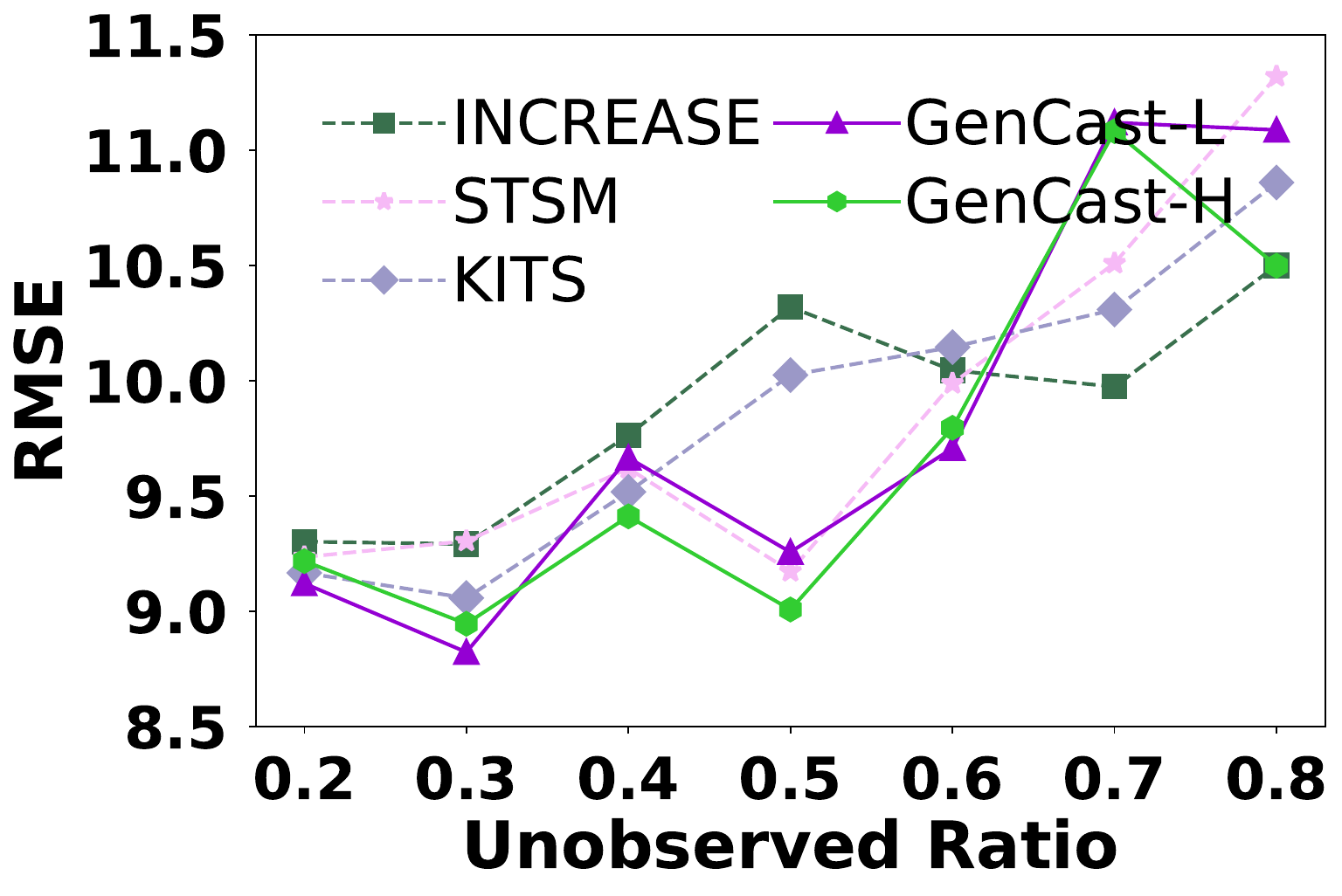}} 
    \caption{Model performance vs. unobserved ratio.}
    \label{app:fig:unknow_ratio}
\end{figure*}

\begin{figure*}[!t]
    \centering
    \subfigure[PEMS07]{\includegraphics[width=0.18\textwidth]{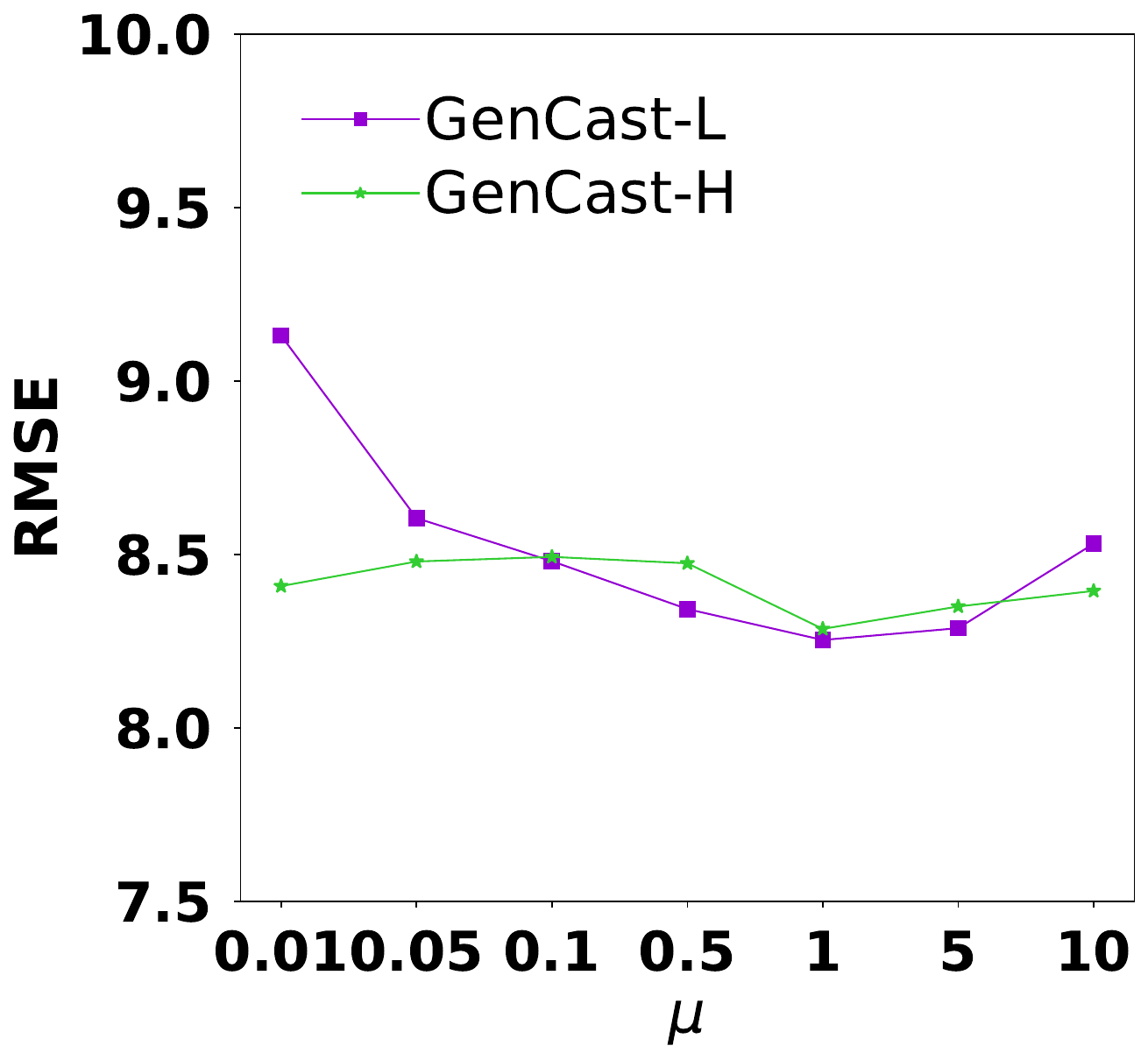}} 
    \hspace{-1mm}
    \subfigure[PEMS08]{\includegraphics[width=0.18\textwidth]{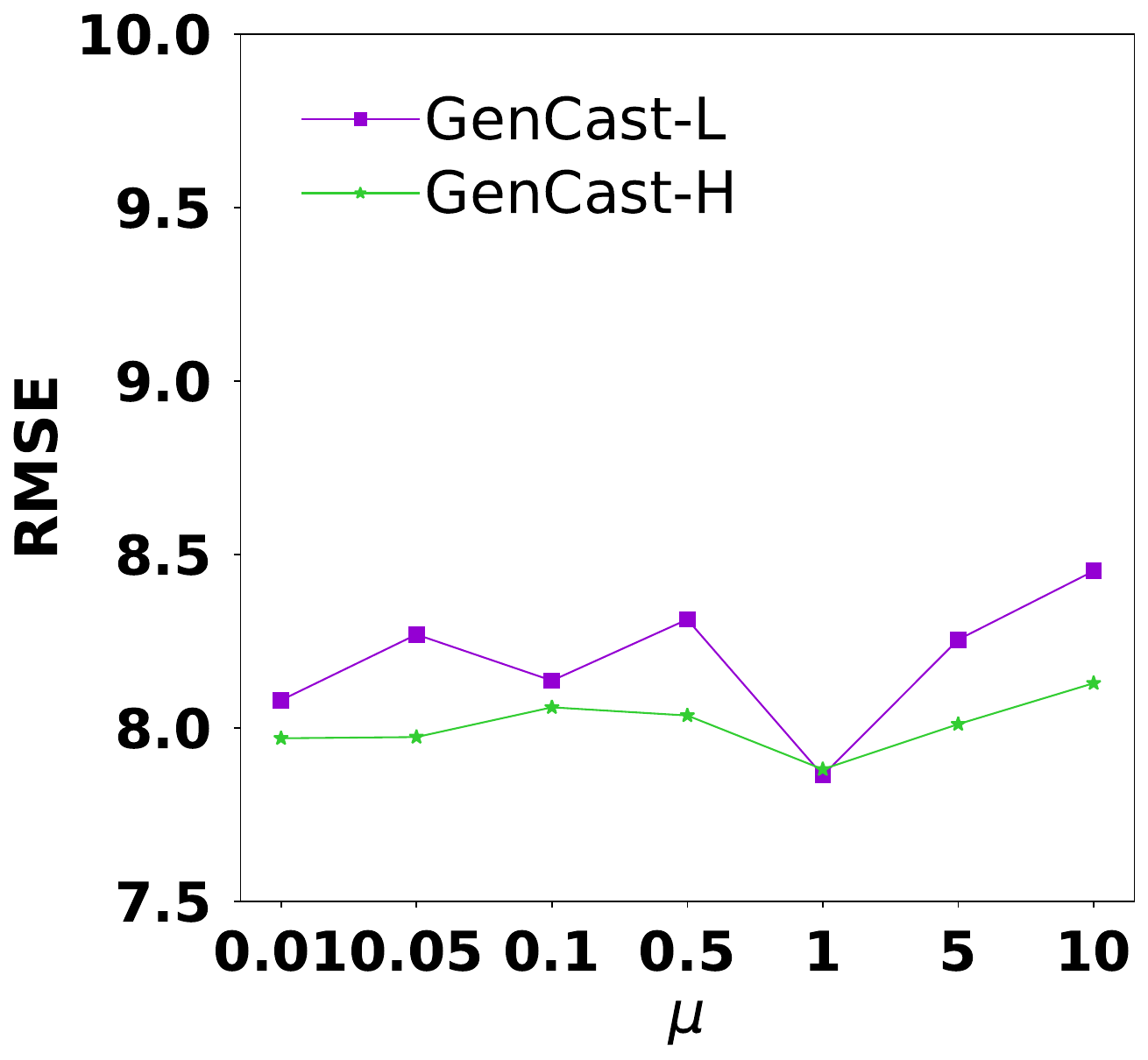}} 
    \hspace{-1mm}
    \subfigure[PEMS-Bay]{\includegraphics[width=0.18\textwidth]{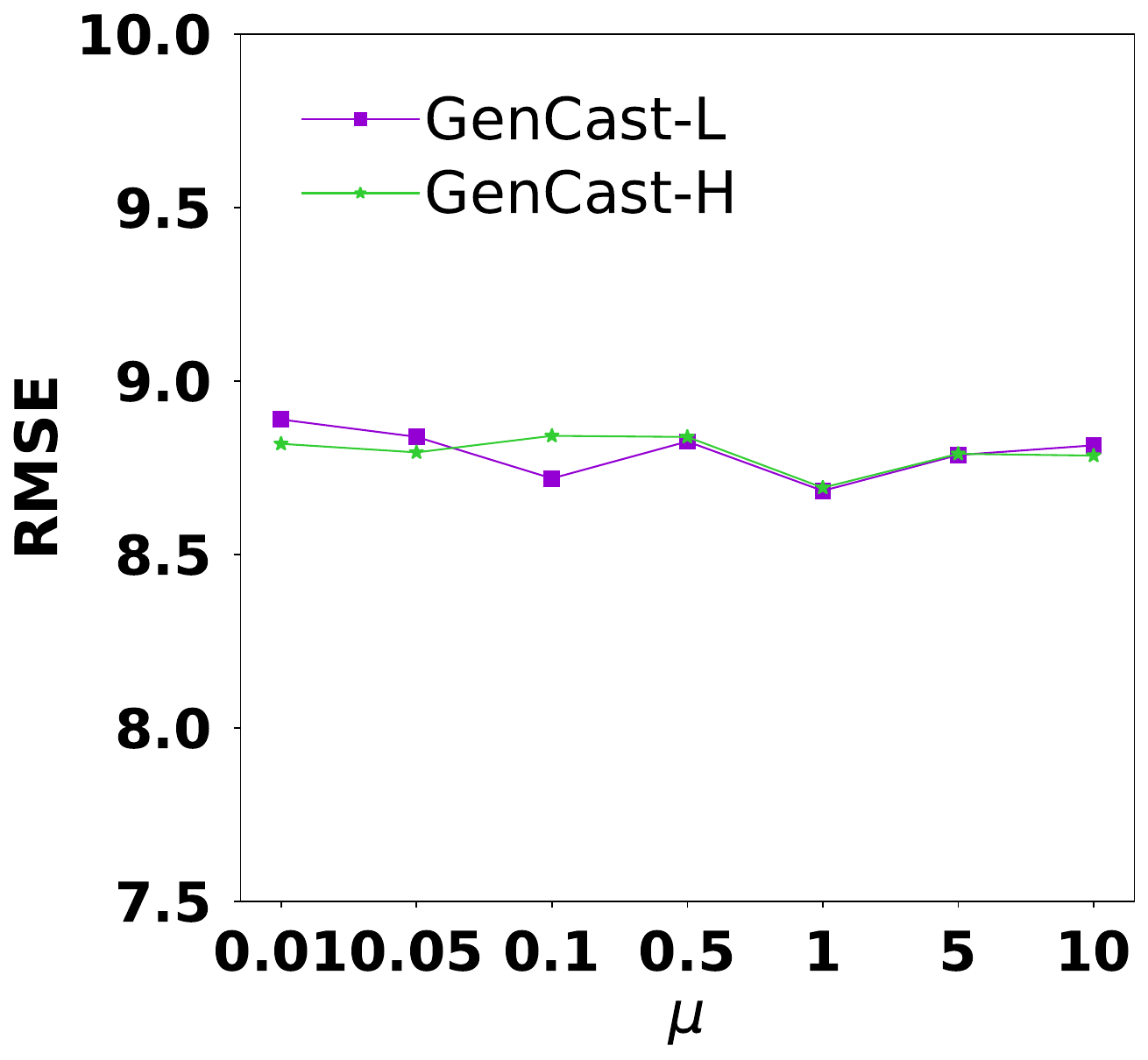}} 
    \hspace{-1mm}
    \subfigure[METR-LA]{\includegraphics[width=0.18\textwidth]{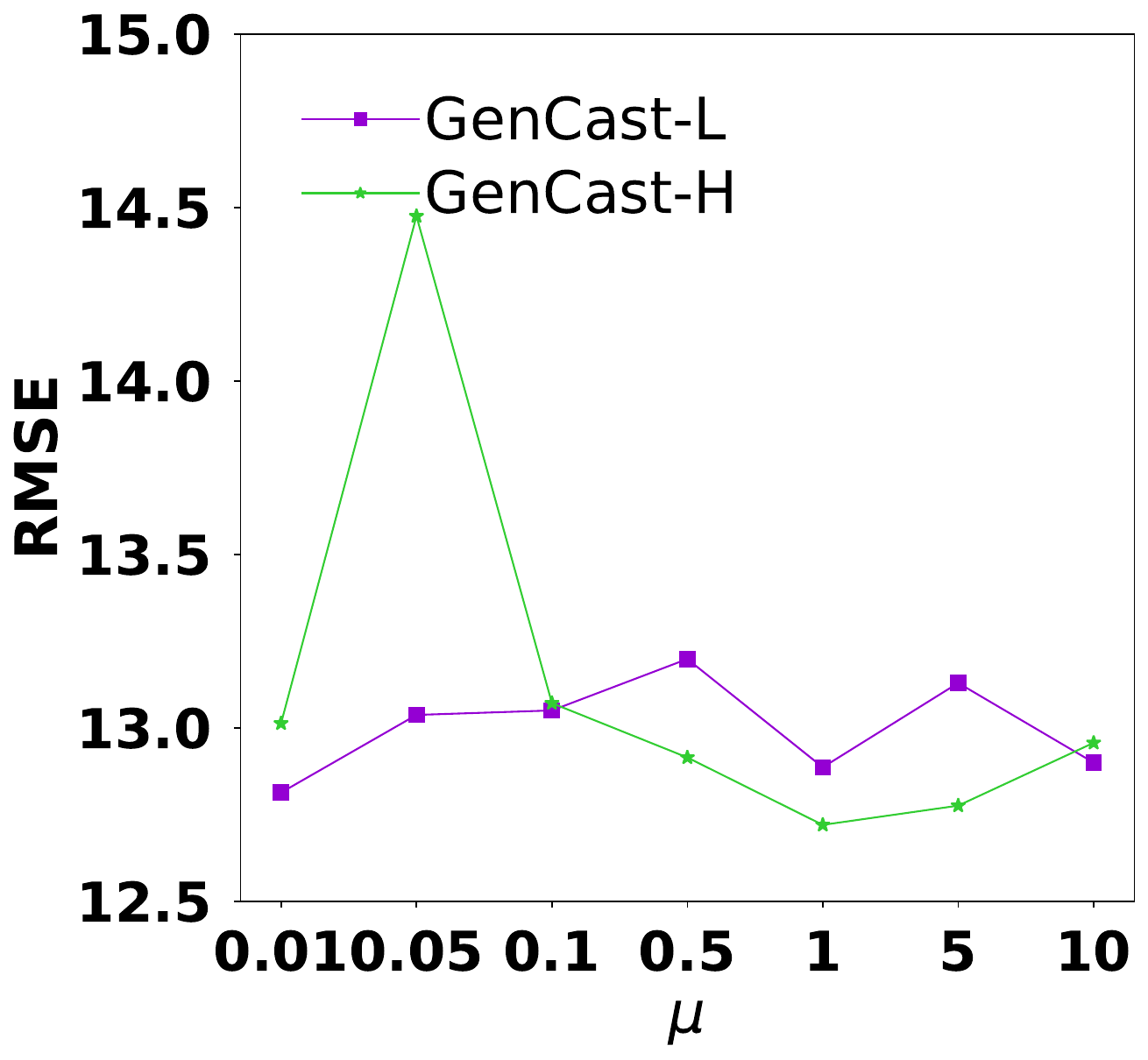}} 
    \hspace{-1mm}
    \subfigure[MEL]{\includegraphics[width=0.18\textwidth]{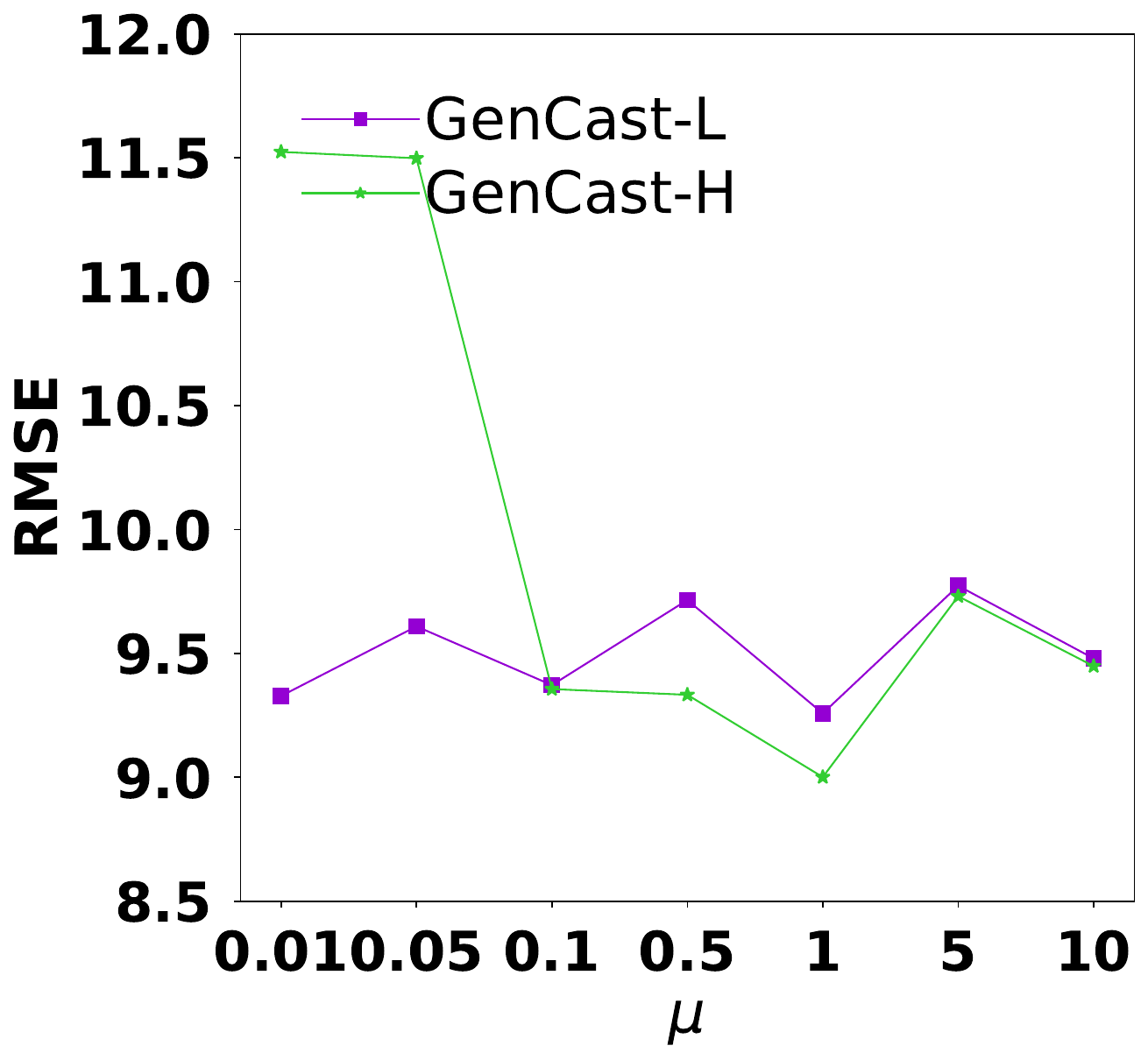}} 

    \subfigure[PEMS07]{\includegraphics[width=0.18\textwidth]{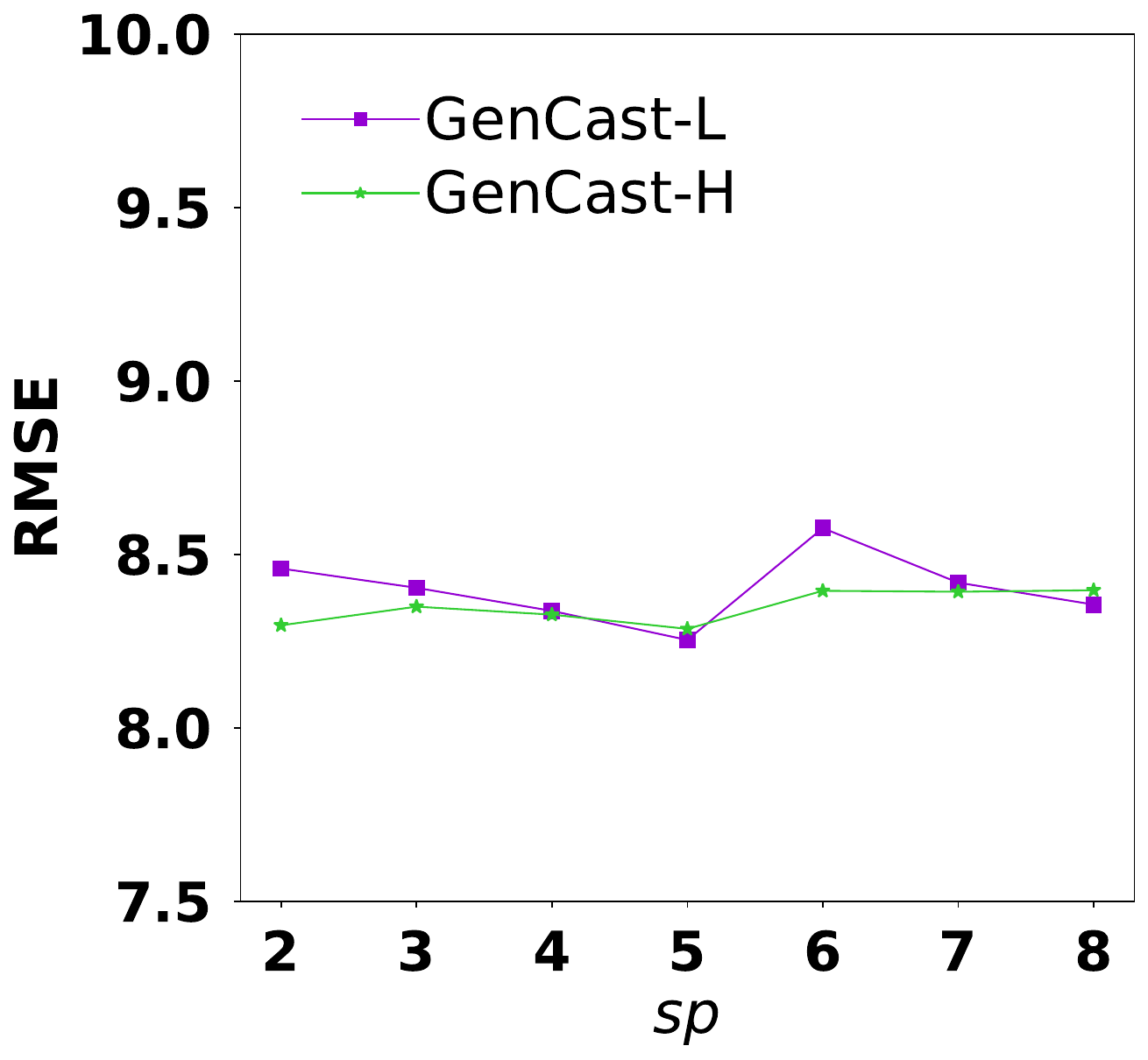}} 
    \hspace{-1mm}
    \subfigure[PEMS08]{\includegraphics[width=0.18\textwidth]{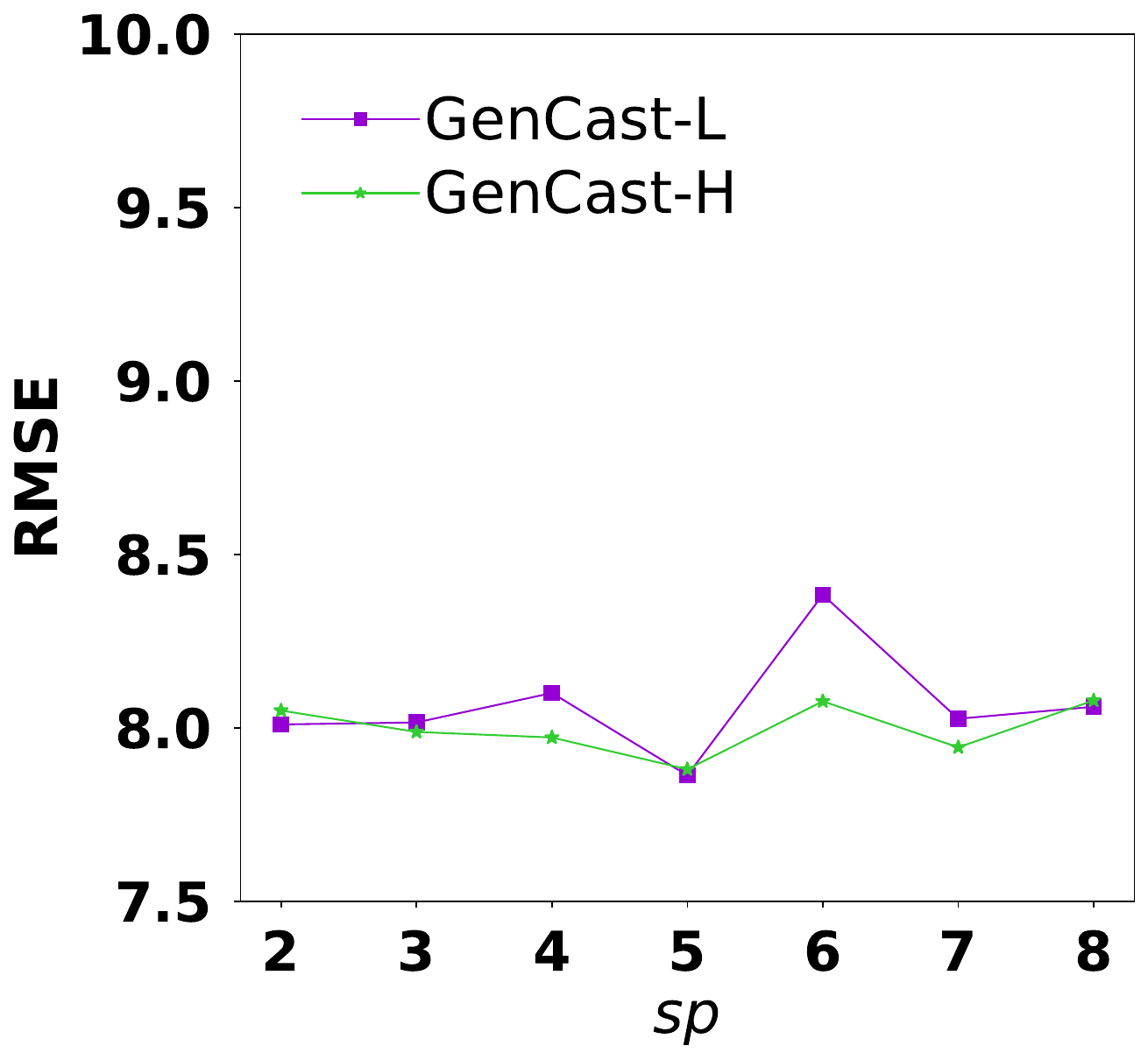}} 
    \hspace{-1mm}
    \subfigure[PEMS-Bay]{\includegraphics[width=0.18\textwidth]{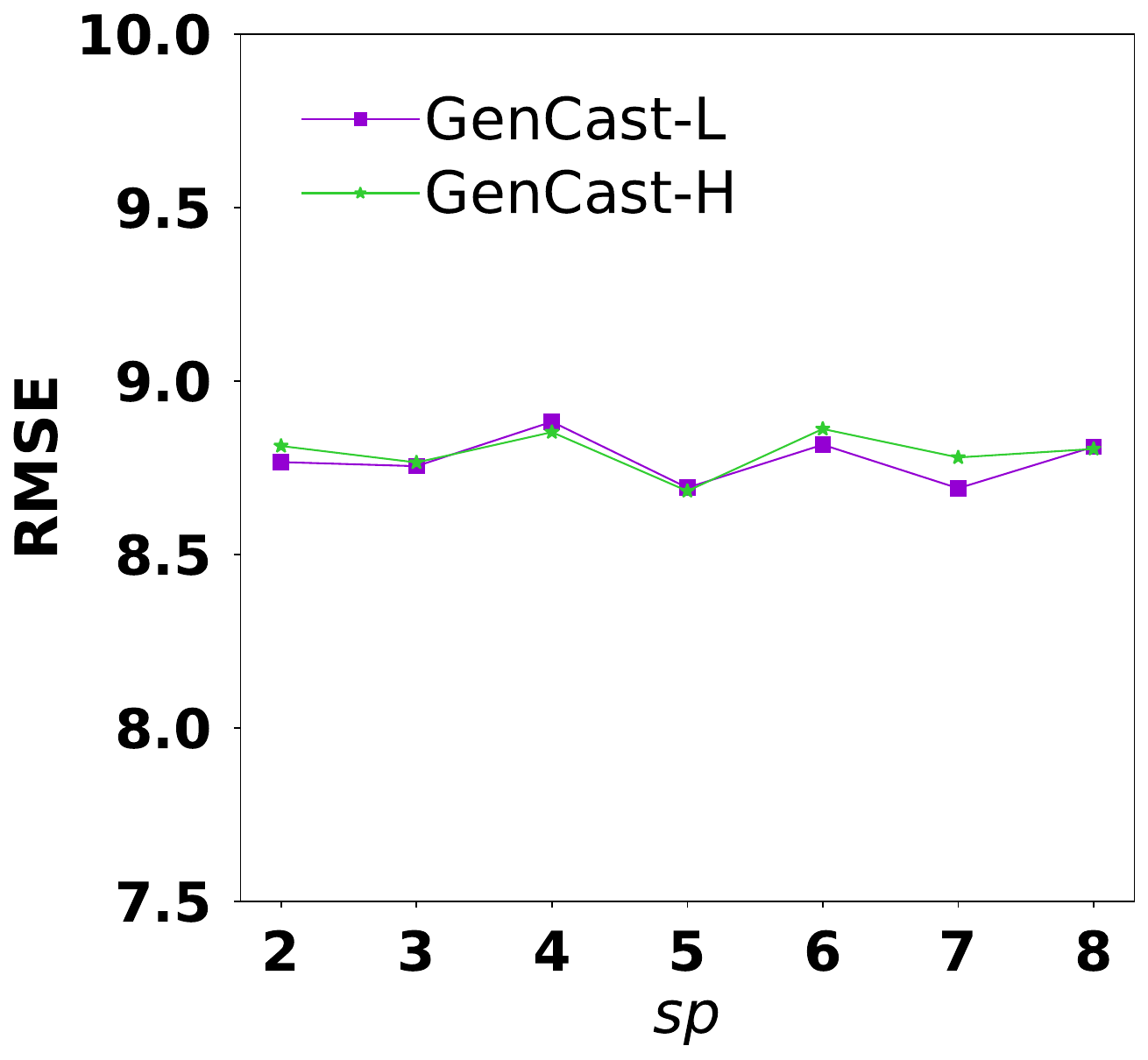}} 
    \hspace{-1mm}
    \subfigure[METR-LA]{\includegraphics[width=0.18\textwidth]{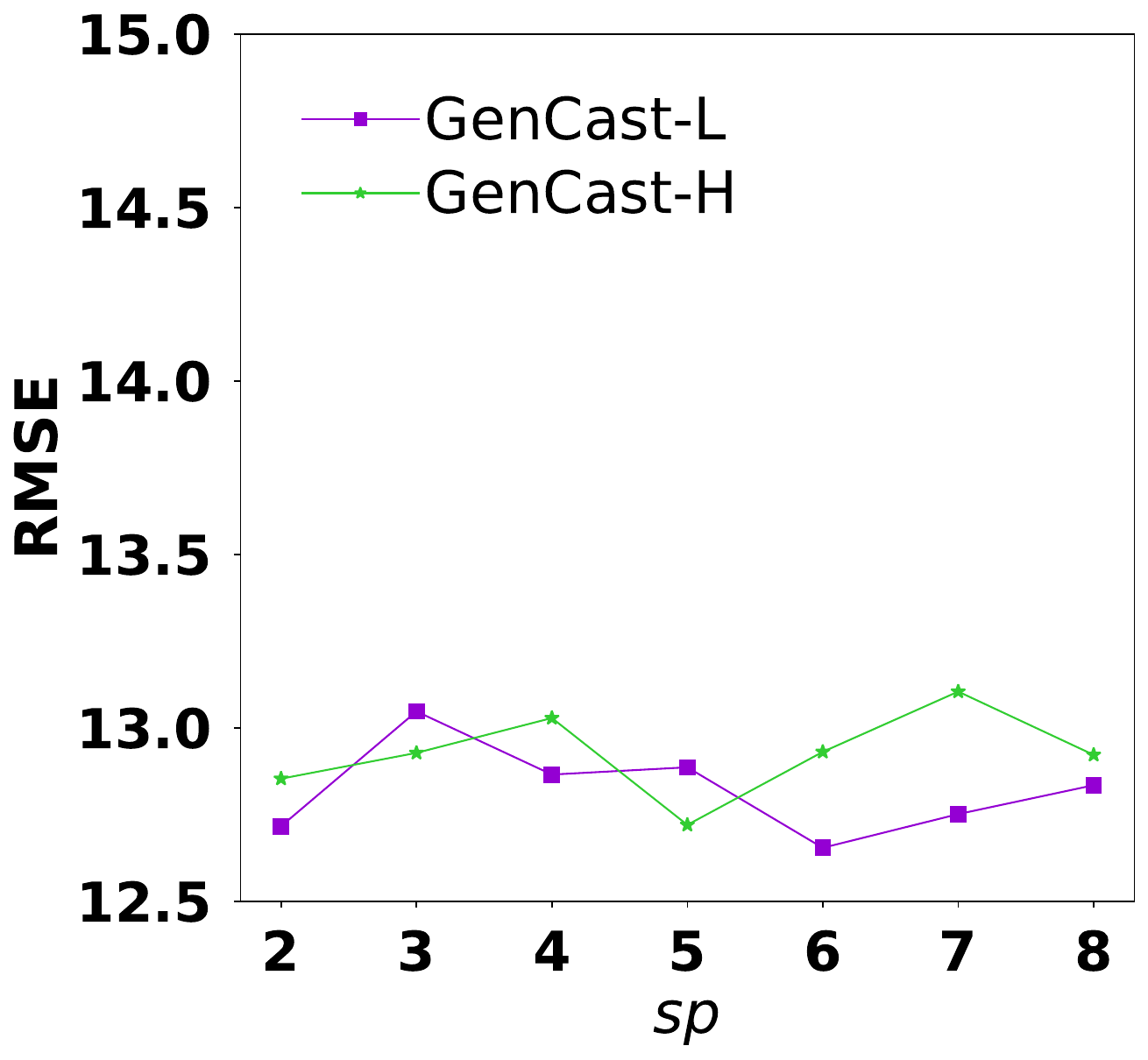}} 
    \hspace{-1mm}
    \subfigure[MEL]{\includegraphics[width=0.18\textwidth]{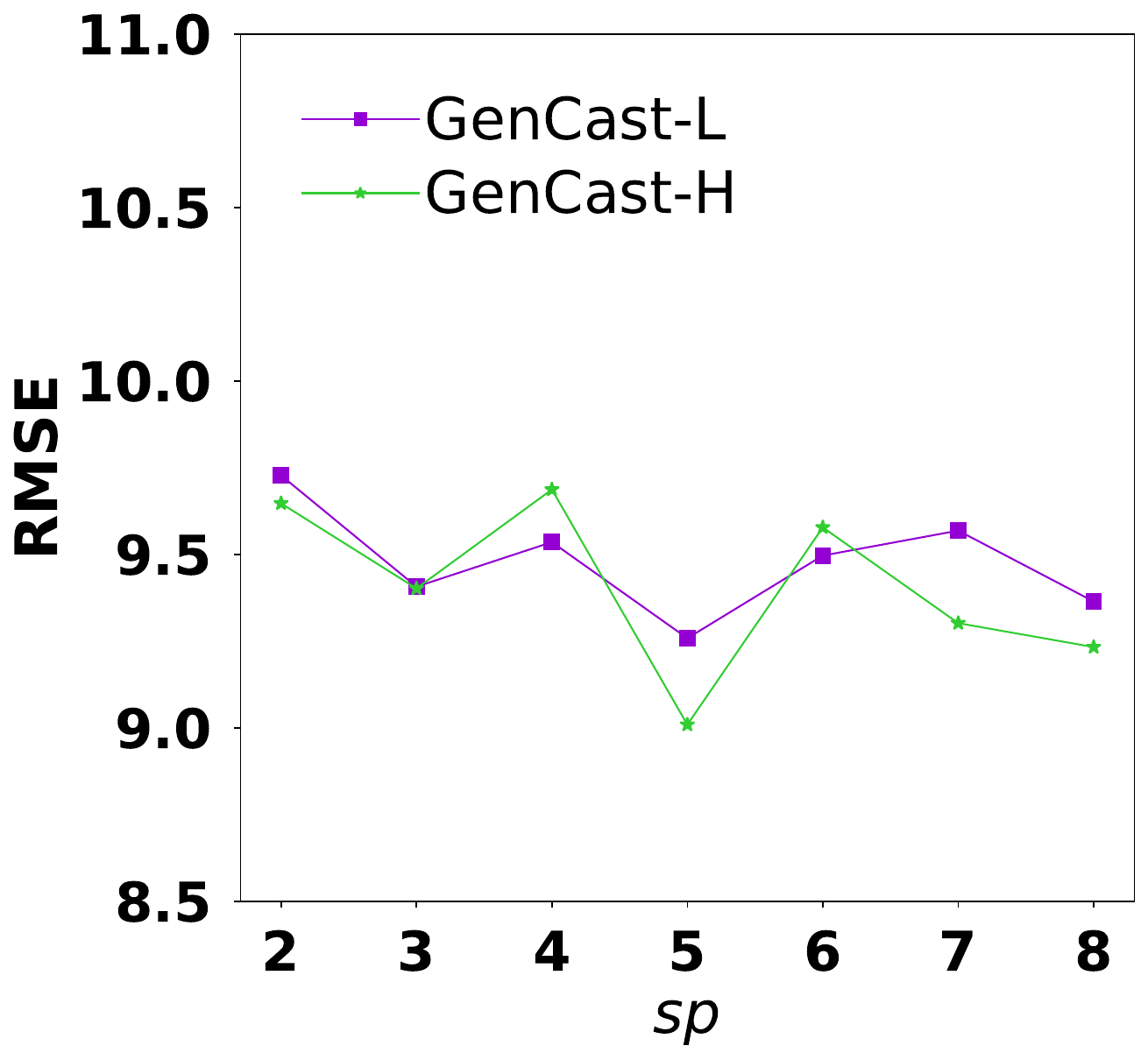}}

    \subfigure[PEMS07]{\includegraphics[width=0.18\textwidth]{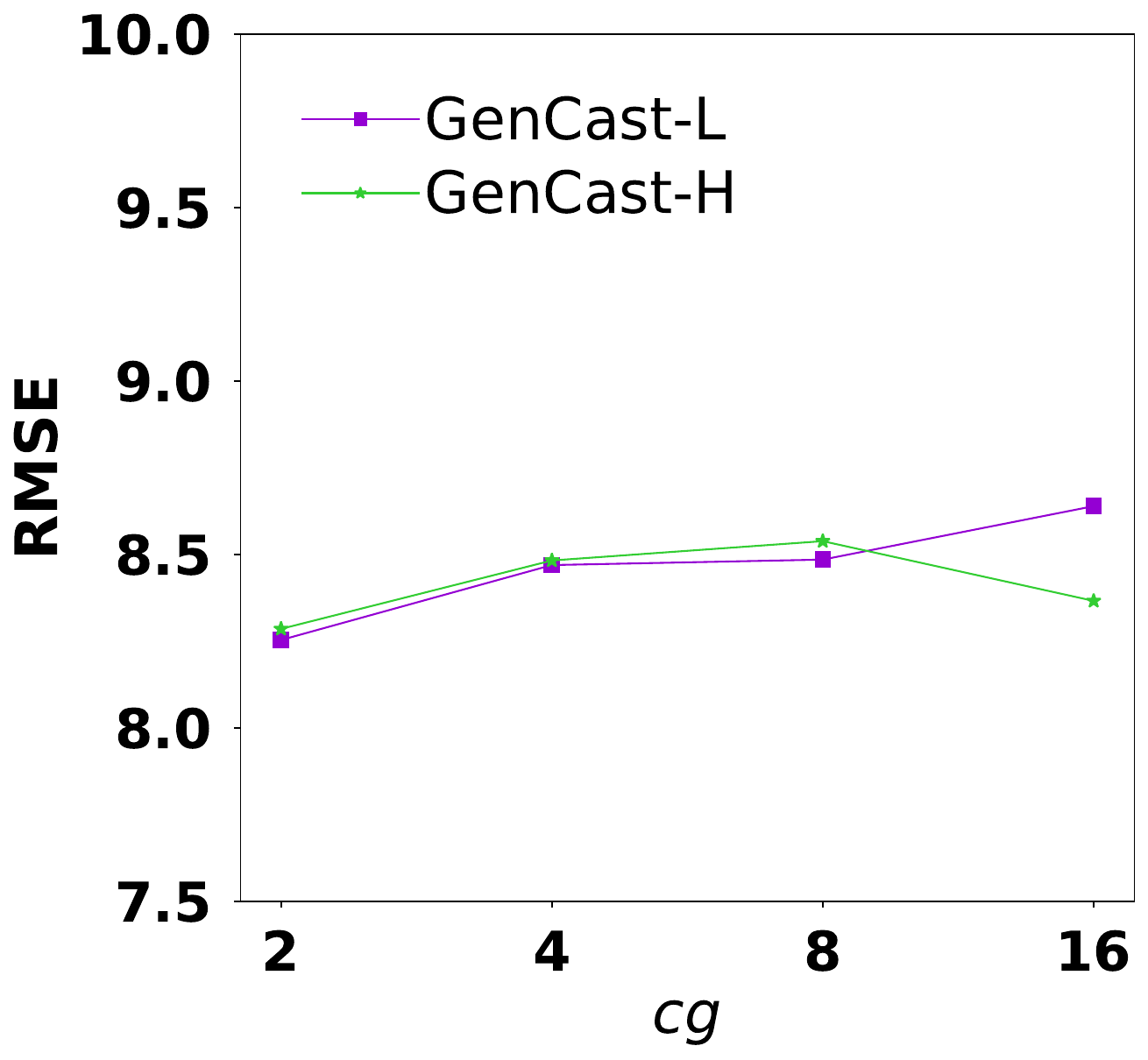}} 
    \hspace{-1mm}
    \subfigure[PEMS08]{\includegraphics[width=0.18\textwidth]{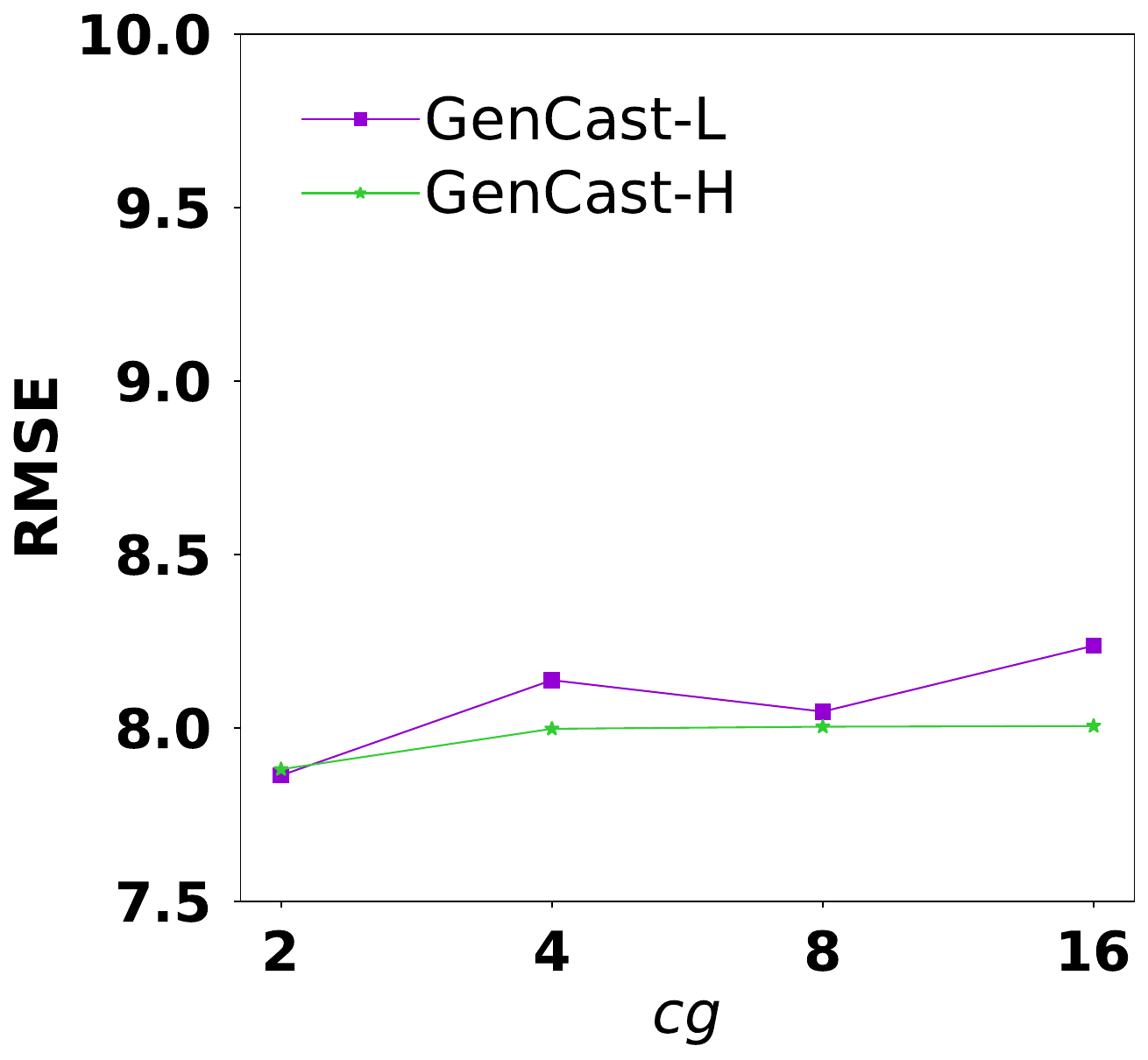}} 
    \hspace{-1mm}
    \subfigure[PEMS-Bay]{\includegraphics[width=0.18\textwidth]{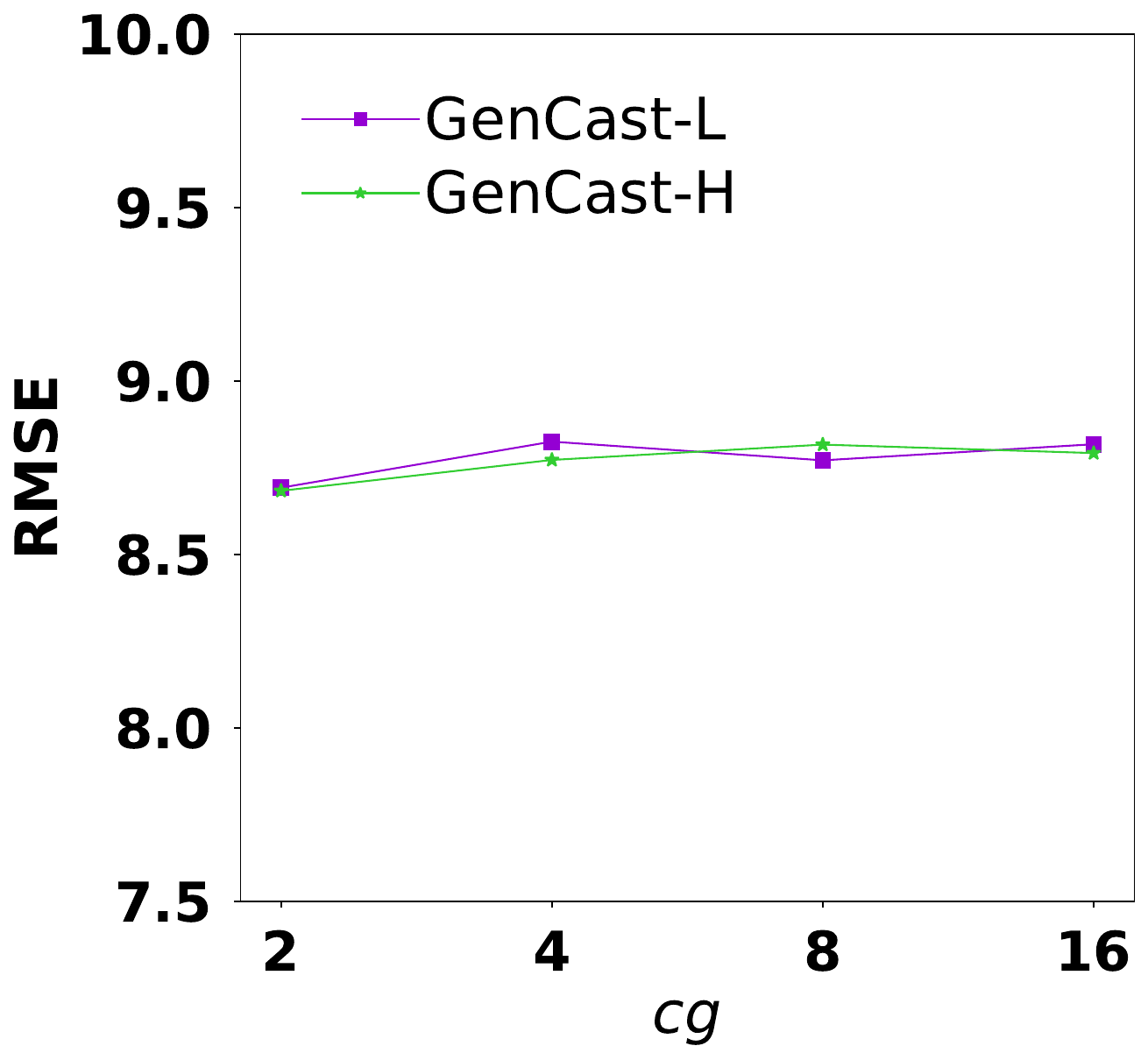}} 
    \hspace{-1mm}
    \subfigure[METR-LA]{\includegraphics[width=0.18\textwidth]{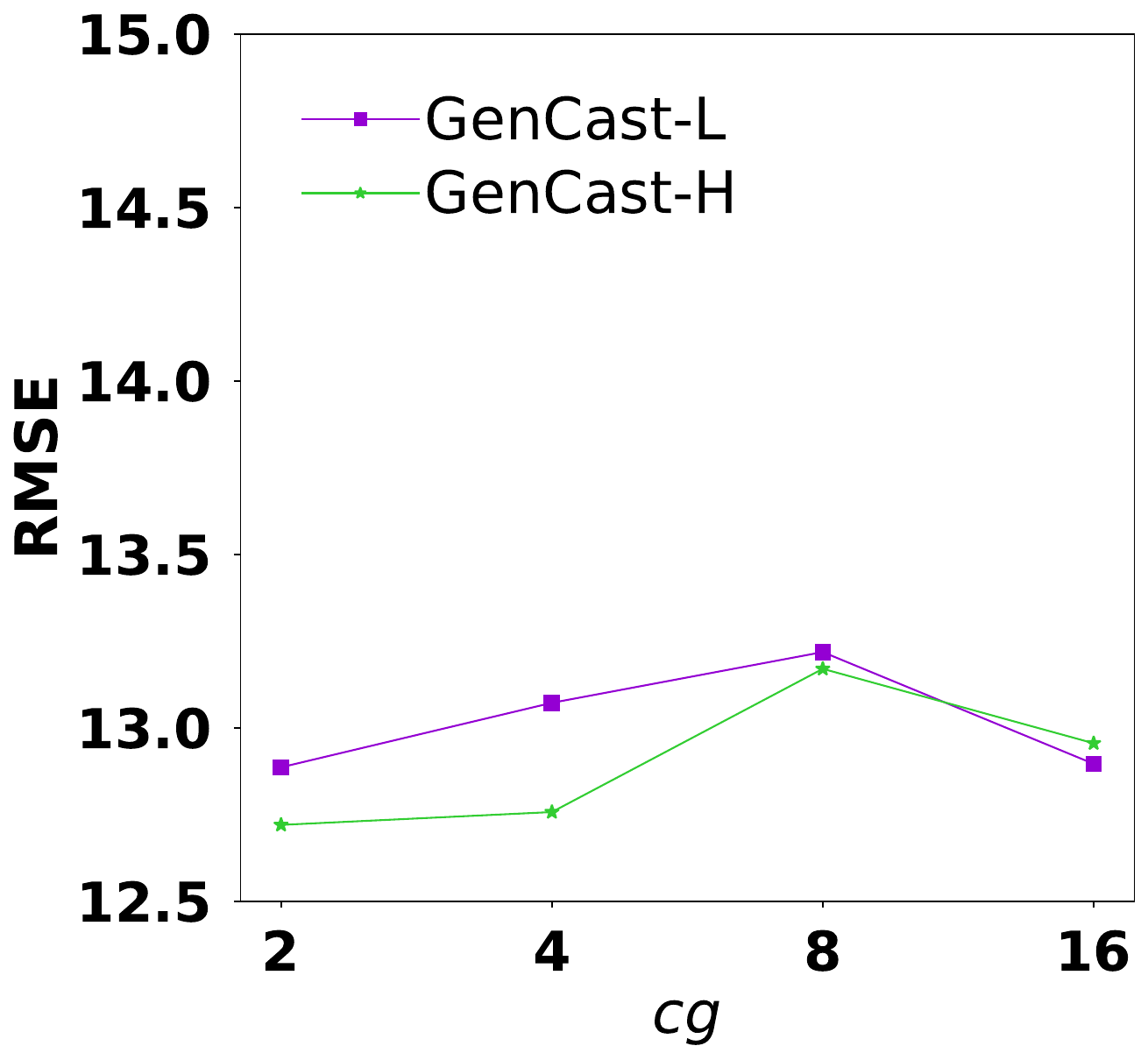}} 
    \hspace{-1mm}
    \subfigure[MEL]{\includegraphics[width=0.18\textwidth]{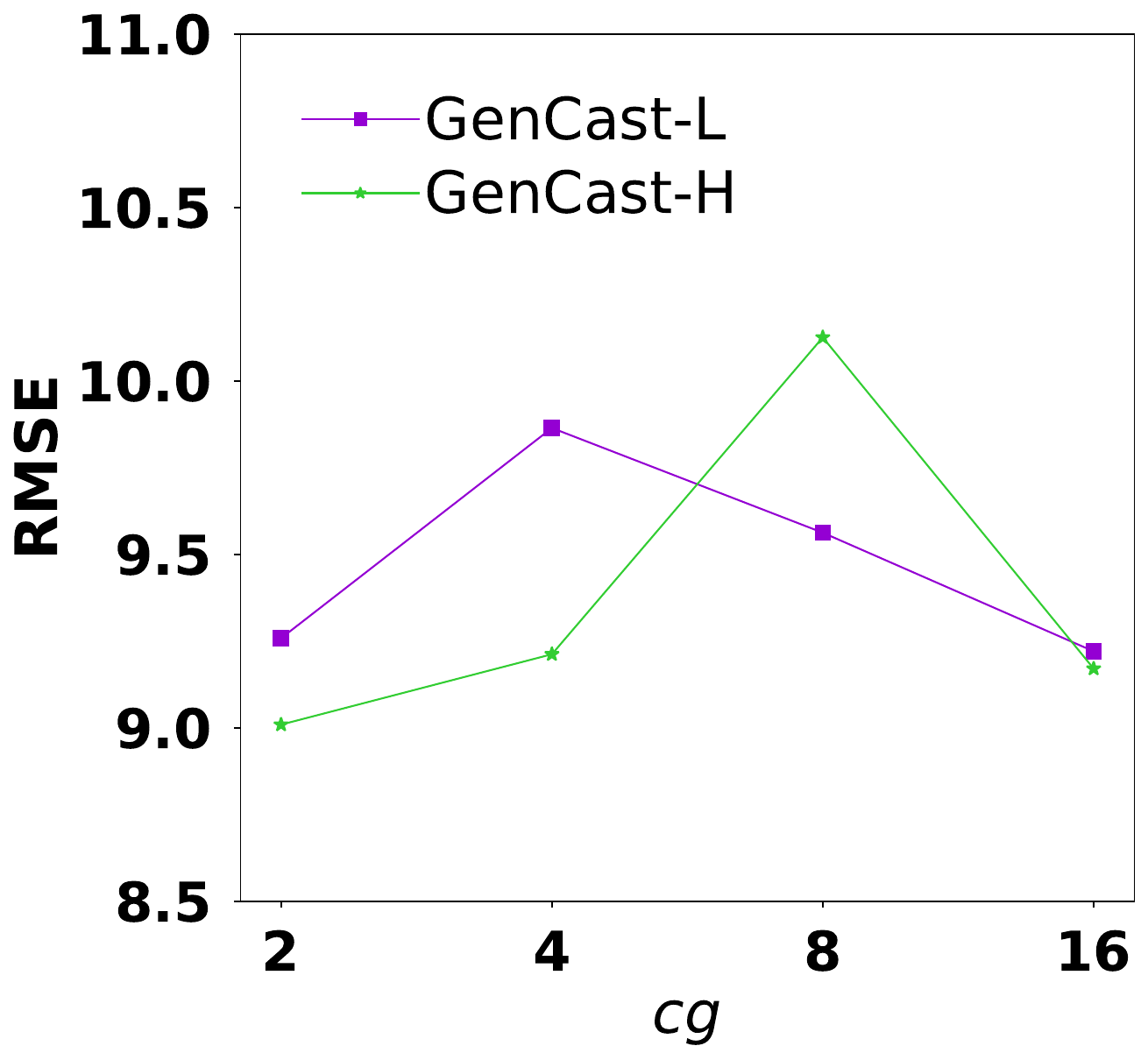}} 

\caption{Impact of hyperparameters (I).}
    \label{fig:exp_hp1}
\end{figure*}

\begin{figure*}[!t]
\centering
    \subfigure[PEMS07]{\includegraphics[width=0.18\textwidth]{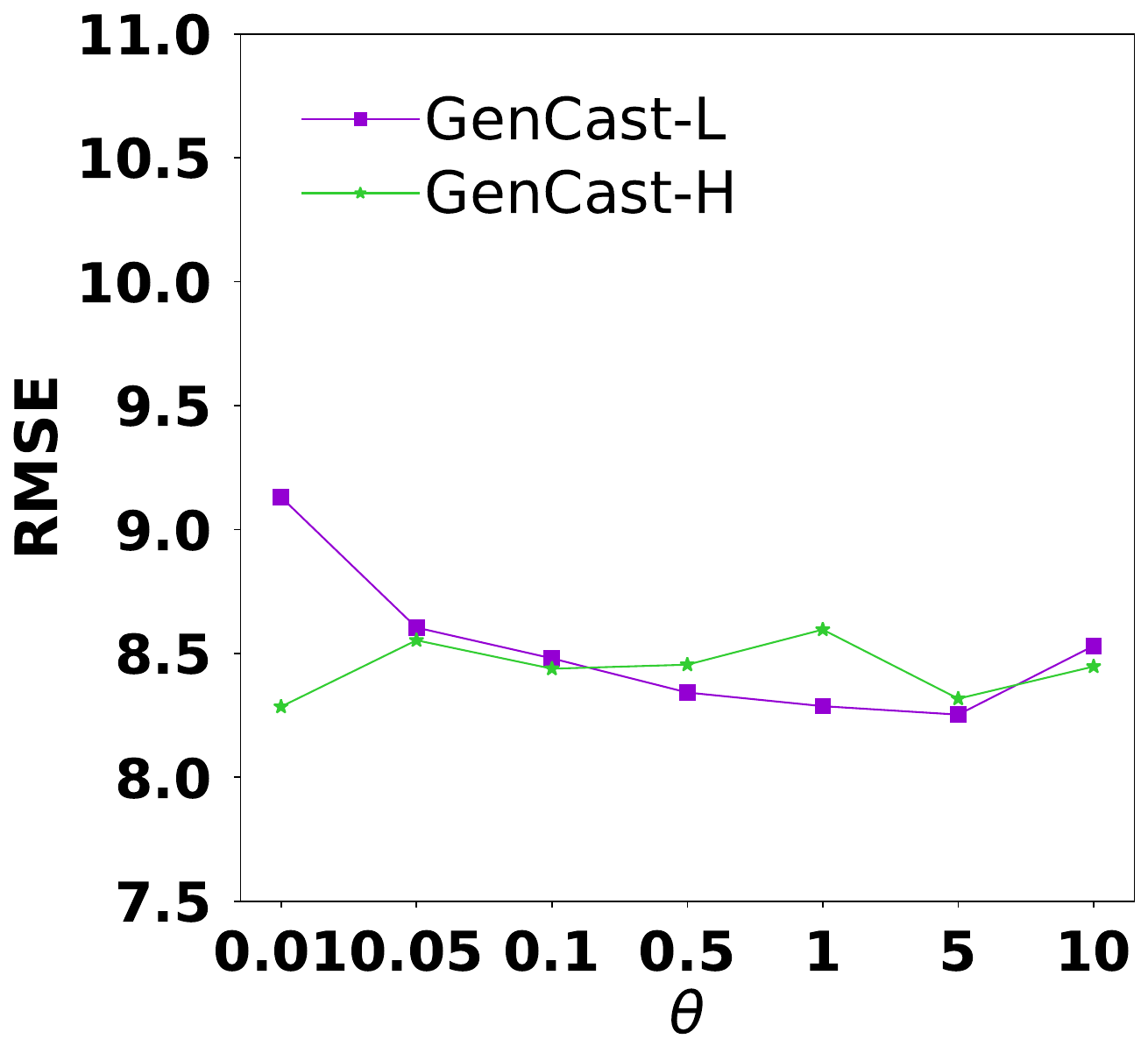}} 
    \hspace{-1mm}
    \subfigure[PEMS08]{\includegraphics[width=0.18\textwidth]{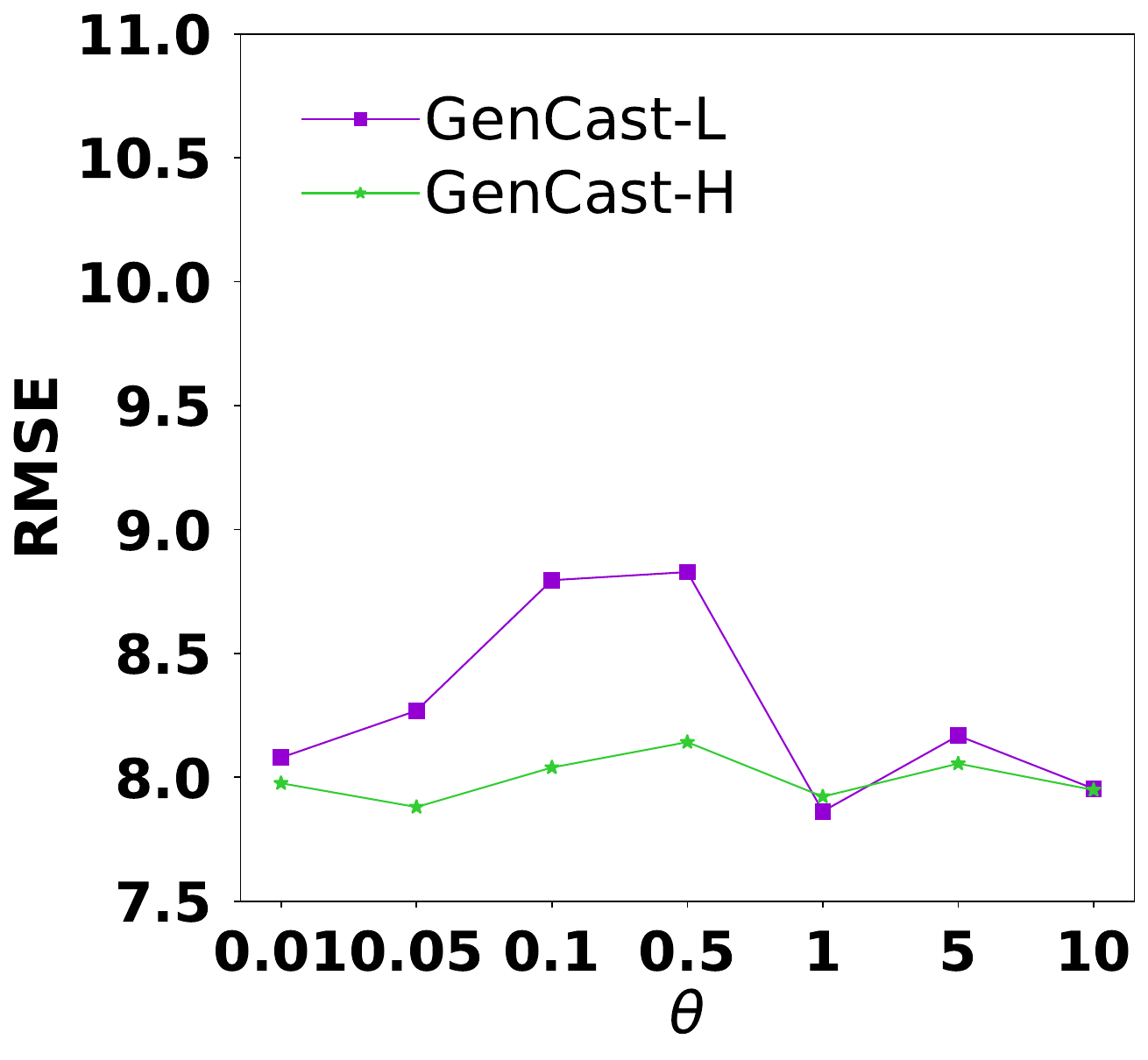}} 
    \hspace{-1mm}
    \subfigure[PEMS-Bay]{\includegraphics[width=0.18\textwidth]{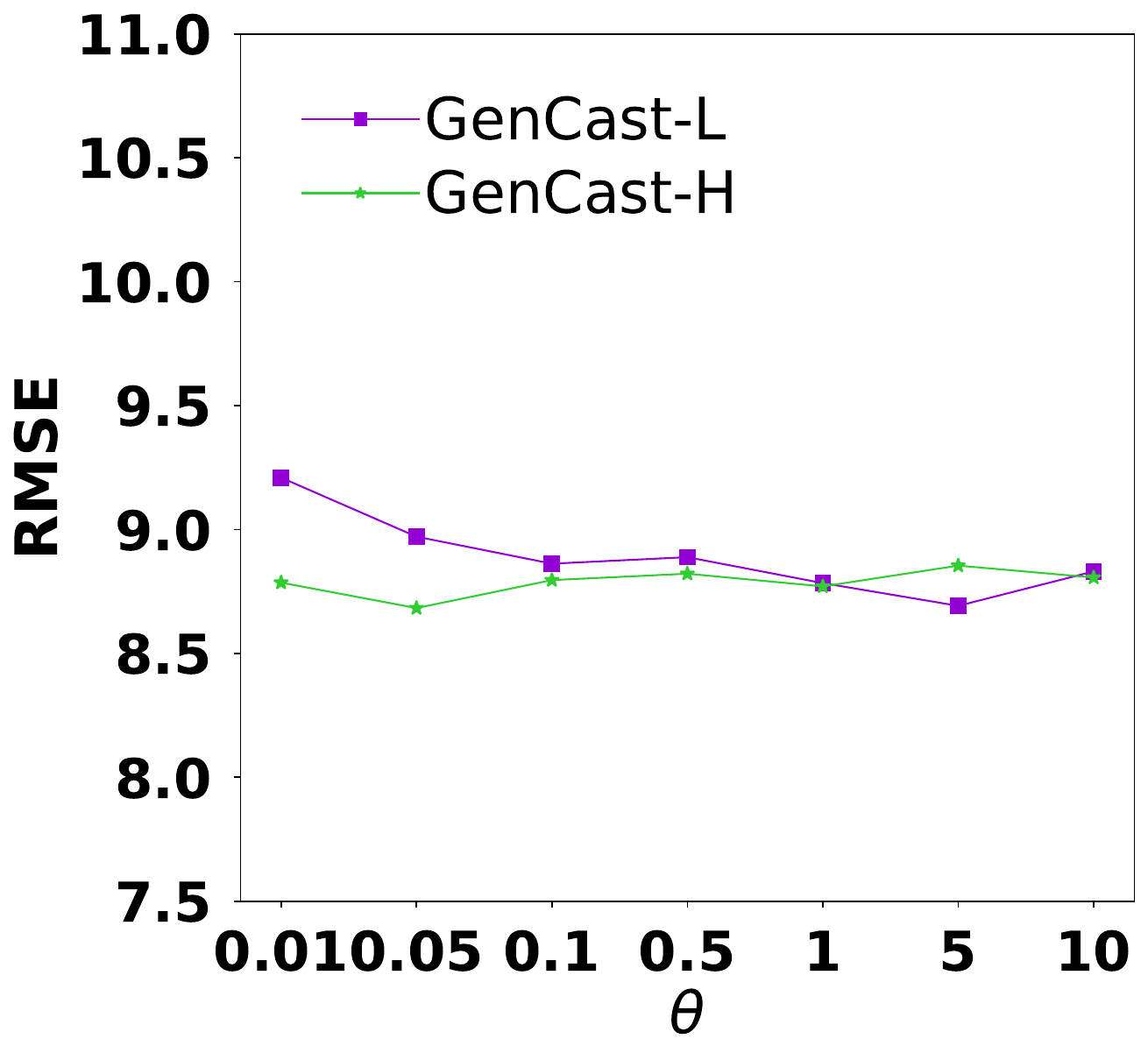}} 
    \hspace{-1mm}
    \subfigure[METR-LA]{\includegraphics[width=0.18\textwidth]{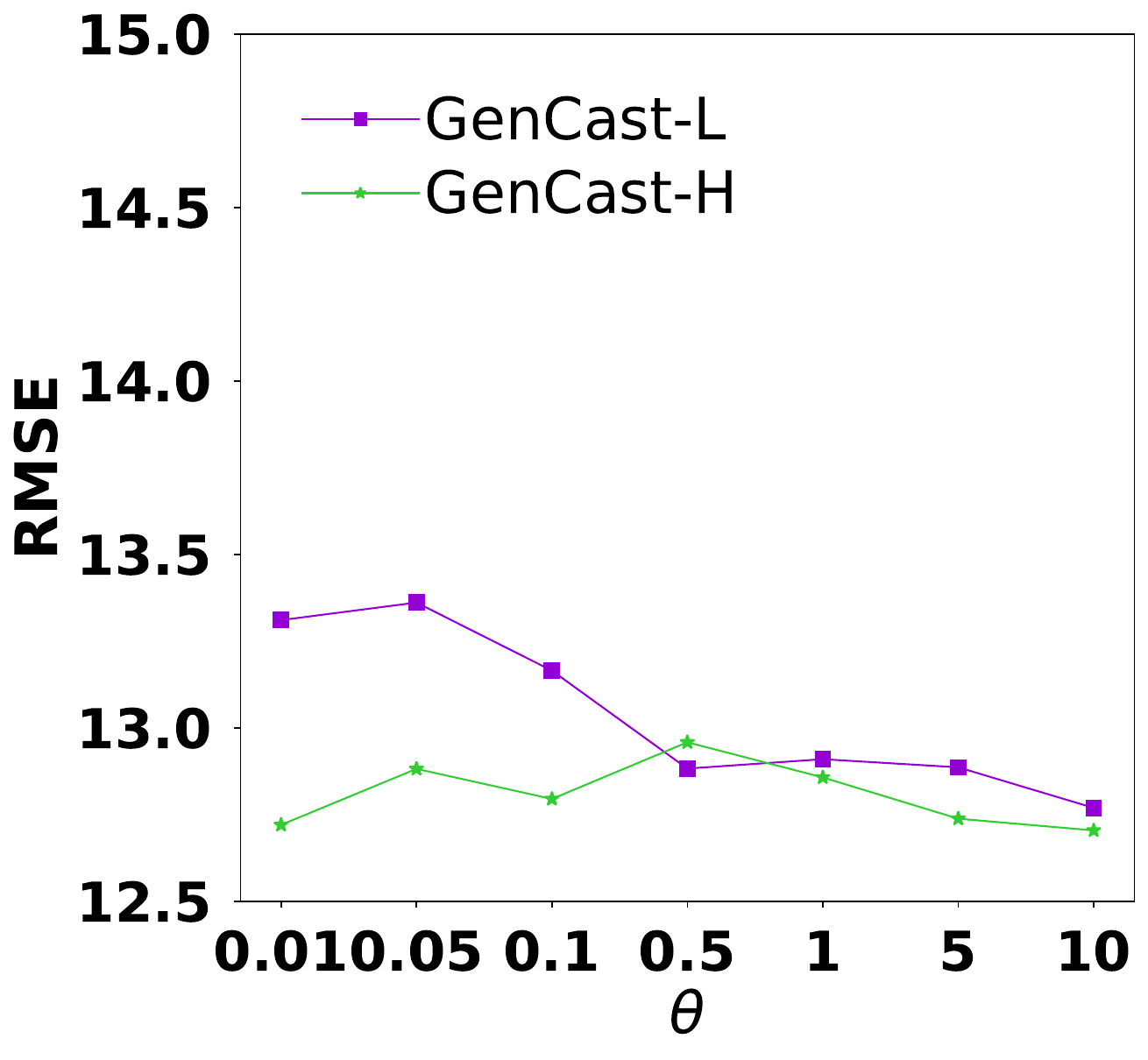}} 
    \hspace{-1mm}
    \subfigure[MEL]{\includegraphics[width=0.18\textwidth]{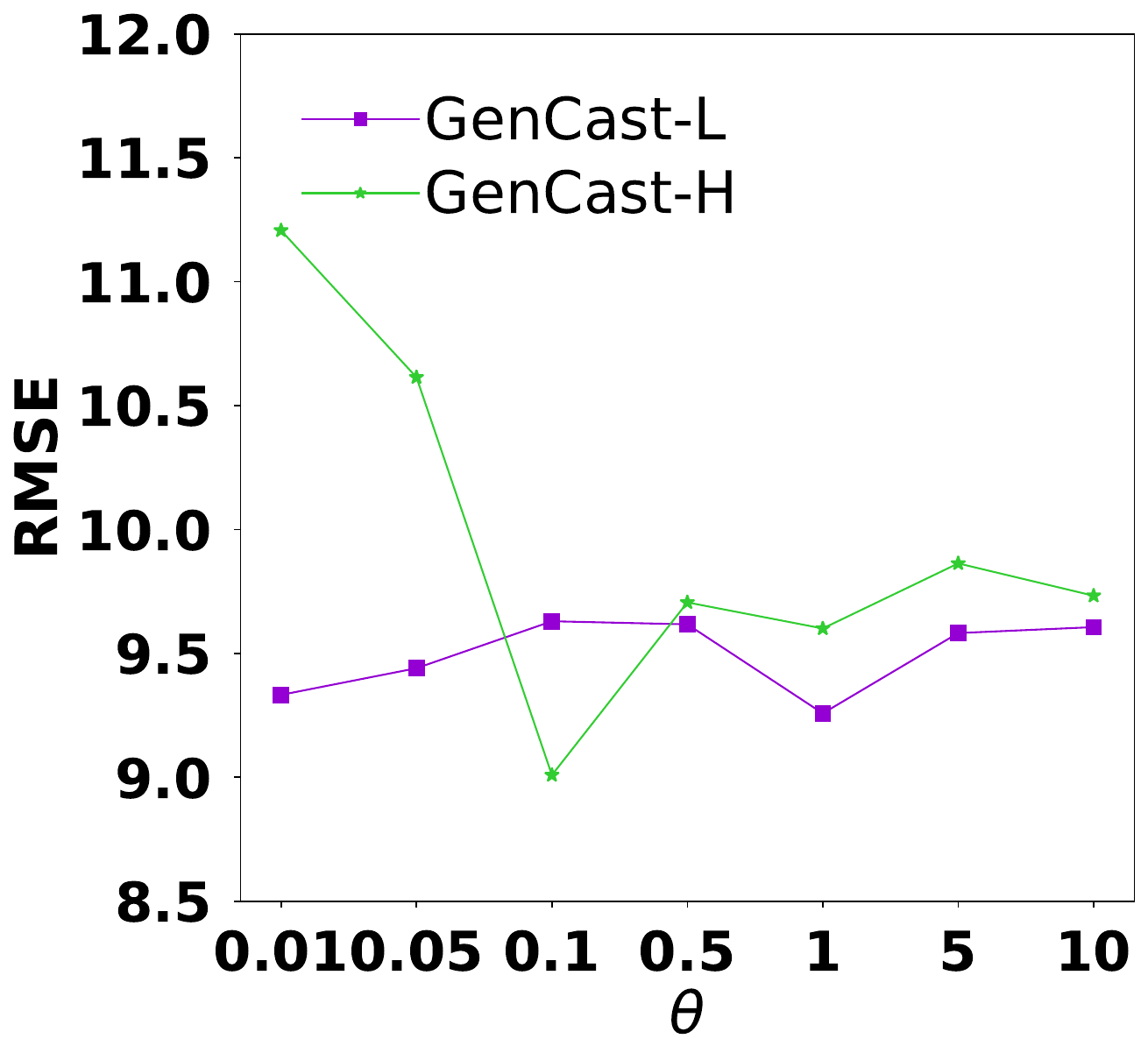}} 
    
    \subfigure[PEMS07]{\includegraphics[width=0.18\textwidth]{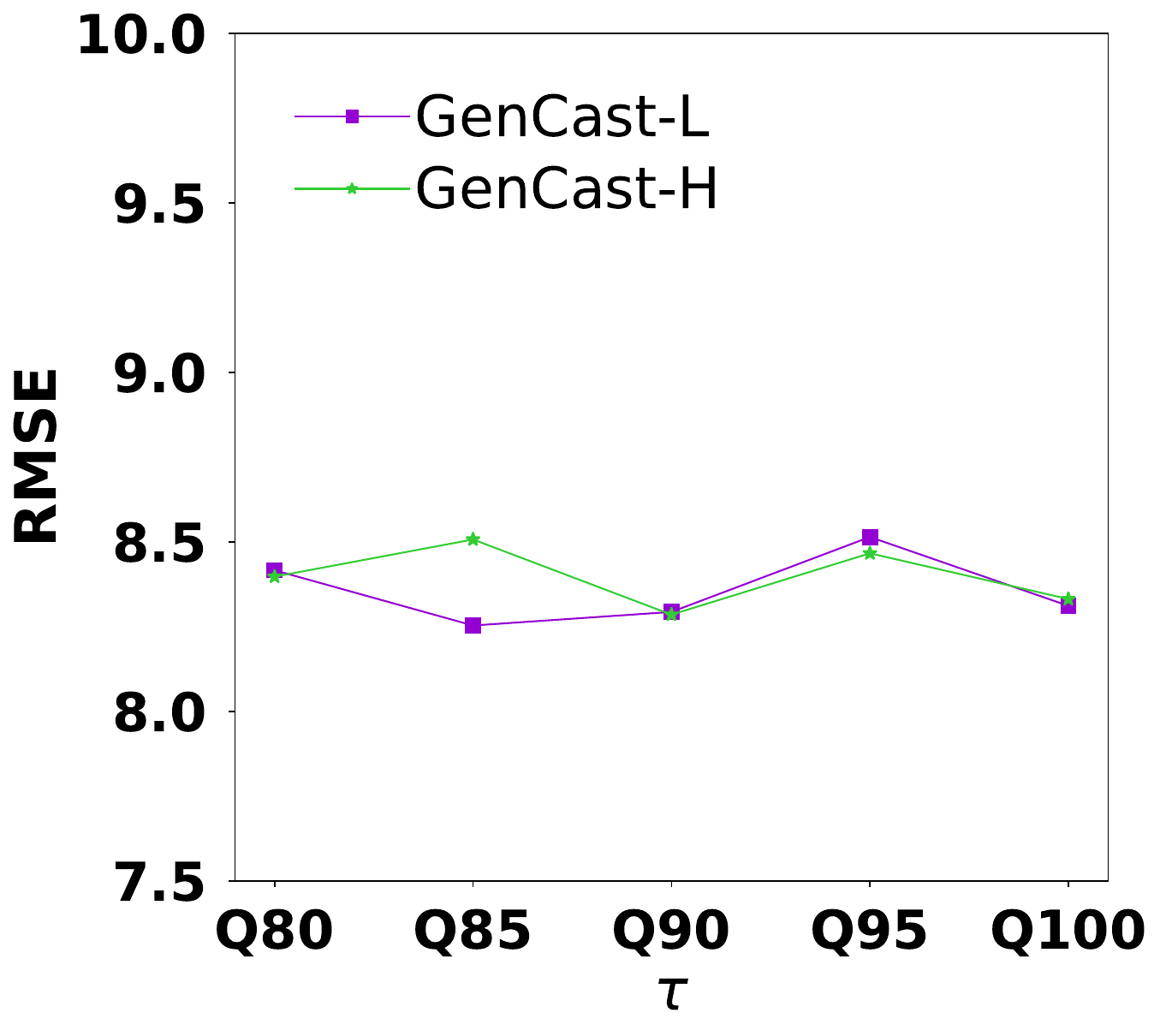}} 
    \hspace{-1mm}
    \subfigure[PEMS08]{\includegraphics[width=0.18\textwidth]{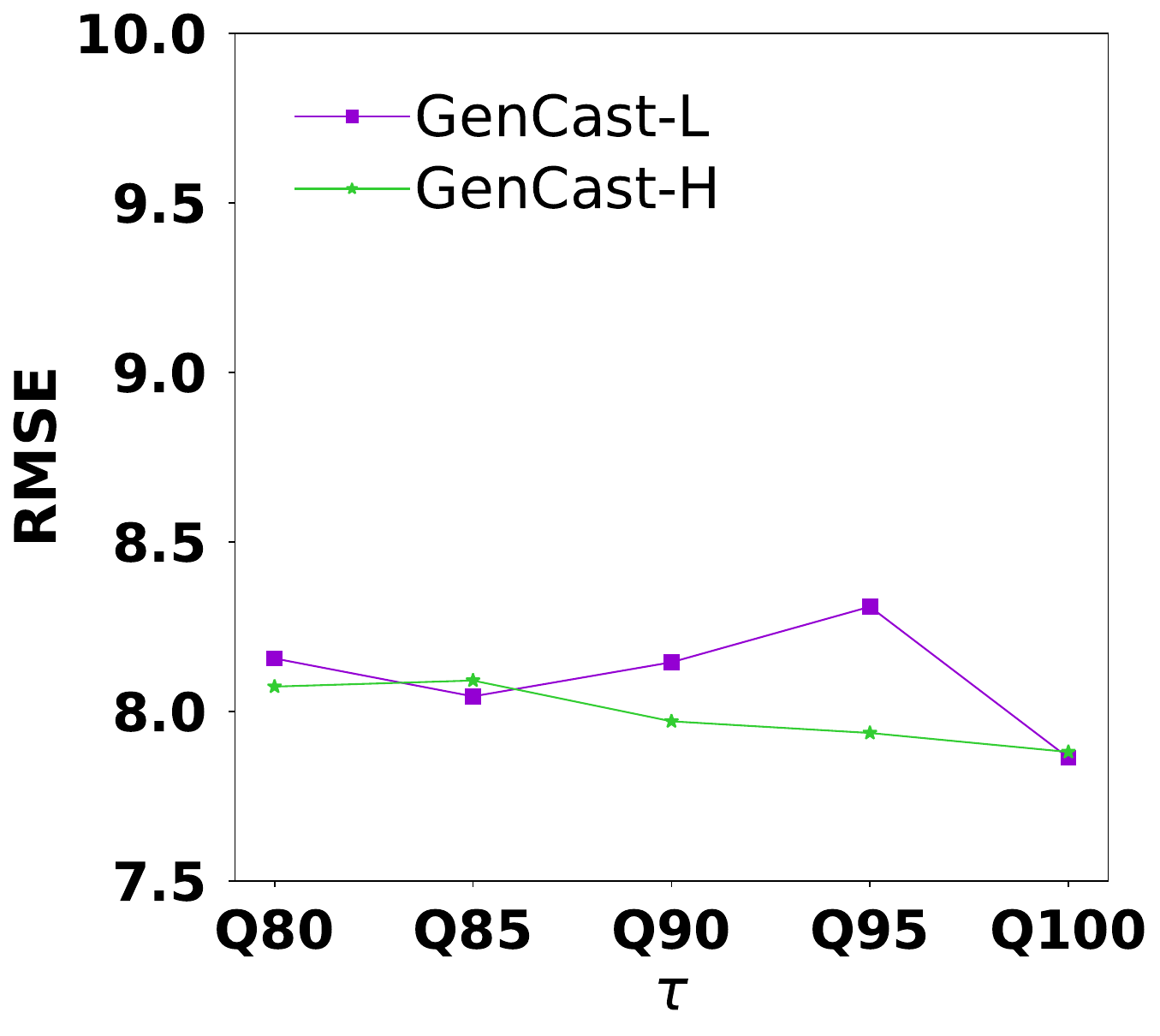}} 
    \hspace{-1mm}
    \subfigure[PEMS-Bay]{\includegraphics[width=0.18\textwidth]{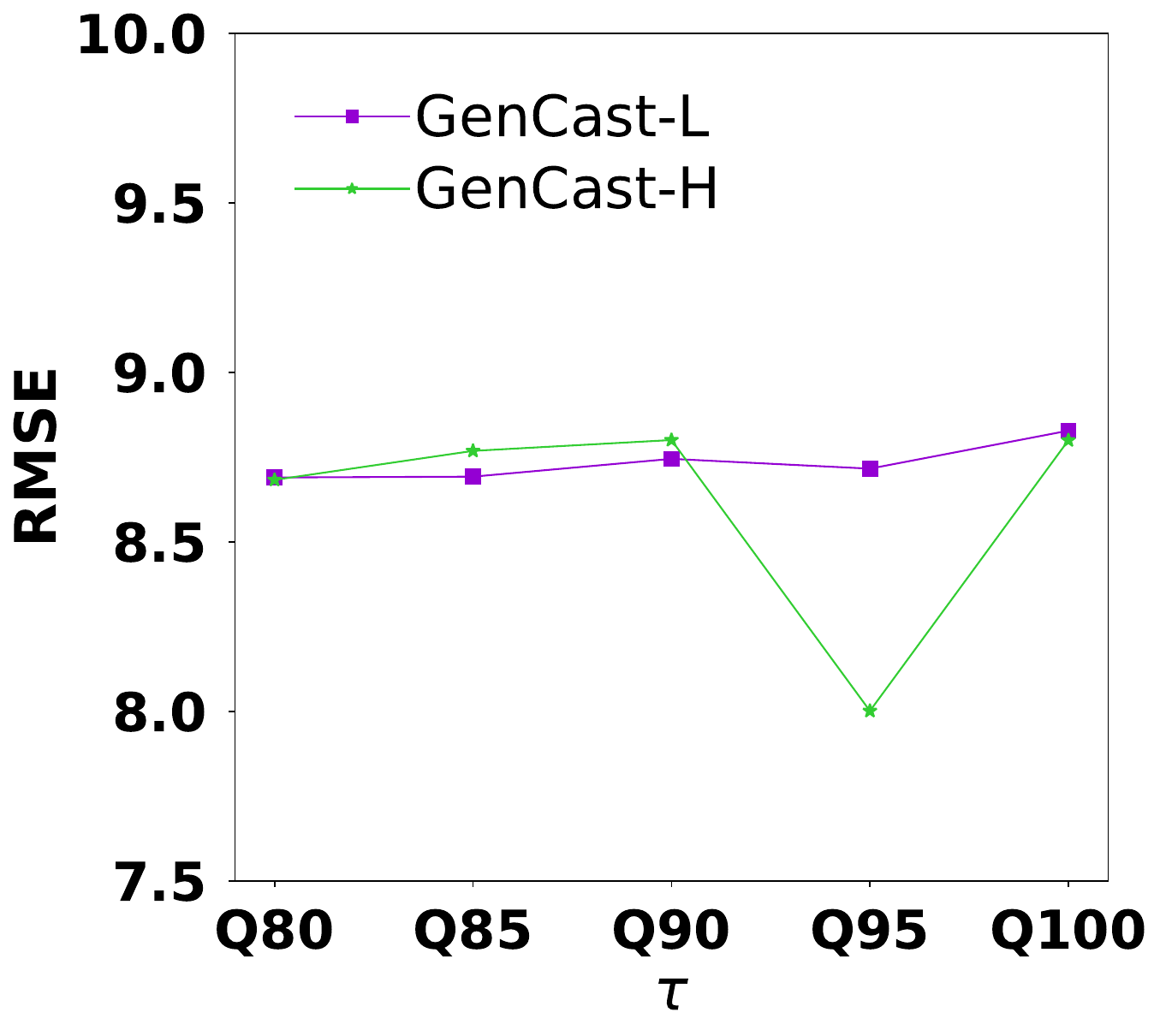}} 
    \hspace{-1mm}
    \subfigure[METR-LA]{\includegraphics[width=0.18\textwidth]{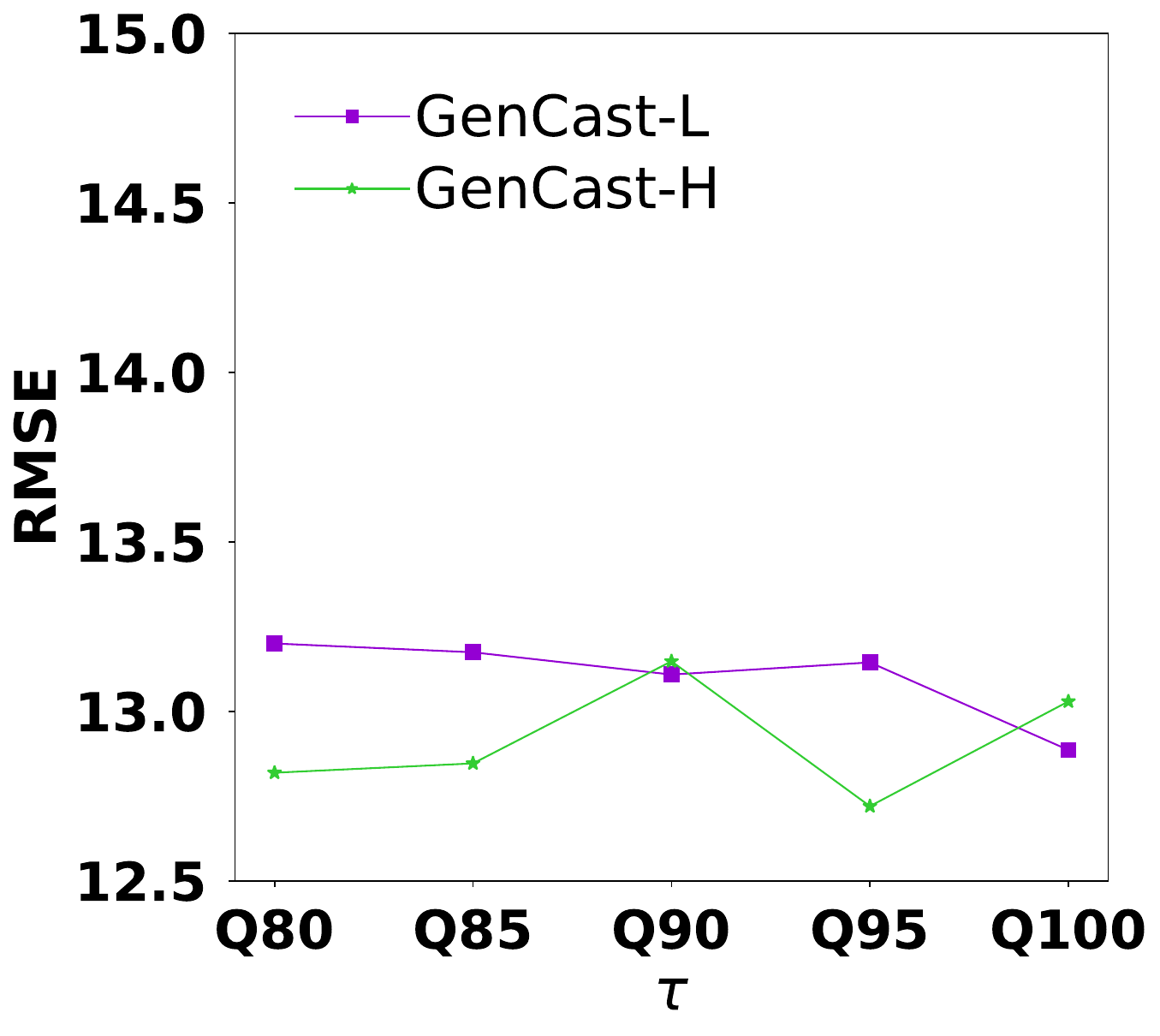}} 
    \hspace{-1mm}
    \subfigure[MEL]{\includegraphics[width=0.18\textwidth]{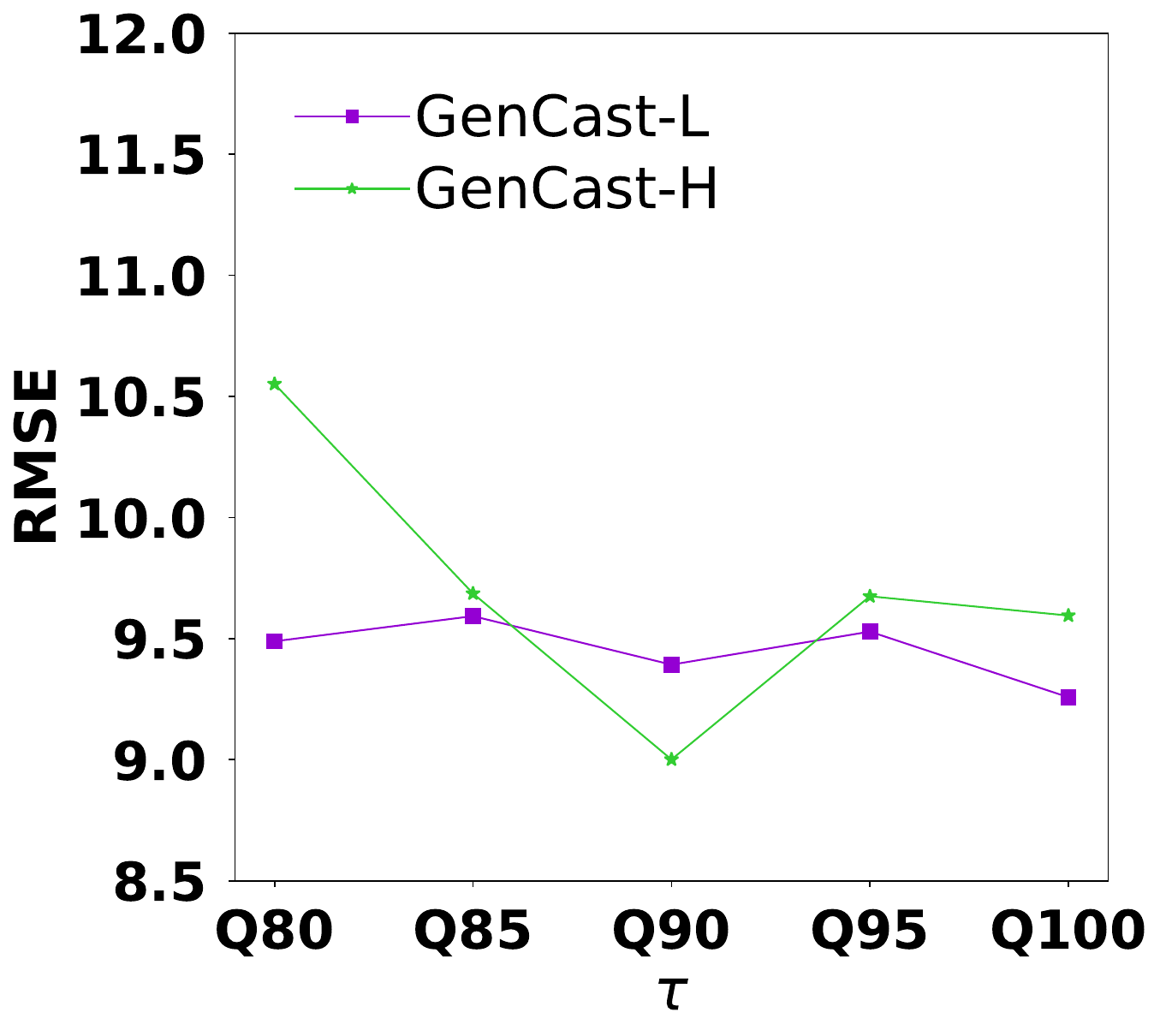}} 
    \centering
    \subfigure[PEMS07]{\includegraphics[width=0.18\textwidth]{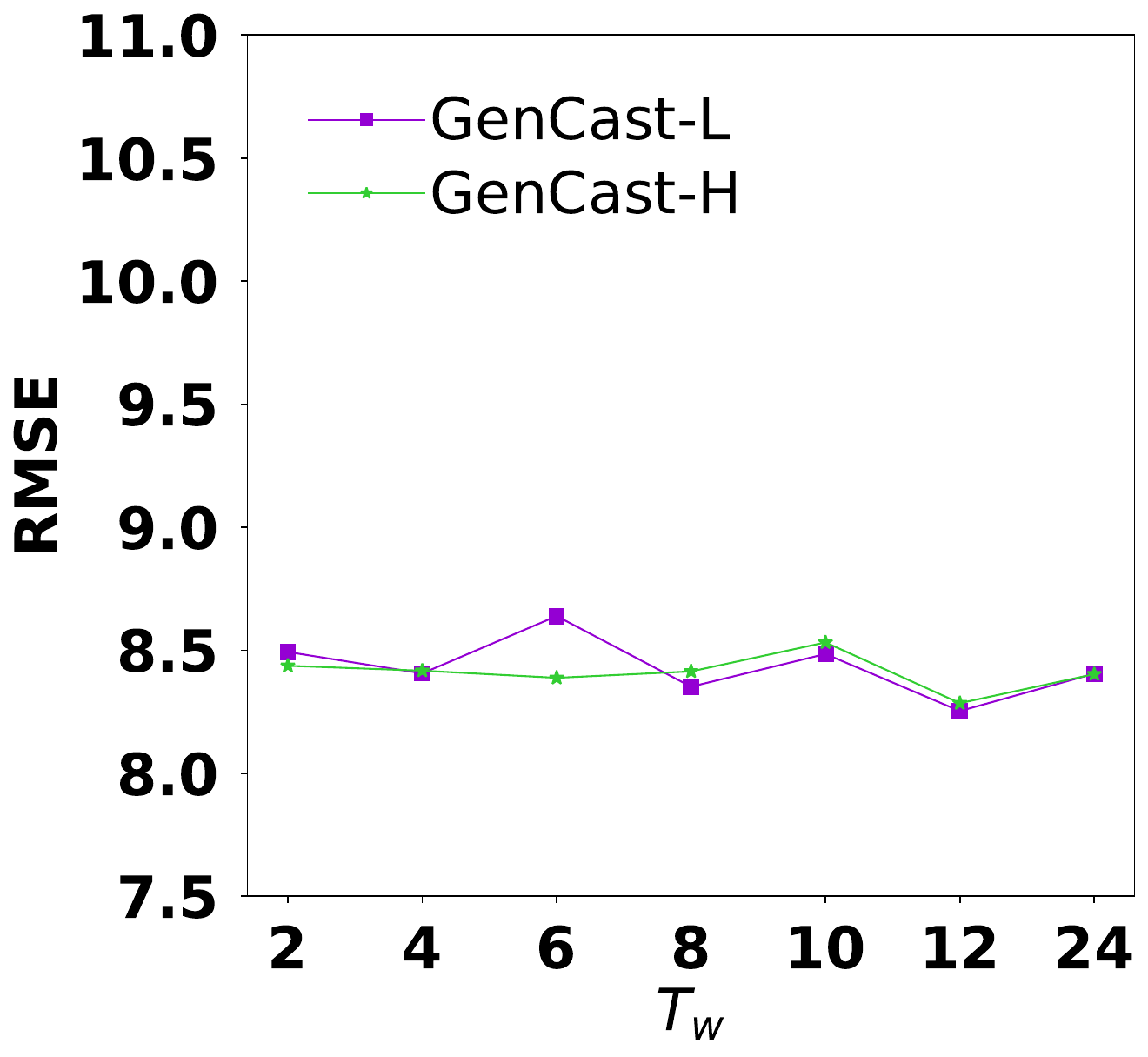}} 
    \hspace{-1mm}
    \subfigure[PEMS08]{\includegraphics[width=0.18\textwidth]{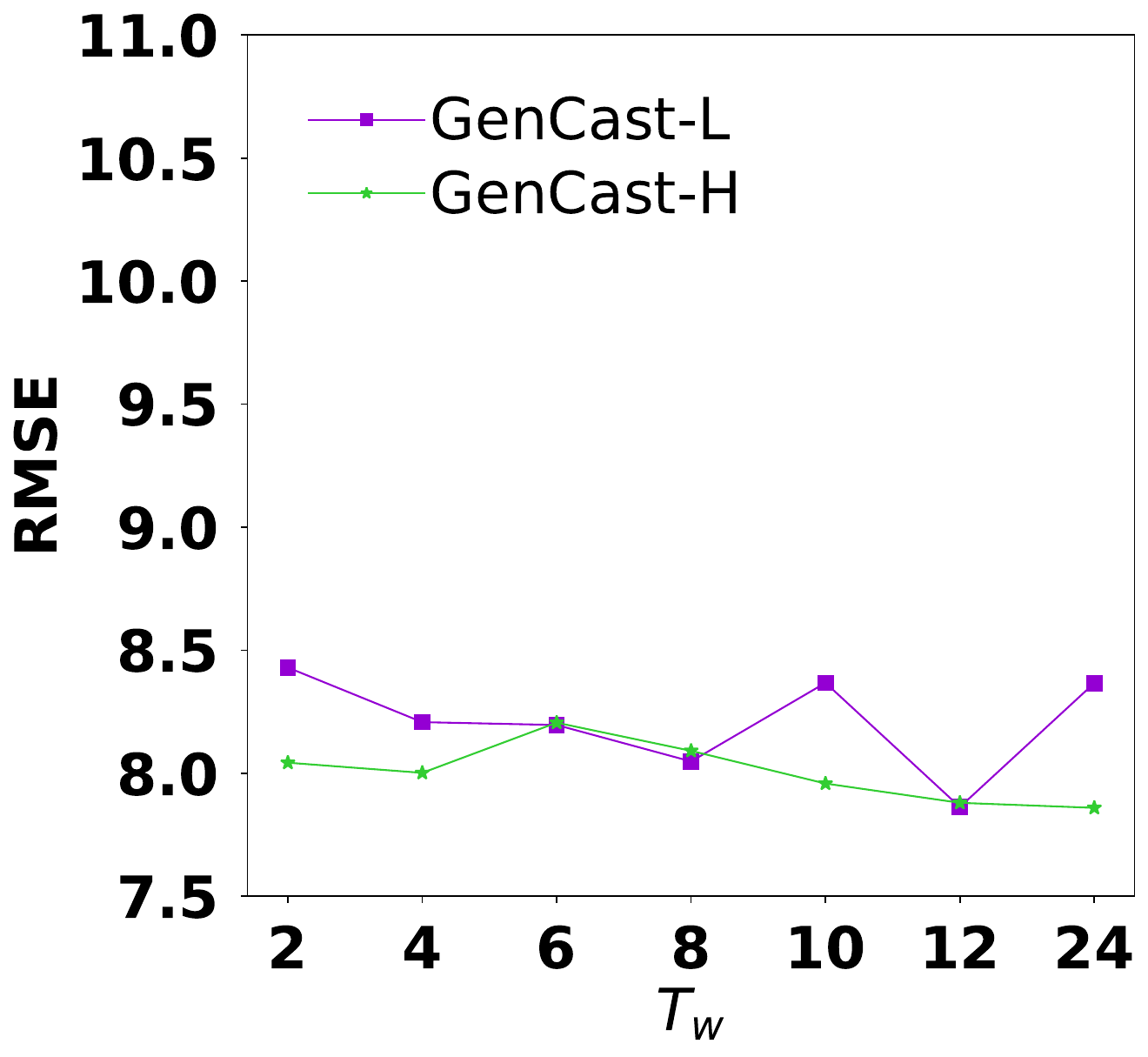}} 
    \hspace{-1mm}
    \subfigure[PEMS-Bay]{\includegraphics[width=0.18\textwidth]{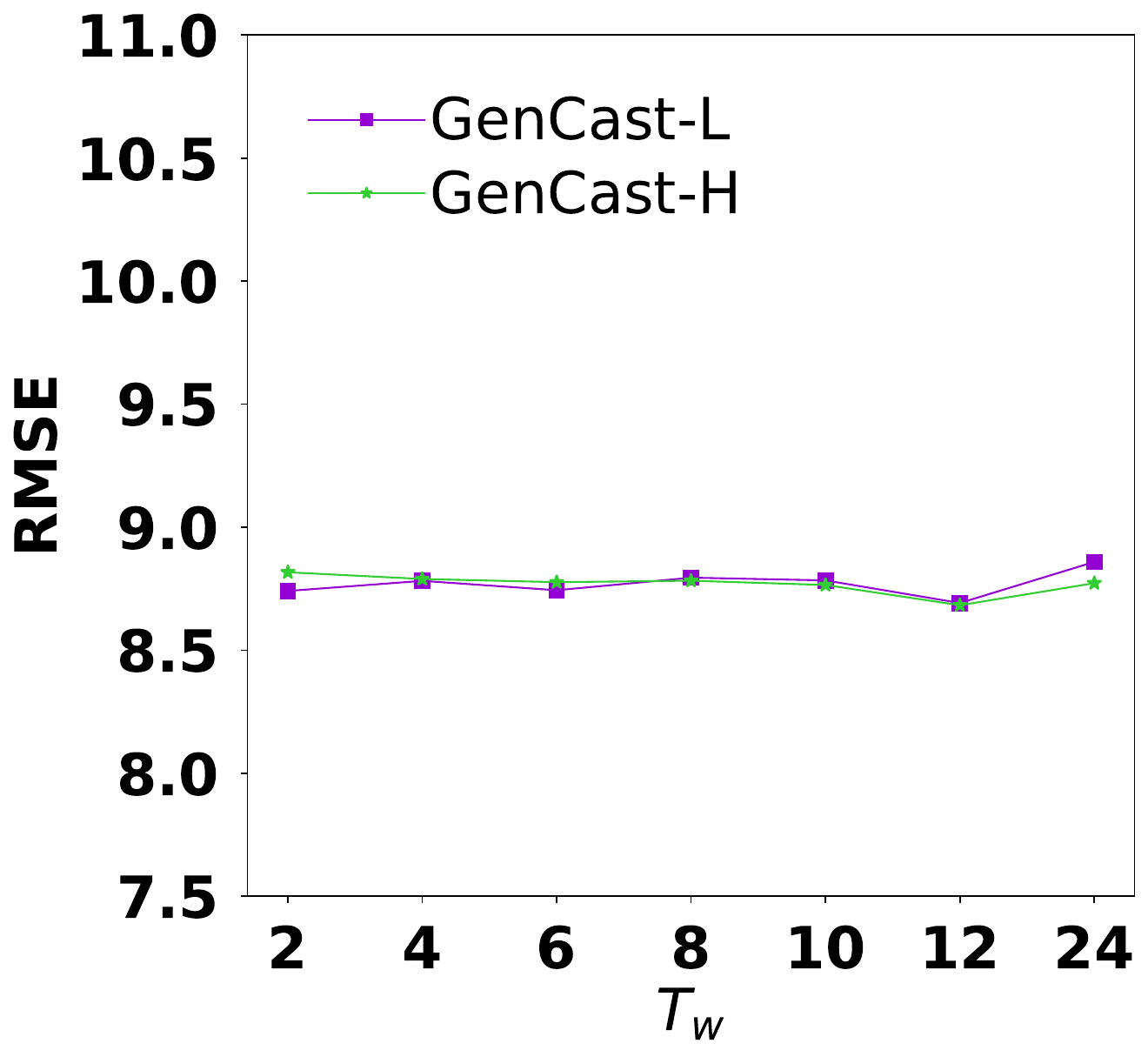}} 
    \hspace{-1mm}
    \subfigure[METR-LA]{\includegraphics[width=0.18\textwidth]{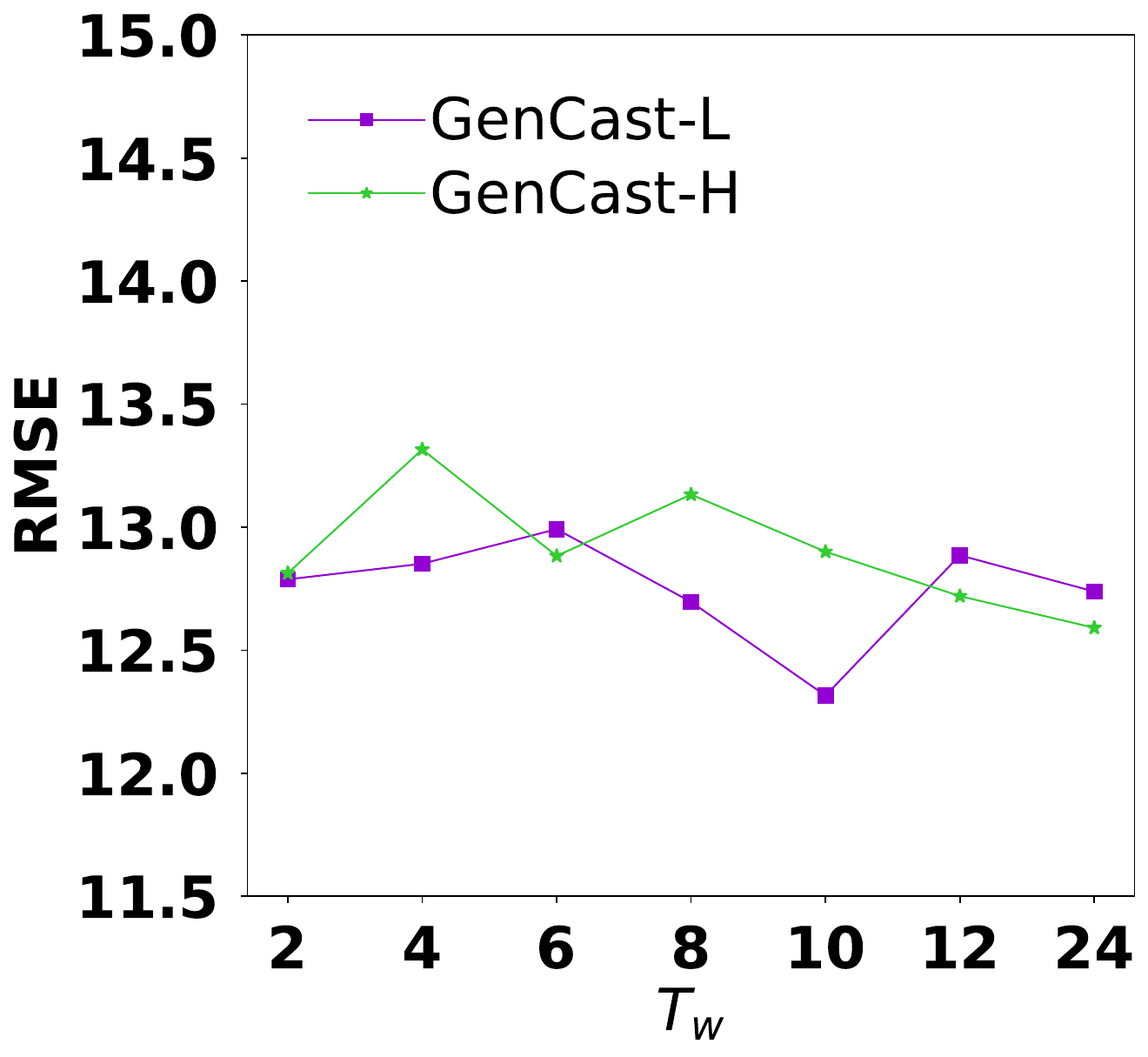}} 
    \hspace{-1mm}
    \subfigure[MEL]{\includegraphics[width=0.18\textwidth]{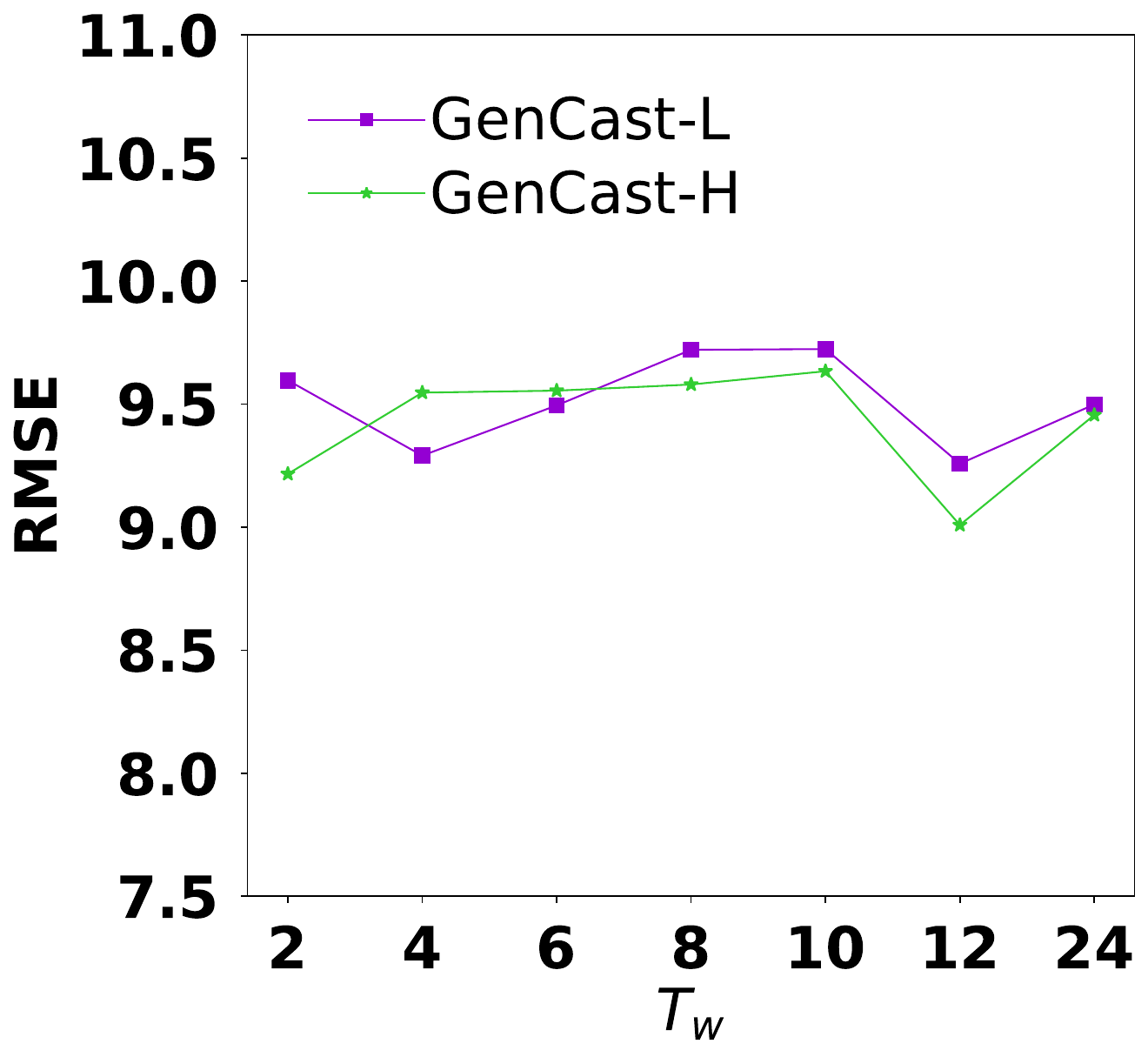}} 
    
    \caption{Impact of hyperparameters (II).}
    \label{fig:exp_hp2}
    %\caption{Model performance vs. ${T_w}$.}
    %\label{fig:look_back}
\end{figure*}

\paragraph{Impact of Hyperparameters.} 
We study the impact of key hyperparameters, including the number of spatial/channel groups in the spatial grouping module, loss weights $\theta$, $\mu$, $\tau$, and weather data input window length $T_w$. 

(1)~\emph{Impact of the weight $\mu$ of the spatial grouping loss.} This parameter controls the strength of the spatial grouping loss. As Fig.~\ref{fig:exp_hp1}a to Fig.~\ref{fig:exp_hp1}e show, $\mu=1$ consistently produces lower errors across datasets and variants, indicating the robustness of our framework to this hyperparameter.

(2)~\emph{Impact of the number of spatial groups $sp$.} We present the impact of varying the number of spatial groups $sp$ in Fig.~\ref{fig:exp_hp1}f to Fig.~\ref{fig:exp_hp1}j. 
We observe that $sp=5$ consistently produces the best performance across all datasets and model variants, except for \model-L on METR-LA where $sp=6$ is  slightly better (though $sp=5$ still achieves competitive results). Notably, this value was selected based solely on \model-L over PEMS08 without further tuning.

The forecast errors show a U-shaped pattern as $sp$ increases: overly small number of groups fails to capture transferable spatial patterns, while overly large ones may capture overly specific, localised features that do not generalise well to unobserved regions. A moderate setting (e.g., $sp=5$ or $sp=6$) strikes a balance, effectively capturing shared structures while mitigating local noise. 

(3)~\emph{Impact of the number of channel groups $cg$.} As shown in Fig.~\ref{fig:exp_hp1}k to Fig.~\ref{fig:exp_hp1}o, $cg=2$ produces  consistently low errors across datasets and model variants. This may be attributed to the fact that the number of feature channels $D$ is typically much smaller than the number of spatial nodes $N$, and $cg$ is required to divide $D$ evenly. Using a small $cg$ such as 2 offers a good trade-off between preserving feature coherence and enabling effective grouping. In contrast, overly large values of $cg$ may fragment the features, making it more difficult to align with spatial groupings and weakening model generalisability.

(4)~\emph{Impact of th weight $\theta$ of the physics  loss.} This parameter controls the contribution of the physics loss. As shown in Fig.~\ref{fig:exp_hp2}a to Fig.~\ref{fig:exp_hp2}e, {\model-L} has lower errors when $\theta$ has larger value (i.e., 1 or above), while {\model-H} has lower errors when  $\theta$ has smaller values (i.e., 0.1 or below). These are expected, as the underlying spatial embeddings used in \model-L and \model-H differ, resulting in large variations in the magnitude of the physics loss (e.g., 95th percentile: 21 vs. 992 on PEMS07). Consequently, the appropriate value of $\theta$ depends on the scale of the physical residuals induced by each embedding type.

(5)~\emph{Impact of the quantile used in dynamic Huber loss $\tau$.} This parameter is used during the warm-up stage to optimise threshold $\delta$ for the Huber loss applied to physics residuals. It specifies the percentile of RMSE loss values, determining what proportion of residuals is treated as outliers in the adaptive calibration. As shown in Fig.~\ref{fig:exp_hp2}f to Fig.~\ref{fig:exp_hp2}j, \model-L shows greater robustness to this hyperparameter. We set $\tau=100\%$ for it across most datasets, except for PEMS-Bay and PEMS07 where $\tau=85\%$. This may be attributed to the use of frozen LLM-based spatial embeddings in \model-L, which leads to more stable physics residuals during training. \model-H relies on trainable GeoHash embeddings, making it more susceptible to outliers. Thus, suppressing the impact of large residuals via an adaptive Huber loss is particularly beneficial for \model-H.

(6)~\emph{Impact of the weather look-back window length $T_w$.} As shown in Fig.~\ref{fig:exp_hp2}k to Fig.~\ref{fig:exp_hp2}o, our model forecast performance varies with $T_w$, which controls how many previous hours of weather data are used. We observe that $T_w=12$ generally yields the lowest errors. This suggests that a 12-hour window provides a good temporal context to capture the delayed or cumulative impacts of weather on traffic dynamics. In contrast, short windows (e.g., $T_w=2$ or $4$) may miss such dependencies, while overly long windows (e.g., $T_w=24$) may introduce irrelevant or outdated signals, thus diluting the more useful signals.

\paragraph{Generalisability of the External Signal Encoder.} As shown in Table~\ref{tab:with_wx}, incorporating our external signal encoder (\textbf{-w/wx}) consistently improves the performance of both INCREASE and KITS across all metrics. This confirms that our design, leveraging dynamic external signals to guide learning and enhance model generalisability to unobserved regions, is applicable beyond our  model \model. The improvements are particularly pronounced in MAPE and R\textsuperscript{2}, indicating that external signals help reduce magnitude errors and better capture large-scale patterns. These results highlight the potential of integrating weather and other exogenous data as general-purpose enhancements to existing advanced models.

\begin{table}[!t]
    \centering
    {\small
    \begin{tabular}{lcccc}
        \hlineB{3}
        \textbf{Models} & \textbf{RMSE}$\downarrow$ & \textbf{MAE}$\downarrow$ & \textbf{MAPE}$\downarrow$ & \textbf{R$^2$}$\uparrow$ \\
        \hline \hline
        {INCREASE} & 8.534 & 6.045 & 2.730 & 0.177 \\
        {INCREASE-w/wx} & 8.493 & 6.027 & 2.693 & 0.321 \\
        \hline
        \rowcolor{gray!15}
        Improve & 0.48\% & 0.30\% & 1.37\% & 81.36\% \\
        \hline \hline
        {KITS} & 8.704 & 5.513 & 1.776 & 0.314 \\
        {KITS-w/wx} & 7.715 & 4.582 & 0.746 & 0.496 \\
        \hline
        \rowcolor{gray!15}
        Improve & 12.82\% & 20.32\% & 138.07\% & 57.96\% \\
        \hlineB{3}
    \end{tabular}
}
\caption{Generalisability of the external signal encoder.}
\label{tab:with_wx}
\end{table}

\begin{figure}[!t]
    \centering
    \subfigure[\small Split type 1]{\includegraphics[width=0.18\textwidth]{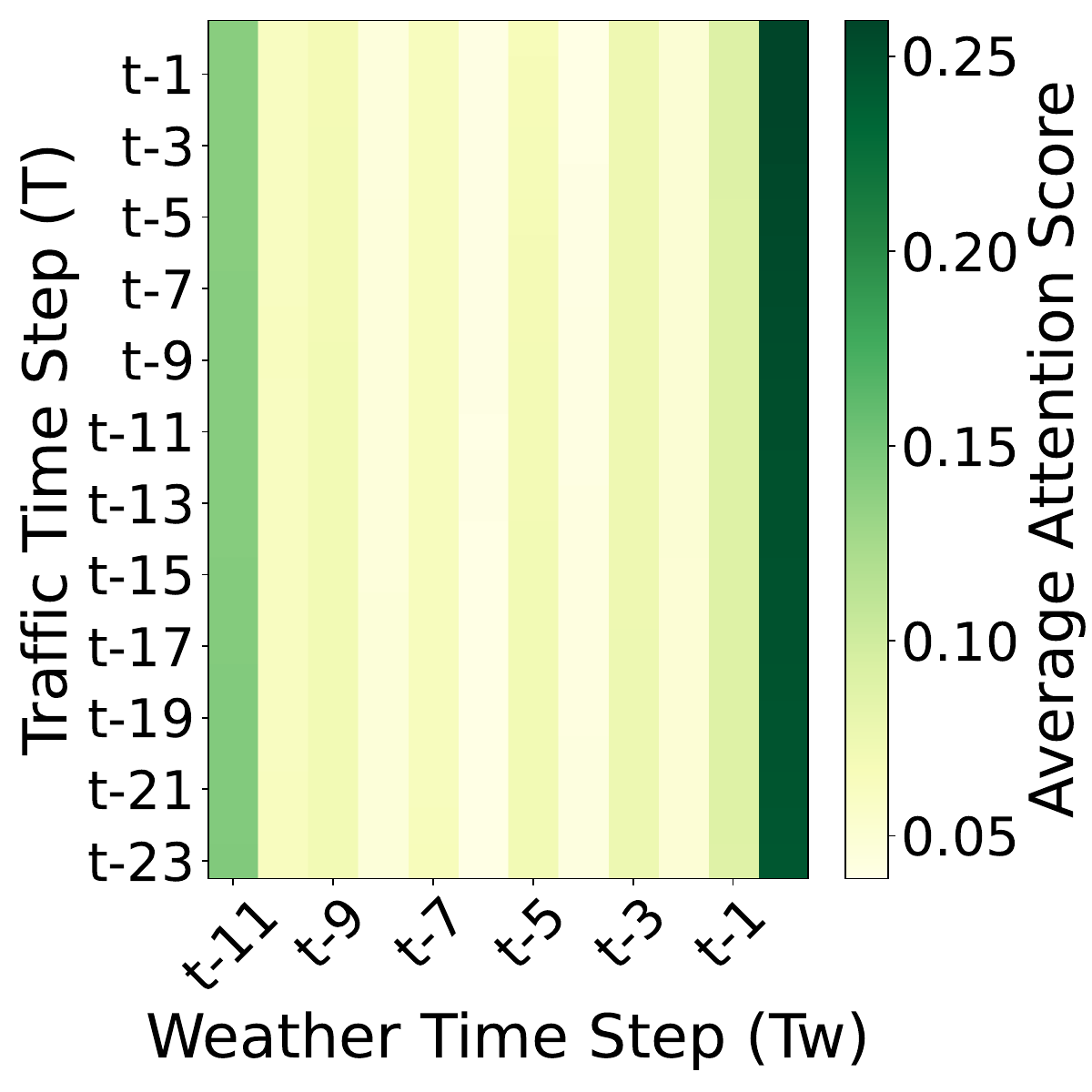}} 
    \hspace{-1mm}
    \subfigure[\small Split type 2]{\includegraphics[width=0.18\textwidth]{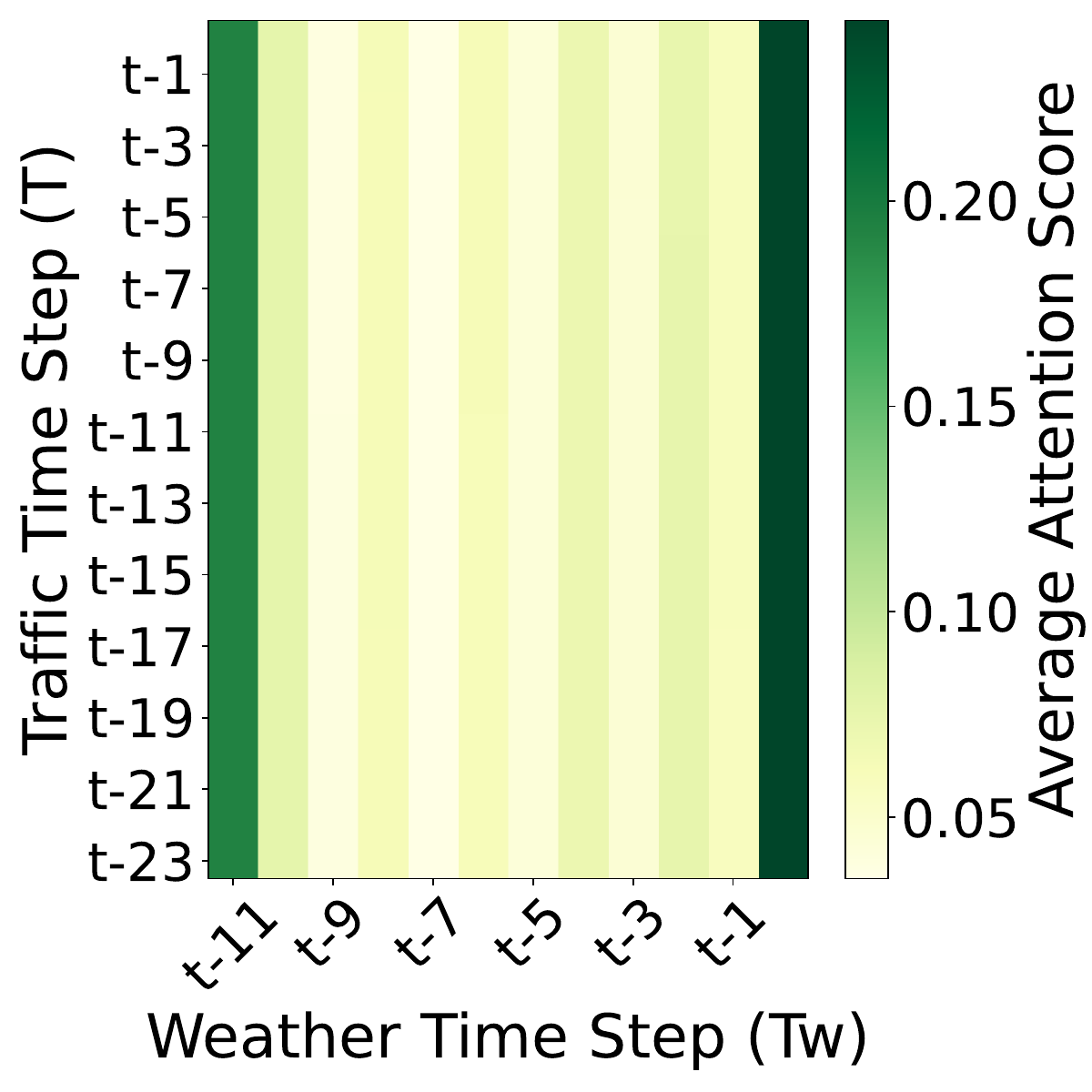}} 
    \hspace{-1mm}
    \subfigure[\small Split type 3]{\includegraphics[width=0.18\textwidth]{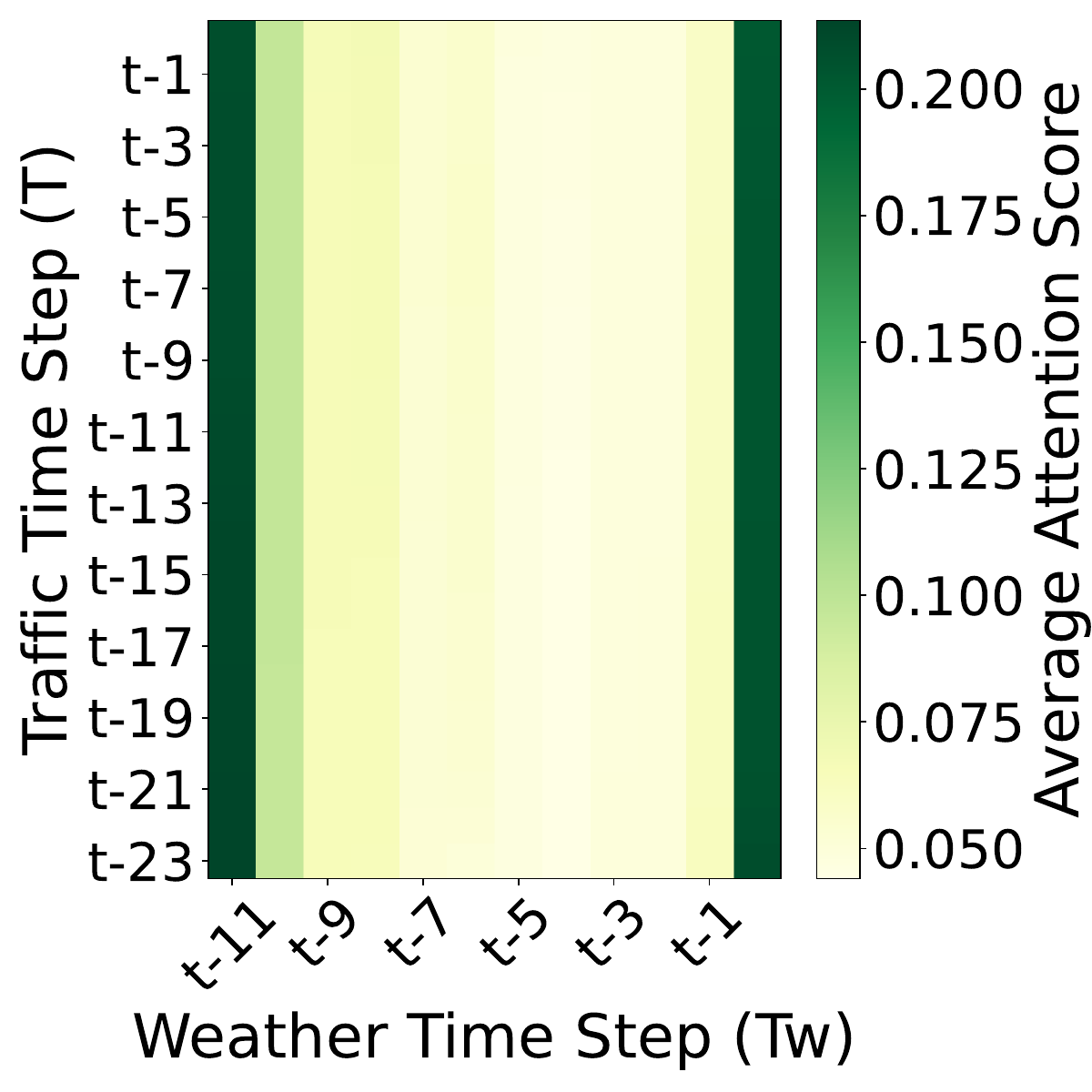}} 
    \hspace{-1mm}
    \subfigure[\small Split type 4]{\includegraphics[width=0.18\textwidth]{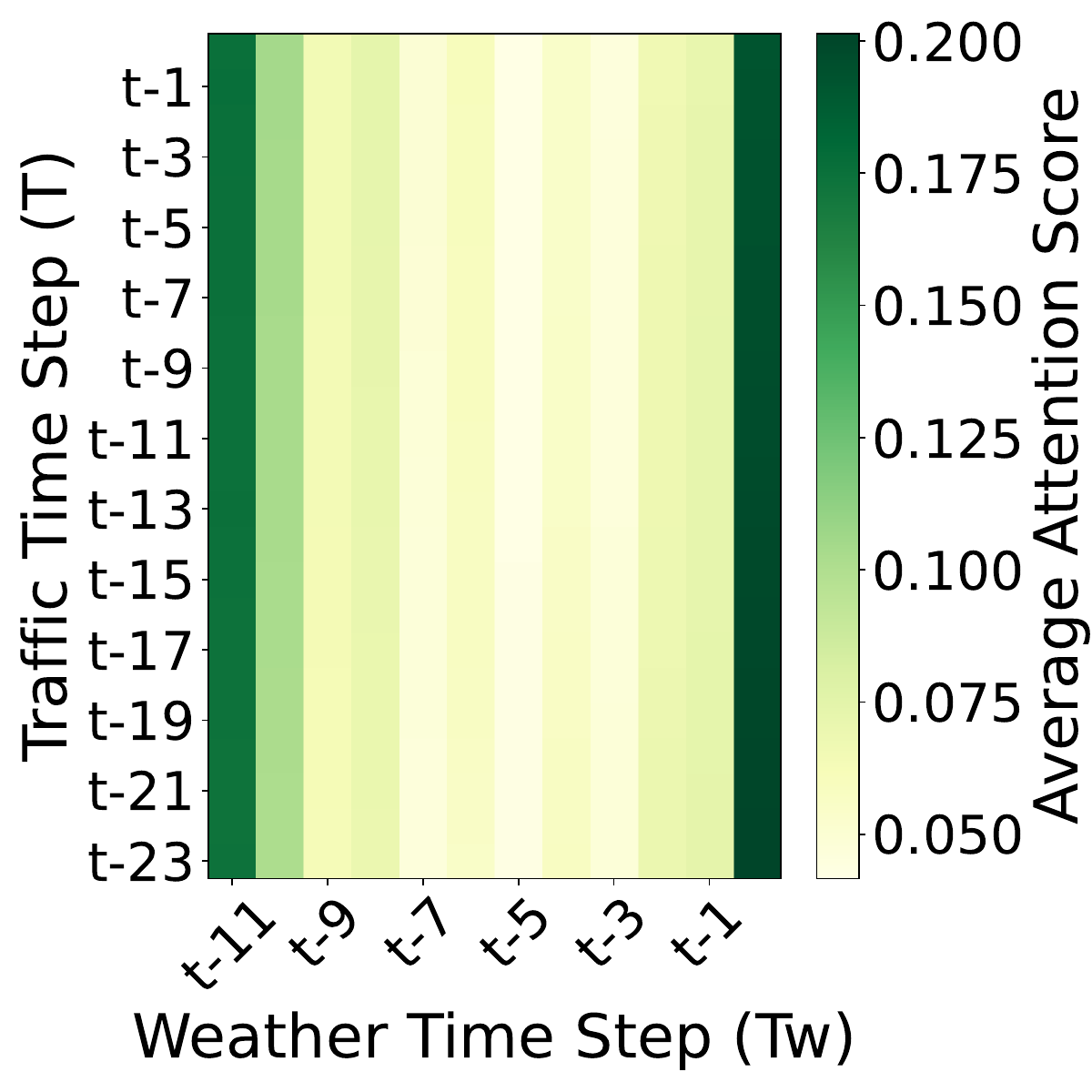}} 
    \caption{\small Average temporal attention to weather on the PEMS07.}
    \label{fig:attn_vis}
\end{figure}

\paragraph{Visualisation Weather-Traffic Attention.}
To understand the overall temporal alignment between traffic forecasts and weather input, we compute an average attention heatmap over the PEMS07 dataset. We aggregate attention scores across all nodes and all test samples, resulting in a $T \times T_w$ matrix, where each entry indicates the average attention weight assigned from a given weather time step to a  traffic prediction step. This global view reveals how \model\ utilises weather information at different temporal offsets.

The heatmap exhibits a bi-modal attention distribution. \model\ consistently focuses on both immediate weather observations (e.g., at $t$) and longer-range inputs (notably, at $t-11$ and $t-10$). This pattern suggests a reliance on both short-term conditions and broader temporal contexts. The periodicity in the distribution is likely a by-product of the 2-hour sliding window used during data generation, which introduces recurring emphasis on specific time steps.
Furthermore, we observe slight variations in the distribution across different data splits, indicating that weather–traffic correlations may vary subtly between regions.

\end{document}